\documentclass[10pt,twocolumn,letterpaper]{article}

\usepackage{wacv}       

\usepackage{booktabs}
\usepackage{placeins}

\makeatletter
\renewcommand\paragraph{%
  \@startsection{paragraph}{4}{\z@}%
    {0pt}%
    {-1em}%
    {\normalfont\normalsize\bfseries}}
\makeatother

\definecolor{wacvblue}{rgb}{0.21,0.49,0.74}
\usepackage[breaklinks,colorlinks,allcolors=wacvblue]{hyperref}

\def\wacvPaperID{2081} 
\def\confName{WACV}
\def\confYear{2027}

\title{DISTRACT: Degradation-Indexed Study of Textual Reference Across Competing Targets}

\author{
Dan Zlotnikov \qquad Ohad Ben-Shahar\\
Ben-Gurion University of the Negev, Israel\\
{\tt\small danzlot@post.bgu.ac.il \quad ben-shahar@cs.bgu.ac.il}
}

\begin{document}
\maketitle
\begin{abstract}

\itshape

Vision-language trackers (VLTs) accept descriptions intended to assist tracking, but accepting text does not establish that its content changes tracking decisions, particularly as visual evidence deteriorates. We introduce DISTRACT, a matched-intervention benchmark that holds video, target, and distractor fixed while crossing four text priors with four image degradations along a controlled severity axis. We evaluate five VLTs on four datasets using behavioral contrasts, layerwise representation probes, and activation patching.

Molmo2 and JointNLT exhibit directed text use. Some effects contract under degradation, whereas others show no significant endpoint change after multiplicity correction. UVLTrack, MMTrack, and DUTrack respond only weakly to changes in text semantics, with no directed behavioral effect surviving FDR correction in the tested settings. Model-internal analyses reveal that at decision-near states, correct-versus-conflicting text is decodable in Molmo2 and JointNLT but not in the other three systems. In the responsive models, decodability peaks at late states where replacement produces little behavioral change, whereas replacing selected earlier states redirects conflict-associated trajectories toward the target. Because matched-correct donors do not significantly outperform unrelated donors, these interventions identify earlier intervention-sensitive states without establishing target-specific semantic rescue. DISTRACT thus exposes when and where text affects tracking under degraded visual evidence.

\end{abstract}
    
\section{Introduction}
\label{sec:intro}

Visual object tracking must preserve an object's identity as appearance,
position, and context change over time. Vision--language trackers (VLTs) add a
natural-language specification---for example, ``the player in white moving
toward the basket''---to distinguish a target from a similar competitor
semantically~\cite{li2017tracking,wang2021tnl2k}. This capability is most
valuable outside clean laboratory video: robots, autonomous cameras, and
other perception--action systems encounter image degradations under which even
subtle perturbations can cause severe perception failures
\cite{pezzementi2018safeperception,michaelis2019detection,kerim2021adverse}.

Accuracy with a target-valid description, however, does not establish that a
tracker uses language's semantic content to select the target over an
on-screen competitor. Prior robustness studies show that degraded visibility
harms trackers and VLMs~\cite{noman2022avist,hu2024dbvot,usama2025vlmcorruptions,saxena2026vlmrobustbench},
while VLT studies show that description content and delivery can alter
performance~\cite{li2024dtllm,li2024vltverse,alansari2025cldtracker}.
They generally vary visual quality and semantic validity separately, leaving
open whether language rescues weak vision, misdirects selection toward a
matching distractor, or is behaviorally redundant on the same clip.

\begin{figure*}[t]
\centering
\includegraphics[width=\textwidth]{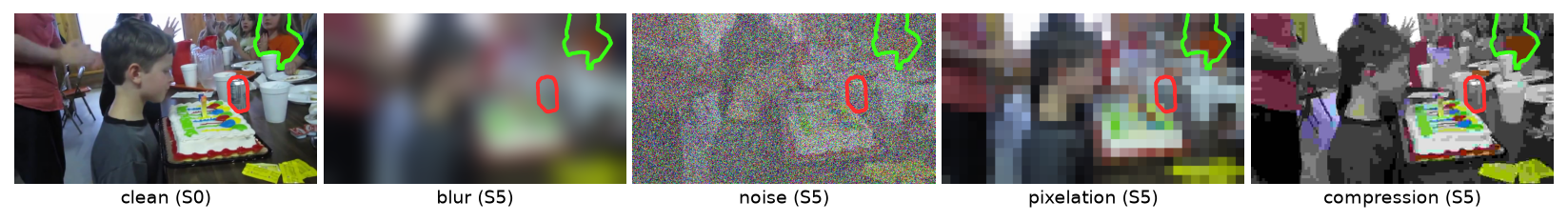}
\caption{A real MeViS target--distractor pair (green = target; red = distractor) at the clean initialization frame and the terminal severity ($S_5$) of blur, noise, pixelation, and compression. Descriptions: target, ``girl in red sitting on the right side of the table''; distractor, ``The can next to the cake''.}
\label{fig:intro_example}
\end{figure*}

We evaluate Molmo2~\cite{clark2026molmo2}, JointNLT~\cite{zhou2023jointnlt},
UVLTrack~\cite{ma2024uvltrack}, MMTrack~\cite{zheng2024mmtrack}, and
DUTrack~\cite{li2025dutrack} on four diverse datasets, crossing four language priors with progressive visual
degradation (Fig.~\ref{fig:intro_example}). UVLTrack, MMTrack, and DUTrack show only
small, inconsistent text-conditioned differences. Molmo2 and
JointNLT respond directionally on clean video, but their severity trajectories
are heterogeneous: some supported effects attenuate toward the null, 
whereas for others—including the single harmful clean-input effect—we detect no 
FDR-significant endpoint change.

However, behavior alone cannot reveal whether a description never reaches a tracking
pathway, remains represented but does not affect localization, or loses its
link to the spatial decision under degradation. We therefore complement the
benchmark with layerwise, source-video-grouped probes across all five systems
and targeted activation patching in the two responsive systems. 
This distinguishes linearly recoverable text-condition information from stages 
at which controlled state replacement changes the trajectory, 
while donor-content controls test whether that redirection is target-specific.

\textbf{Our contributions are:}
\begin{itemize}
    \item A matched four-condition, multi-severity VLT protocol spanning four perturbations, five systems, and four datasets, while preserving models' supported point- or box-tracking interfaces.
    \item A five-model degradation study showing that text effects impact VLTs in meaningfully different ways, depending on the model, dataset, and perturbation.
    \item Layerwise probing and controlled state-replacement analyses that trace text-condition signals across all five systems and identify intervention-sensitive stages in Molmo2 and JointNLT.
\end{itemize}

\section{Related Work}
\label{sec:related}

\paragraph{Evaluating language use in tracking.} Language-annotated benchmarks (OTB99-Lang~\cite{li2017tracking}, LaSOT~\cite{fan2019lasot}, TNL2K~\cite{wang2021tnl2k}, MGIT~\cite{hu2023mgit}, WebUAV-3M~\cite{zhang2023webuav}) and VLT methods (SNLT~\cite{feng2021snlt}, ModaMixer~\cite{guo2022modamixer}, JointNLT~\cite{zhou2023jointnlt}, MMTrack~\cite{zheng2024mmtrack}, UVLTrack~\cite{ma2024uvltrack}, DUTrack~\cite{li2025dutrack}) are evaluated by trajectory accuracy under valid descriptions -- accuracy given truthful language, not whether counterfactually changing that language changes target selection. Nor do the language diagnostics we are aware of isolate visibility: DTLLM-VLT and DTVLT vary detail and update frequency while keeping text target-consistent~\cite{li2024dtllm,li2024dtvlt}; VLTVerse crosses multiple text variants with naturally occurring sequence-level factors, rather than applying controlled degradation~\cite{li2024vltverse}; VLT-MI changes text and the box together~\cite{li2024vltmi}; and CLDTracker's substitutions are neither tied to an on-screen competitor nor tested across severity~\cite{alansari2025cldtracker}. Our protocol addresses both gaps by independently varying text relation and severity on the same clip with a designated distractor.

\paragraph{Robustness under degraded visual evidence.} Common-corruption evaluation applies graded transformations to a fixed source~\cite{hendrycks2019corruptions}, but tracking benchmarks instead use naturally difficult sequences (LaSOT~\cite{fan2019lasot}, TNL2K~\cite{wang2021tnl2k}, AVisT~\cite{noman2022avist}, adverse-weather person tracking~\cite{kerim2021adverse}) that confound visibility with target, motion, and scene difficulty. DBVOT evaluates six progressive defocus levels~\cite{hu2024dbvot}, while adversarial-blur work synthesizes motion-blur attacks against vision-only trackers~\cite{guo2021adversarialblur}, and VLM corruption suites retain a valid prompt while scoring one-shot answers~\cite{usama2025vlmcorruptions,saxena2026vlmrobustbench} -- neither directly measures changing text authority or temporal identity errors. Our graded blur, noise, pixelation, and compression experiments test whether language guides tracking as visual evidence weakens.

\paragraph{Cross-modal conflict and behavioral consequence.} CLASH and CrossCheck-Bench ask models to recognize controlled image--text contradictions, not to reveal their unprompted behavioral consequence in an ongoing spatial task~\cite{popordanoska2026clash,tian2026crosscheck}, and SIGNPOST-Bench shows that embedded scene text can bias still-image geolocation, but does not study a graded visual-degradation axis~\cite{li2026signpost}. By holding scene content fixed, identifying the competing object, and measuring the direction and duration of tracking errors, our benchmark tests whether language remains actionable under imperfect sensing.

\paragraph{Mechanistic interpretability of grounding models.} Causal tracing and activation patching have probed how general vision–language models use visual information, including BLIP~\cite{palit2023blipcausal} and object hallucination in large VLMs~\cite{li2025objcausaltrace}, but not, to our knowledge, vision--language tracking, where a spatial decision is emitted every frame rather than a single caption. We apply the same causal-tracing toolkit to five VLT systems to test whether text condition is decodable at decision-near representations and whether controlled interventions at selected text-reachable states alter localization.

\section{Benchmark and Protocol}
\label{sec:benchmark}

\subsection{Datasets and pair construction}
\label{sec:pair_construction}

We use source videos from MeViS~\cite{ding2023mevis}, a referring video
object segmentation benchmark, and from the multi-object tracking benchmarks
DanceTrack~\cite{sun2022dancetrack}, SportsMOT~\cite{cui2023sportsmot}, and
PersonPath22~\cite{shuai2022personpath} (Table~\ref{tab:datasets}).  MeViS
provides its native referring expressions and masks. For the three MOT
sources, human-written expressions and released point/mask trajectories come
from the public Molmo2-VideoTrackEval annotation release~\cite{clark2026molmo2};
the original MOT datasets provide the source videos. We derive initialization
boxes from the released masks and enumerate ordered target--distractor pairs.
A pair is eligible only if the target $T$ and distractor $D$ are co-visible in at least five frames
and have mean co-visible mask IoU below 0.3. Every selected pair
receives the same text conditions and perturbation protocol.

The MeViS selector takes one eligible pair per source video, while for the
other three datasets, we construct outcome-enriched diagnostic stress
subsets. Molmo2's clean conflicting-text run identifies cases for inclusion,
with seeded random filling of the remaining quota, yielding 16/60 DanceTrack,
2/30 SportsMOT, and 6/30 PersonPath22 persistent-takeover cases in the final
subsets.
These subsets let us track how conflict-relevant cases evolve under
degradation. Because pairs are
partly selected on an outcome related to what we measure, naive effect
estimates on this set risk running high; a supplementary
inverse-probability-weighted check estimates the corresponding effects over the full eligible candidate pool; effect directions are preserved, although some naïve stress-subset estimates—particularly \(C_T\)—are larger. A second, more direct check in the supplementary reruns all four specialized trackers on the complete, outcome-independent eligible pool, reproducing the qualitative separation between JointNLT and the three small-effect specialized trackers.

\begin{table}[t]
  \centering
  \small
  \setlength{\tabcolsep}{4pt}
  \begin{tabular}{@{}llrr@{}}
    \hline
    Dataset & Characterization & Pairs & Seq. \\
    \hline
    MeViS & referring seg., masks & 120 & 120 \\
    DanceTrack & dance MOT, similar look & 60 & 13 \\
    SportsMOT & athlete MOT, fast motion & 30 & 19 \\
    PersonPath22 & pedestrian MOT, crowded & 30 & 20 \\
    \hline
  \end{tabular}
  \caption{Per-dataset pair and source-video coverage. Pairs is the number
  of selected target--distractor pairs. Seq.\ is the number of distinct
  source videos selected.}
  \label{tab:datasets}
\end{table}

\subsection{Targets and distractors}

Let a video contain an ordered pair of distinct annotated objects $(T,D)$, where
$T$ is the object initialized for tracking and $D$ is a co-visible distractor.
At evaluated frame $t$, their annotated spatial supports are
$\mathcal{M}_{T,t}$ and $\mathcal{M}_{D,t}$. Each pair has a human-written
expression $e_T$ referring to $T$ and an expression $e_D$ referring to $D$.

\subsection{Text conditions}
\label{sec:text_conditions}

We vary the text input independently of the video through four conditions:
\begin{itemize}[leftmargin=1.1em,label=\textbullet,itemsep=0pt,topsep=2pt,parsep=0pt]
  \item \textbf{Neutral} ($c_{\mathrm{neutral}}$): a generic tracking
  instruction with no target-identifying description.
  \item \textbf{Unrelated} ($c_{\mathrm{unrelated}}$): a real expression
  $e_U$ from another clip, that is not annotated as referring to either $T$ nor $D$.
  \item \textbf{Correct} ($c_{\mathrm{correct}}$): the target expression
  $e_T$.
  \item \textbf{Conflicting} ($c_{\mathrm{conflicting}}$): the distractor
  expression $e_D$.
\end{itemize}
The unrelated expression is sampled once from a different annotated clip with
the same category label as the pair, then fixed across every model, condition,
and severity.

\subsection{Model initialization, inputs, and outputs}
\label{sec:model_protocol}
\label{sec:text_interfaces}

Every run is anchored to the same clean annotated target instance and uses
the same fixed post-initialization frames for every model, text condition,
and perturbation. Molmo2 is initialized with a
target point and predicts points; UVLTrack, MMTrack, JointNLT, and DUTrack are
initialized with a tight axis-aligned box and predict boxes, evaluating which
object is selected rather than boundary precision. Points/boxes are derived
from mask centroids/extents when masks are available, and otherwise from the
corresponding released box trajectory. We additionally report native-box and
absolute-performance metrics in the supplementary.

Molmo2 receives text through the scaffold ``Track the object indicated by the
point,'' with descriptive conditions appending ``The target is: $e$.''
UVLTrack, MMTrack, and JointNLT receive the raw expression $e$ directly, with
the generic token \texttt{object} as their neutral-condition input. DUTrack
shares this interface but normally regenerates a description from the current
frame, overwriting the assigned condition; we use \emph{static-external}
mode and reset all adapter state between pair--condition--severity runs. A
supplementary dynamic-mode audit verifies this condition-preservation control.

\subsection{Perturbations and fixed severity schedule}
\label{sec:perturbations}

Each corruption is applied independently to every post-initialization frame;
the initialization frame remains clean and corruptions are never composed. We use Gaussian blur, additive Gaussian noise, pixelation, and JPEG
compression---common corruption families with fixed, ordered severity ladders
in robustness benchmarks~\cite{hendrycks2019corruptions}. Table~\ref{tab:severity}
gives our disclosed, perturbation-specific parameter schedules; these were
chosen to produce visibly distinct changes at the native video resolution.
$S0$ is an exact no-op. Severity is an ordinal dose axis within each
perturbation, not a claim of perceptual equivalence across perturbation types;
the supplementary reports descriptive PSNR/SSIM diagnostics. For noise, a
video--frame hash fixes one standard-normal field shared across conditions and
severities, so contrasts never compare different noise realizations.

\begin{table}[!h]
  \centering
  \small
  \setlength{\tabcolsep}{4pt}
  \begin{tabular}{@{}ll@{}}
    \hline
    Perturbation & Fixed schedule values, $S0$--$S5$ \\
    \hline
    Blur, $k_s$ & $0,\ .004,\ .008,\ .016,\ .032,\ .064$ \\
    Noise, $\sigma_s$ & $0,\ .12,\ .20,\ .30,\ .45,\ 1.5$ \\
    Pixelation, $r_s$ & $1,\ .35,\ .25,\ .18,\ .12,\ .04$ \\
    Compression, $q_s$ & no-op$,\ 15,\ 10,\ 7,\ 3,\ 1$ \\
    \hline
  \end{tabular}
  \caption{Fixed severity schedule.}
  \label{tab:severity}
\end{table}

\subsection{Outcomes and metrics}
\label{sec:metrics}

Each post-initialization evaluated sample receives exactly one of the following labels:
\begin{itemize}[leftmargin=1.1em,label=\textbullet,itemsep=0pt,topsep=2pt,parsep=0pt]
  \item \textbf{absent} (no usable predicted point at that timestamp).
  \item \textbf{target} ($\hat p_t\in\mathcal{M}_{T,t}$).
  \item \textbf{distractor} ($\hat p_t\notin\mathcal{M}_{T,t}$ and $\hat p_t\in\mathcal{M}_{D,t}$).
  \item \textbf{other}.
\end{itemize}
Target takes precedence
when masks overlap, and absent samples remain in the denominator of all
frame-level rates.

For pair $p$, condition $c$, and severity $s$, let
$A_T^{p}(c,s)$, $A_D^{p}(c,s)$, $A_O^{p}(c,s)$, and
$A_{\emptyset}^{p}(c,s)$ be the fractions of post-initialization samples classified as
target, distractor, other, and absent. They sum to one. We aggregate frames
within a pair before aggregating pairs, preventing longer tracks from receiving
greater weight. In addition to condition-specific rates, the primary
contrastive semantic effects anchor every comparison at the \emph{correct}
condition -- the observed target accuracy when the model receives the ground-truth referring expression
 -- and measure the gap to each of the
other three conditions:
\begin{align}
 B_T(s) &= A_T(\mathrm{correct},s)-A_T(\mathrm{unrelated},s),\label{eq:bt}\\
 N_T(s) &= A_T(\mathrm{correct},s)-A_T(\mathrm{neutral},s),\label{eq:nt}\\
 C_T(s) &= A_T(\mathrm{correct},s)-A_T(\mathrm{conflicting},s).\label{eq:ct}
\end{align}
$B_T$ is the correct-versus-unrelated effect (a same-type but unrelated
description), $N_T$ is the correct-versus-neutral effect (no
target-identifying description), and $C_T$ is the correct-versus-conflicting
effect; each is signed and can be
negative. Two further effects, the directed distractor-capture effect $S_D$
and the conflict-specific target-loss effect $L_T$, are defined in the
supplementary alongside the full endpoint test family.

\paragraph{Severity-integrated effect magnitude.} To summarize the size of a
condition effect across the severity ramp, we use integrated absolute
deviation (IAD). Severities are indexed $s\in\{0,\dots,5\}$ (
Table~\ref{tab:severity}); for conditions $c_1,c_2$, let
$\Delta_{c_1,c_2}(s)=A_T(c_1,s)-A_T(c_2,s)$ and define
\begin{equation}
 \mathrm{IAD}(c_1,c_2) = \sum_{s=0}^{4}\tfrac12\big[|\Delta_{c_1,c_2}(s)|+|\Delta_{c_1,c_2}(s{+}1)|\big].
 \label{eq:iad}
\end{equation}
Taking the absolute value \emph{before} integrating preserves effects that
change direction over severity. IAD is nonnegative and symmetric in its two
conditions; with six severity levels, $\mathrm{IAD}/5$ is the mean absolute
condition gap per severity step. We report all $\binom{4}{2}=6$ unordered
condition contrasts as the upper triangle of a $4\times4$ matrix per model
and dataset. Per-severity signed contrasts ($B_T$, $N_T$, and $C_T$) remain
the primary directional measurements and are used for endpoint tests.

\paragraph{Behavioral inference.} We compute every contrast within pair and
use source video as the resampling unit, keeping all selected pairs from one
video together. We report 95\% source-video-clustered bootstrap CIs and use
paired source-video sign-flip tests for clean effects and severity-endpoint
changes. Benjamini--Hochberg correction is applied within each stated
hypothesis family; IAD is descriptive rather than inferential. The
supplementary specifies the resampling algorithm, test statistics,
hypothesis families, and formulas in full.

\subsection{Model-internal and mechanistic measurements}
\label{sec:internal_metrics}

For internal analyses, we save compact pooled states at model-specific
functional stages---text encoding/projection, early and late multimodal
fusion, and the decision-near localization state---rather than treating raw
layer indices as comparable across architectures. Exact hooked tensors,
layer counts, and semantics per model are listed in the supplementary.

At a matched video frame, let $h_l^{c}$ be a pooled state at stage $l$ under
text condition $c$. Let $m_t$ be the target--distractor output margin at
evaluated frame $t$,
\begin{equation}
 m_t=\mathrm{IoU}(\hat b_t,b_{T,t})-\mathrm{IoU}(\hat b_t,b_{D,t}),
 \label{eq:margin}
\end{equation}
with the analogous target/distractor label used for point output. We use
three procedures to characterize the text pathway before defining the
metrics computed from them.

\paragraph{Text-condition probe.} Following low-capacity probing practice
\cite{hewitt2019control,belinkov2022probing}, at every saved stage we
standardize the pooled state using statistics fit only on the training fold,
project it onto a training-fold SVD basis (at most 20 components), and fit a
ridge-regularized linear classifier for correct-versus-conflicting condition.
Cross-validation holds out complete source videos, so the probe cannot exploit
sentence- or token-identity shortcuts shared within a video; group-preserving
label permutations provide its null distribution. We report the probe's
held-out balanced accuracy as \textbf{Text BA} at stage $l$ ($0.5$ is chance).
Within conflicting runs, we repeat the same procedure for the observed
target-versus-distractor outcome. This establishes whether the prompt
distinction is linearly present in a state, not whether the tracker acts on it;
causal relevance is tested with the interventions below~\cite{elazar2021amnesic}.

\paragraph{Activation patching.} We intervene on the forward pass directly, following intervention-based interpretability, causal-tracing, and activation-patching
work~\cite{vig2020causal,meng2022rome,zhang2024activation}. For a fixed video,
template, and evaluated-frame sequence, we run the network under a
\emph{target} condition (typically conflicting) but replace its unpooled
native state at the identified text boundary with the corresponding state
from an independently run \emph{source} condition
($\mathrm{source}\in\{\mathrm{correct},\mathrm{neutral},\mathrm{unrelated},\mathrm{zero}\}$)
on the same visual input,
\begin{equation}
 h_l^{\mathrm{target}} \leftarrow h_l^{\mathrm{source}},
 \label{eq:patch}
\end{equation}
before continuing the forward pass and subsequent sequential tracking
unmodified. Every patch site is validated with two controls: an exact
self-patch ($\mathrm{source}=\mathrm{target}$), required to reproduce the
unpatched trajectory exactly, and an unpatched repeat, which confirms that
reset and replay are themselves deterministic.

For each activation-patching intervention, we report the paired change in trajectory-mean $m_t$ and the count of
target--distractor switches, with source-video bootstrap CIs and paired
sign-flip tests; these controls establish patch integrity and deterministic replay. 
They support interpreting differences as consequences of the intervention, 
but do not by themselves establish target-specific semantic mediation; 
that stronger interpretation requires donor-content controls.

\paragraph{Full-depth layer sweeps.} Rather than committing in advance to one
decision-near stage, we apply the text-condition probe at every saved
functional stage for every model, (for Molmo2, at all 37 pre-coordinate
decoder depths). This traces the depth at which the
correct/conflicting distinction becomes linearly decodable, from text
encoding through fusion to the localization output, reducing the risk that a null result at one selected stage is incorrectly generalized to the model’s entire saved pathway.

\begin{table*}[!t]
  \centering
  \scriptsize
  \setlength{\tabcolsep}{3pt}
  \resizebox{\textwidth}{!}{%
  \begin{tabular}{@{}lccccccccccccccc@{}}
    \toprule
     & \multicolumn{3}{c}{Molmo2} & \multicolumn{3}{c}{JointNLT} & \multicolumn{3}{c}{UVLTrack} & \multicolumn{3}{c}{MMTrack} & \multicolumn{3}{c}{DUTrack} \\
    \cmidrule(lr){2-4} \cmidrule(lr){5-7} \cmidrule(lr){8-10} \cmidrule(lr){11-13} \cmidrule(lr){14-16}
    Dataset & $C_T$ & $B_T$ & $N_T$ & $C_T$ & $B_T$ & $N_T$ & $C_T$ & $B_T$ & $N_T$ & $C_T$ & $B_T$ & $N_T$ & $C_T$ & $B_T$ & $N_T$ \\
    \midrule
    DanceTrack   & $\underline{+25\%}$ & $\underline{+13\%}$ & $\underline{+8\%}$  & $\underline{+11\%}$ & $\underline{+11\%}$ & $+6\%$           & $+1\%$ & $-1\%$          & $+2\%$ & $-1\%$ & $+0\%$ & $+0\%$          & $+0\%$ & $+1\%$ & $+0\%$ \\
    SportsMOT    & $+6\%$           & $+0\%$           & $+4\%$           & $-2\%$           & $\underline{-11\%}$ & $-6\%$           & $-1\%$ & $+1\%$          & $-1\%$ & $+1\%$ & $-2\%$ & $-4\%$          & $-1\%$ & $-2\%$ & $+0\%$ \\
    PersonPath22 & $+14\%$          & $+1\%$           & $+4\%$           & $\underline{+27\%}$ & $\underline{+18\%}$ & $\underline{+17\%}$ & $+5\%$ & $\underline{+5\%}$ & $+5\%$ & $+2\%$ & $+1\%$ & $\underline{+3\%}$ & $+0\%$ & $-2\%$ & $\underline{+0\%}^{\dagger}$ \\
    MeViS        & $\underline{+8\%}$  & $\underline{+4\%}$  & $+2\%$           & $\underline{+14\%}$ & $\underline{+11\%}$ & $\underline{+8\%}$  & $+2\%$ & $+1\%$          & $-1\%$ & $+0\%$ & $+0\%$ & $+1\%$          & $+0\%$ & $+0\%$ & $+0\%$ \\
    \bottomrule
  \end{tabular}
  }
  \caption{Clean-input ($S_0$) text effects for all five systems
  on all four datasets. Underlined: 95\% source-video
  cluster-bootstrap CI excludes zero. Among UVLTrack/MMTrack/DUTrack: 0/12 $C_T$, 1/12 $B_T$, 2/12 $N_T$ CIs
  exclude zero; the sole $B_T$ case (UVLTrack/PersonPath22) fails the
  clustered sign-flip test ($p=.08$, BH $q=.668$), so is not treated as a
  reproducible effect.
  \emph{$^{\dagger}$} Near-zero-variance bootstrap artifact, not interpreted.}
  \label{tab:s0_baseline}
\end{table*}

\section{Results}
\label{sec:results}

\begin{figure}[htbp]
  \centering
  \includegraphics[width=\linewidth]{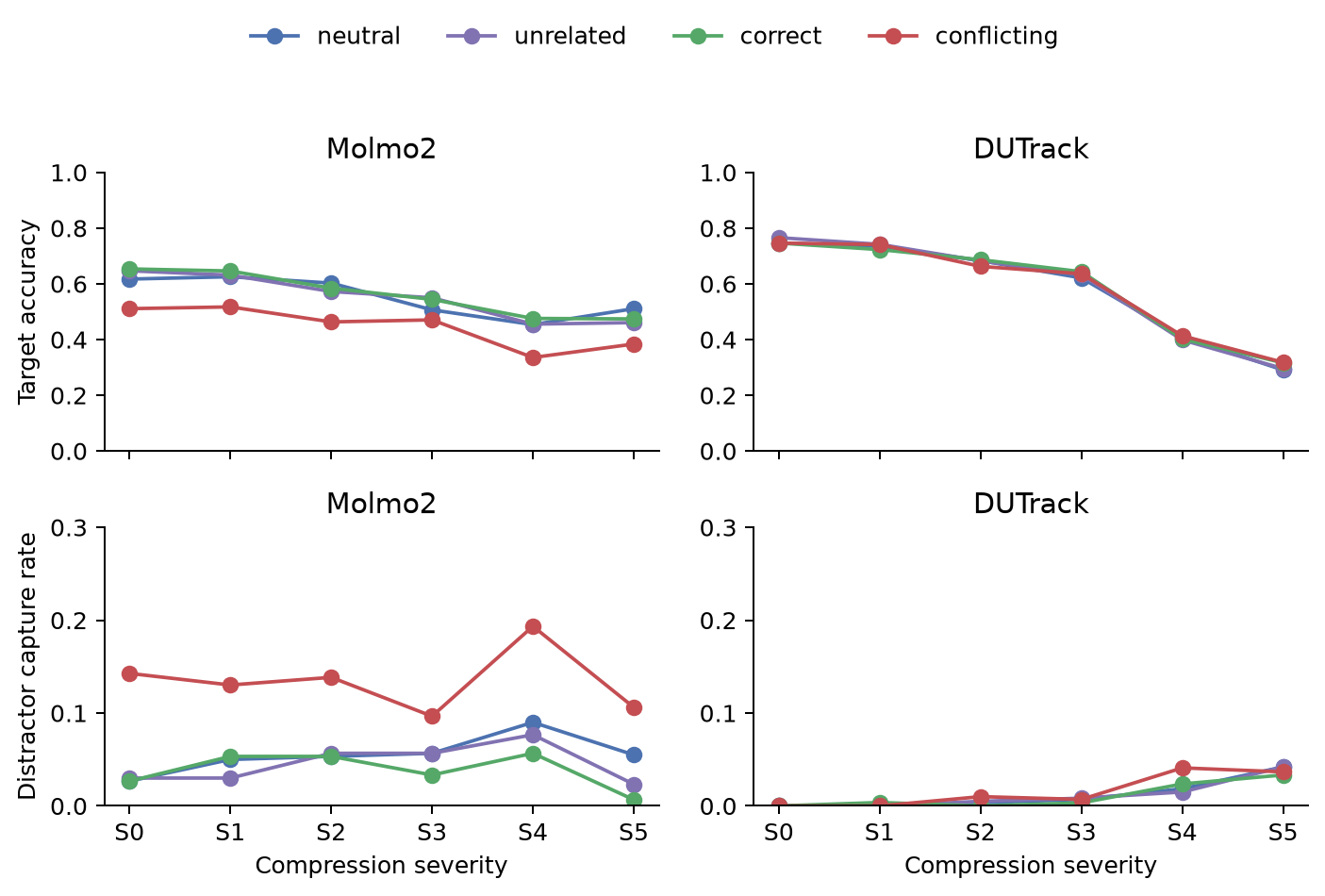}
  \caption{Representative compression trajectories on PersonPath22.
  Molmo2 (left) separates text conditions in target accuracy and
  distractor capture; DUTrack (right) remains nearly unchanged.}
  \label{fig:qual_example}
\end{figure}

Fig.~\ref{fig:qual_example} illustrates the behavioral contrast that the
aggregate analyses quantify: Molmo2's target
accuracy and distractor capture diverge across text conditions, whereas DUTrack’s trajectories closely overlap. 
We use the full matched grid below to assess the 
breadth and uncertainty of this qualitative pattern across the tested settings.

\begin{figure}[htbp]
  \centering
  \includegraphics[width=0.97\columnwidth]{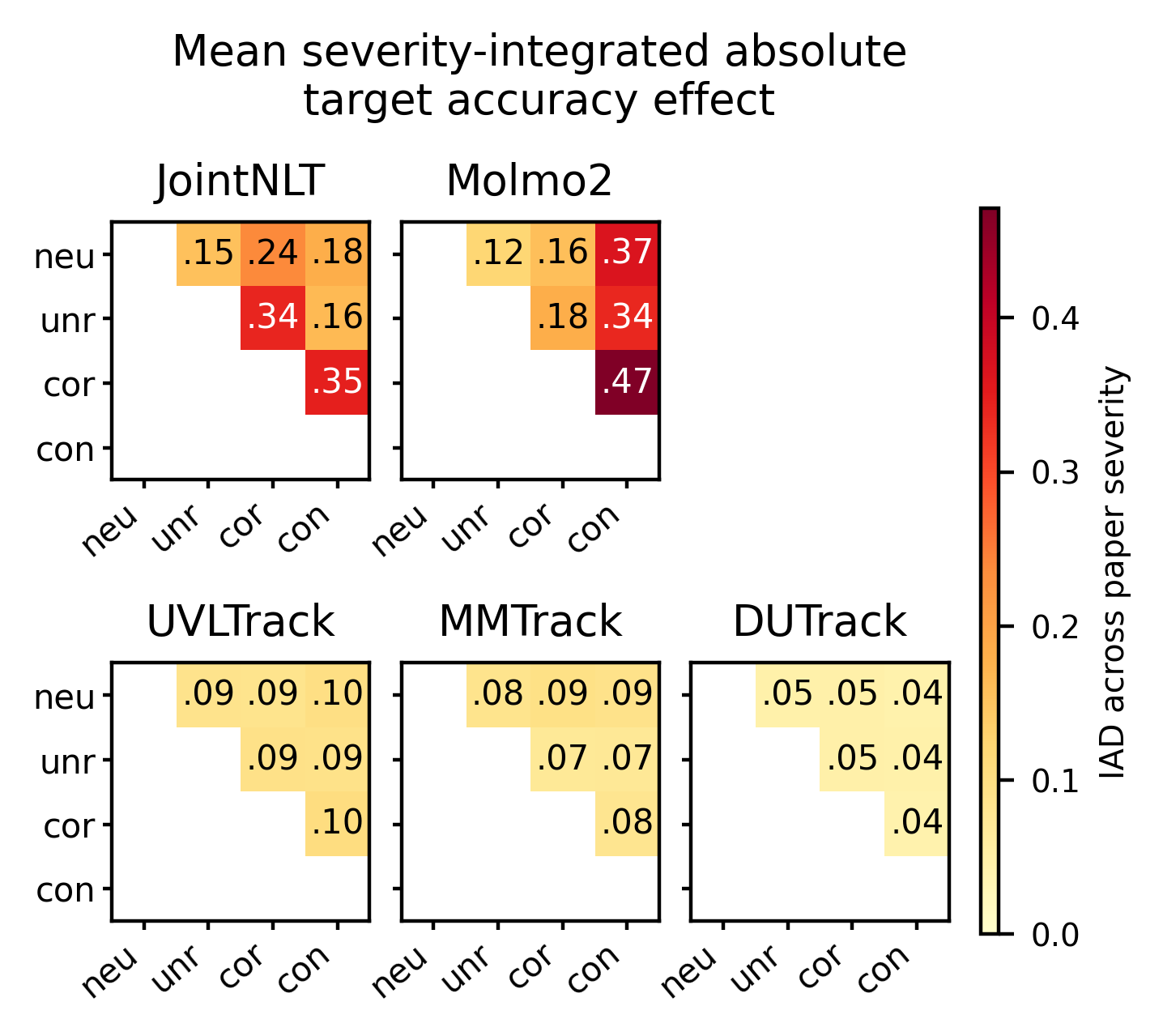}
  \caption{Mean severity-integrated absolute target-accuracy effect
  ($\mathrm{IAD}(c_1,c_2)$, Eq.~\ref{eq:iad}) across all four perturbations
  and four datasets, per model. Top row: JointNLT, Molmo2. Bottom row:
  UVLTrack, MMTrack, DUTrack, same color scale. Condition labels abbreviate
  neutral/unrelated/correct/conflicting.}
  \label{fig:iad_pooled}
\end{figure}

\subsection{UVLTrack, MMTrack, and DUTrack show limited behavioral sensitivity to text semantics}
\label{sec:null-models}

Fig.~\ref{fig:iad_pooled} plots each system's severity-integrated
absolute target-accuracy effect as a $4\times4$ grid over the four text
conditions, one grid per model on a shared color scale, where larger,
warmer cells mark condition pairs whose accuracy diverges more across the
full clean-to-severe range as text effects differ. Two grids show consistently larger effects; 
the remaining three stay pale in every cell. This qualitative
split holds numerically: the three less-responsive systems' largest target-accuracy IAD
cells are $0.05$ (DUTrack), $0.09$ (MMTrack), and $0.10$ (UVLTrack),
versus JointNLT's mean $0.24$ (maximum $0.35$) and Molmo2's mean $0.27$
(maximum $0.47$), both peaking at the correct--conflicting contrast. 
These observed separations remain small after taking absolute values before integration, 
so their small IAD values are not caused by cancellation between opposite-signed severities. 
The corresponding distractor-capture IAD is also small
($0.02$/$0.05$/$0.08$ for DUTrack/MMTrack/UVLTrack vs.\ $0.29$/$0.37$ for
JointNLT/Molmo2), again largest at the correct--conflicting contrast,
consistent with conflict-directed capture rather than a uniform
output-quality change.

Thus, we detect no FDR-supported directed semantic-text effect in these three systems, without implying absent internal language processing yet. Their 62--68\% correct-text \(S_0\) target accuracy argues against a simple floor/ceiling explanation and leaves room for change in either direction.

\subsection{Directed language effects degrade unevenly, including harmful effects}
\label{sec:severity-ramp}

The deployment-relevant question for the two responsive systems is not
merely whether accuracy falls with degradation, but whether a
CI-supported clean-input text effect reliably weakens as visual evidence degrades. We
test every effect with a CI-supported $S_0$ signal (95\% CI excludes zero) for
FDR-significant contraction toward zero across all four main perturbations:
14 target-accuracy cases and 4 distractor-capture cases, spanning both
responsive systems and all four datasets (Table~\ref{tab:contraction_heterogeneity}).

\begin{table}[htbp]
  \centering
  \scriptsize
  \setlength{\tabcolsep}{4pt}
  \begin{tabular}{@{}lr@{}}
    \hline
    Contraction level (of 4 perturbation ramps) & Cases (of 14) \\
    \hline
    4/4    & 1 \\
    1--3/4 & 3 \\
    0/4    & 10 \\
    \hline
  \end{tabular}
  \caption{FDR-significant contraction toward zero, across the 14
  target-accuracy cases with a CI-supported clean-input signal.}
  \label{tab:contraction_heterogeneity}
\end{table}

Of the 14 target-accuracy effects with a CI-supported clean-input value, four show FDR-significant contraction under at least one perturbation, including one that contracts under all four; the remaining ten show no FDR-significant endpoint change under any perturbation. None of Molmo2's five cases yields a significant
contraction after correction. These nonsignificant change tests should not
be interpreted as evidence that the effects remain unchanged.

One of the ten cases is JointNLT's SportsMOT \(B_T\), the study's only
significantly harmful clean-input effect (\(-11.2\%\), 95\% CI excluding zero). Under the primary point-in-support metric, its estimated
S5 changes substantially reduce or reverse the clean effect, but these
changes do not survive FDR correction.

The distractor-capture effect \(S_D\) exhibits a similar inferential
pattern. Of four model--dataset combinations with a CI-supported
clean-input misdirection effect, JointNLT/MeViS shows significant
contraction under three of four perturbations. For Molmo2/PersonPath22,
the largest clean capture effect in the study (\(+11.3\) pp), no endpoint
change survives correction; nevertheless, the point estimates contract
almost completely under blur, noise, and pixelation and remain more
clearly positive under compression.

Thus, contraction is established in some settings, while the remaining changes are statistically inconclusive rather than evidence of persistence. Their behavior varies across models, datasets, effects, metrics, and perturbations.

\subsection{Model internals}
\label{sec:internals}

The behavioral results identify FDR-supported directed text effects for JointNLT and Molmo2, in contrast to the demonstrably smaller effects observed for MMTrack, DUTrack, and UVLTrack. They do not, however, localize \emph{where} the supported effects arise. We probe this with the
internal-mechanism captures of
Sec.~\ref{sec:internal_metrics}, restricted to MeViS: its pairs are the only
ones with one independent source video each (Table~\ref{tab:datasets}); its mask-native annotations pool a hooked
state over exactly the target/distractor's tokens, not a
background-containing box; and its native task is itself referring
expression segmentation.

Of the 120 one-pair-per-video MeViS entries, 115 have at least two co-visible
target/distractor frames. JointNLT's
causal study further uses an outcome-enriched
90-video subset (23 persistent captures, 67 retention controls) so the
rarer capture outcome is well represented.

\subsubsection{Decision-near text decodability aligns with the behavioral split}
\label{sec:internals_overview}

For every system we probe Text BA (Sec.~\ref{sec:internal_metrics}) at its
most decision-near hooked stage. Table~\ref{tab:internals_text_ba} reports
the FDR-corrected result.

\begin{table}[htbp]
  \centering
  \scriptsize
  \setlength{\tabcolsep}{3pt}
  \resizebox{\linewidth}{!}{%
  \begin{tabular}{@{}llrrr@{}}
    \hline
    Model & Decision-near stage & Text BA & 95\% CI & $q$ \\
    \hline
    Molmo2   & final coord.\ query (L36) & \underline{$0.574$} & [$0.543$,$0.605$] & $0.015$ \\
    JointNLT & late-fusion $T{-}D$       & \underline{$0.695$} & [$0.647$,$0.744$] & $0.002$ \\
    \hline
    UVLTrack & pre-box-head search & $0.530$ & [$0.505$,$0.554$] & $0.174$ \\
    MMTrack  & coord.\ decoder input & $0.488$ & [$0.469$,$0.509$] & $0.898$ \\
    DUTrack  & pre-CENTER-head visual & $0.504$ & [$0.499$,$0.512$] & $0.506$ \\
    \hline
  \end{tabular}%
  }
  \caption{Correct-vs-conflicting Text BA ($0.5$=chance) at each system's
  decision-near hooked stage (115-video MeViS manifest). Underlined:
  FDR-significant ($q<.05$).}
  \label{tab:internals_text_ba}
\end{table}

Only Molmo2 and JointNLT carry an FDR-robust text-condition signal to the
selected decision-near state: MMTrack and DUTrack are near chance ($95\%$
CIs spanning $0.5$); UVLTrack's CI ($[0.505,0.554]$) is nominally above
chance but neither FDR-significant nor precise enough for equivalence --
converging with Sec.~\ref{sec:null-models}'s behavioral result that these
same three systems carry no reliable signal at their decision-near
states.

\begin{figure}[htbp]
  \centering
  \includegraphics[width=\columnwidth]{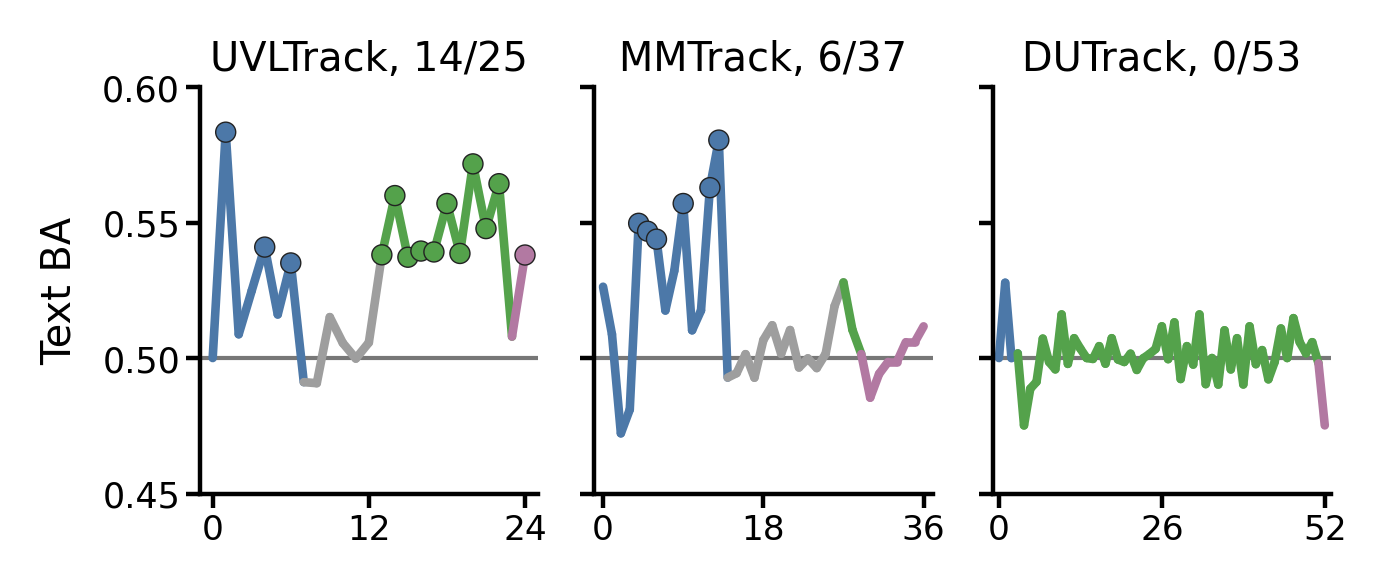}
  \caption{Exhaustive correct-vs-conflicting Text BA sweeps for
  the three behaviorally small-effect systems. Colors denote functional
  stage: text encoder/projection (blue), pre-fusion vision (gray),
  multimodal fusion (green), and localization/pre-box-head (purple).
  Filled circles mark FDR-significant states; titles report the number of FDR-surviving states and the total evaluated.}
  \label{fig:null_layerwise_text_ba}
\end{figure}

Fig.~\ref{fig:null_layerwise_text_ba} sweeps every text-reachable state
in these three systems, rather than only the decision-near endpoint in
Table~\ref{tab:internals_text_ba}.

The full sweeps show different routes to a decision-near effect that is not FDR-supported. 
UVLTrack is significant at 14/25 states, spanning its text encoder and fusion stages,
but not at the pre-localization endpoint. MMTrack is significant at 6/37
states, all in its text encoder/projection and none at or beyond fusion.
DUTrack has no FDR-significant state in its 53-state sweep (33 are formally
chance-equivalent).

\subsubsection{Layerwise text-condition decodability follows distinct routes in the two responsive systems}
\label{sec:internals_propagation}

\begin{figure}[htbp]
  \centering
  \includegraphics[width=\columnwidth]{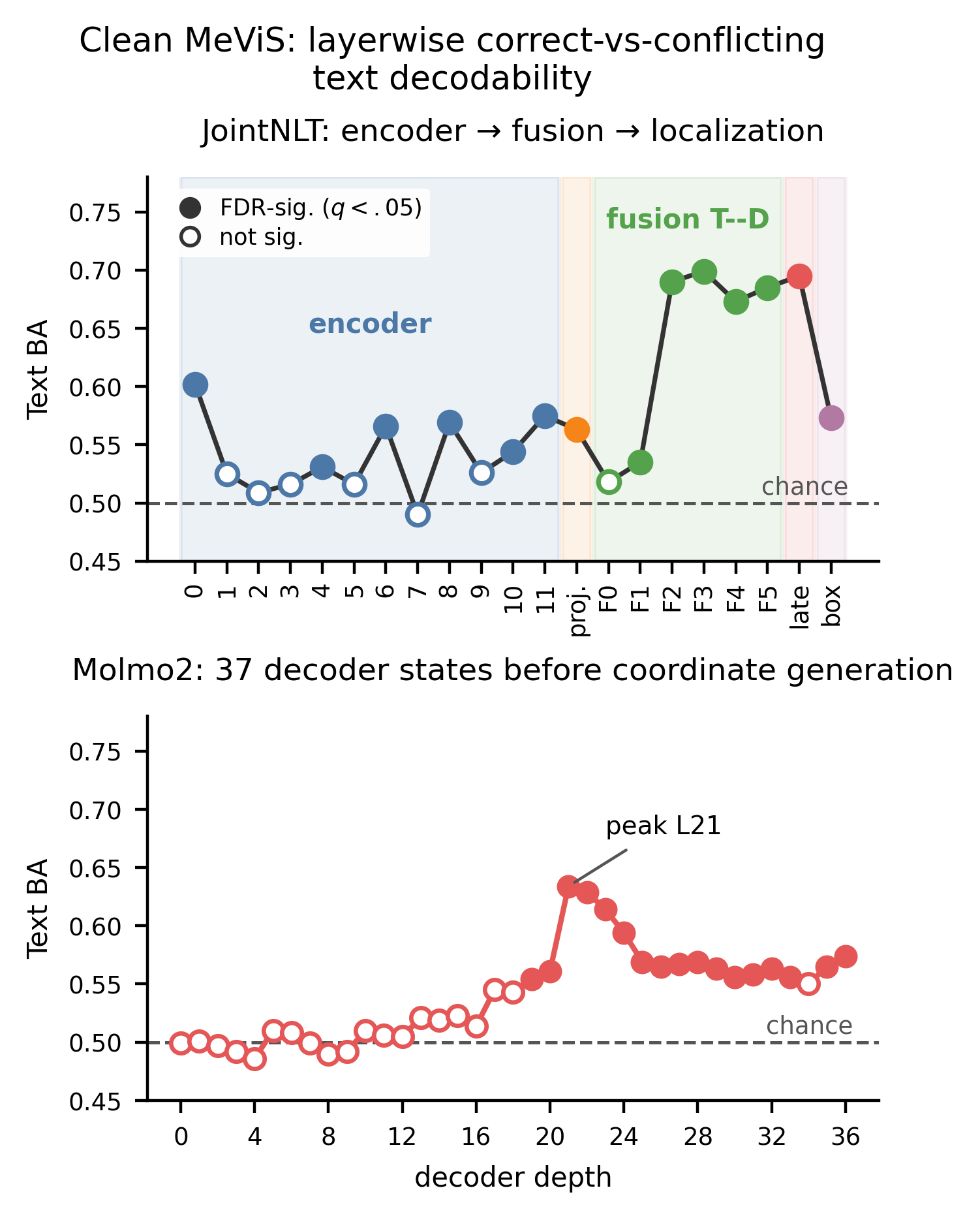}
  \caption{Exhaustive correct-vs-conflicting Text BA sweeps for
  Molmo2 and JointNLT. Solid trajectory segments indicate the
  functional state at their left endpoint: text encoder/projection (blue),
  pre-fusion vision (gray), multimodal fusion (green), and
  localization/pre-box-head (purple).}
  \label{fig:layerwise_text_ba}
\end{figure}

Fig.~\ref{fig:layerwise_text_ba} sweeps correct-vs-conflicting Text BA
through each Molmo2 and JointNLT's functional pathway.

JointNLT carries text-condition information through its encoder--fusion
pathway, though the early encoder signal is intermittent. The fusion pathway starts
weak, then turns consistently strong from its second block onward, staying
robust at the selected late-fusion state ($0.695$, $q=.002$) -- text is
linearly recoverable near localization, not uniformly strong throughout.
Molmo2 shows a different pattern: no decoder state through layer~16 is FDR-significant; decodability becomes significant at layer~19, peaks at layer~21 (\(0.634\)), and decreases but remains significant (except one) through the final state. This onset is consistent with its token order: image tokens precede the appended description, so condition information can emerge only after the description is processed. Thus, both systems carry decodable text to decision-near states through distinct pathways; intervention effects are tested next.

\subsubsection{State-replacement interventions redirect JointNLT and Molmo2 trajectories}
\label{sec:internals_patching}

\begin{figure}[htbp]
  \centering
  \includegraphics[width=\columnwidth]{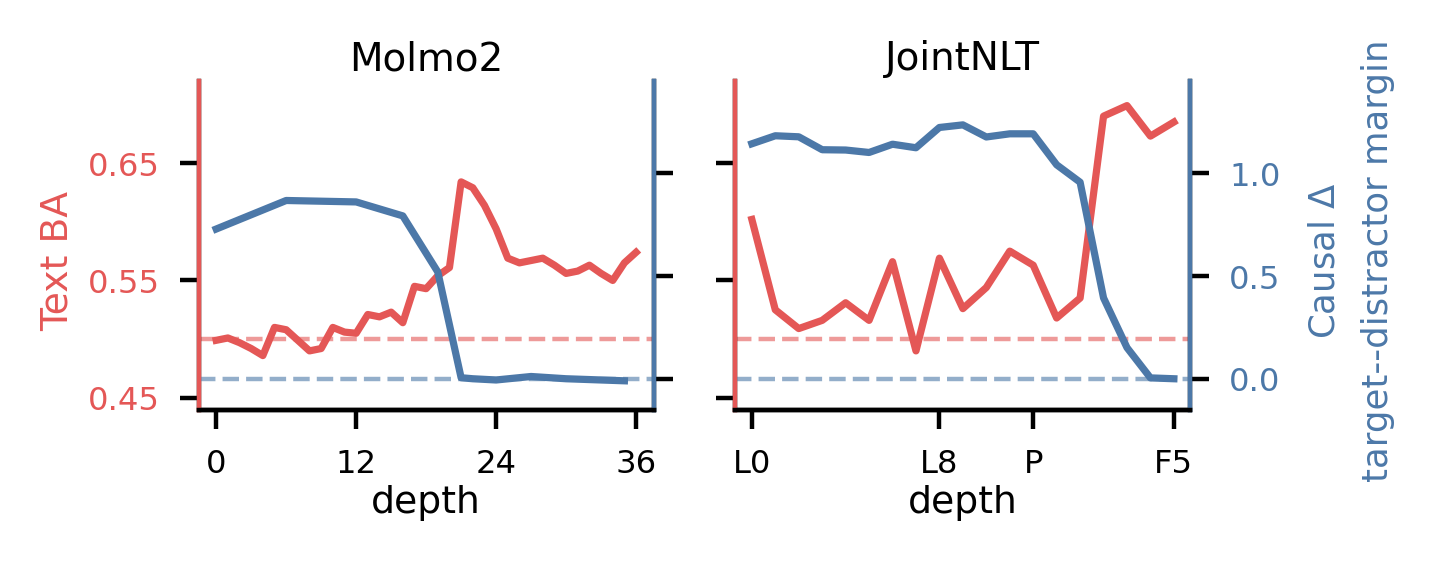}
  \caption{Text-condition decodability and matched-correct state-replacement
  effect across depth. Red, left ordinate: held-out Text BA. Blue, right
  ordinate: mean paired change in target--distractor margin after replacing a
  conflicting-run state with its same-video correct-run state. Both evaluated on the persistent-capture stratum. Dashed horizontal lines mark chance Text
  BA and zero replacement effect.}
  \label{fig:responsive_causal_dissociation}
\end{figure}

Fig.~\ref{fig:responsive_causal_dissociation} contrasts the full Text-BA
trajectory with matched-correct state replacement at each tested depth.

Both systems dissociate linear decodability from intervention efficacy:
replacement produces larger observed trajectory changes at selected earlier sites and 
smaller changes at each system’s Text-BA peak, so
the most linearly decodable state is not necessarily the most effective
intervention point.

\begin{figure}[htbp]
  \centering
  \includegraphics[width=\columnwidth]{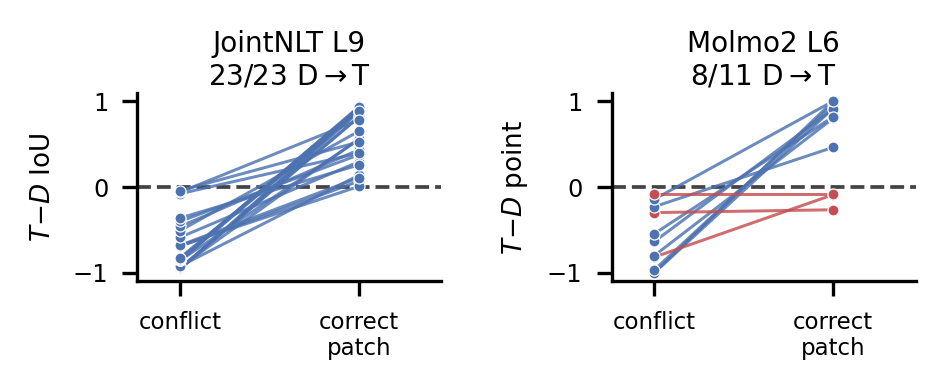}
  \caption{Matched-correct state replacement redirects conflict-associated trajectories in both responsive systems, 
  each shown at its best-recovery site; sites maximize the observed capture-stratum effect on the displayed data. Each line connects the unpatched conflicting run with its
  correct-text-state patch; blue: distractor-to-target transition, red: none.}
  \label{fig:jointnlt_molmo_patch_paired}
\end{figure}

Fig.~\ref{fig:jointnlt_molmo_patch_paired} shows this recovery per
persistent-distractor-capture video, at each model's best-recovery site.

JointNLT's BERT-layer-9 patch reverses all 23/23 persistent captures from
distractor- to target-preferred, and Molmo2's layer-6 patch does so in 8 of 11. On this stratum, however, neither system's correct-donor
effect significantly exceeds an unrelated donor's once corrected (JointNLT
$q=.124$, Molmo2 $q=.092$). Both correct and unrelated donors can redirect the
conflict-associated trajectory, so the grounding-to-output association
interventions do not establish target-specific semantic content as the 
cause of the redirection.
\FloatBarrier

\subsubsection{Under degradation, JointNLT retains fused text information while grounding-to-output association becomes inconsistent}
\label{sec:internals_degradation}

\begin{figure}[htbp]
  \centering
  \includegraphics[width=\columnwidth]{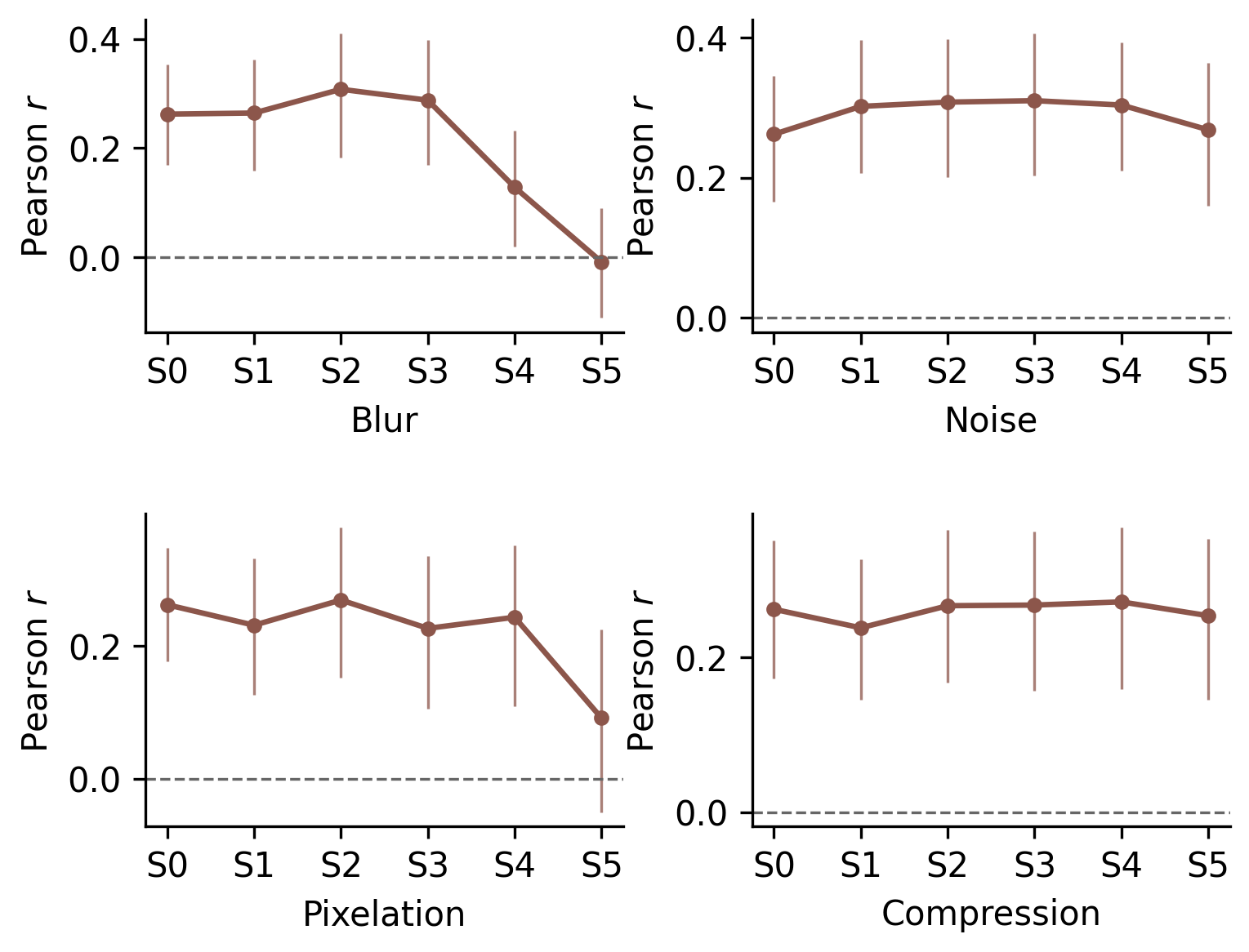}
  \caption{JointNLT's pre-localization state–output association (Pearson
  $r$) across $S_0$--$S_5$; dashed line: zero.}
  \label{fig:jointnlt_degradation_grounding}
\end{figure}

Sec.~\ref{sec:severity-ramp} showed that JointNLT's behavioral text effects
sometimes contract under degradation, although late-fusion Text BA
(Fig.~\ref{fig:layerwise_text_ba}) remains above chance. The
grounding-to-output association asks whether
the pre-localization target--distractor margin predicts the final output margin.
Figure~\ref{fig:jointnlt_degradation_grounding} traces this association, and
Table~\ref{tab:jointnlt_degradation_terminal_main} contrasts its $S_5$ value
with late visual separation and Text BA.

\begin{table}[htbp]
  \centering
  \scriptsize
  \setlength{\tabcolsep}{3pt}
  \begin{tabular}{@{}lrrr@{}}
    \hline
    Perturbation & late vis.\ sep.\ & late fused-text BA & output assoc.\ $r$ \\
    \hline
    blur        & $.173$ & $.582$ & $-.009$ [$-.111,.089$] \\
    noise       & $.042$ & $.566$ & $\underline{+.268}$ [$ .160,.363$] \\
    pixelation  & $.148$ & $.539$ & $+.091$ [$-.051,.225$] \\
    compression & $.253$ & $.577$ & $\underline{+.254}$ [$ .145,.353$] \\
    \hline
  \end{tabular}
  \caption{Terminal $S_5$ JointNLT summaries. Underlined: CI excludes zero.}
  \label{tab:jointnlt_degradation_terminal_main}
\end{table}

This is not a uniform collapse (Fig.~\ref{fig:jointnlt_degradation_grounding},
Table~\ref{tab:jointnlt_degradation_terminal_main}): blur and pixelation
fall to a CI spanning zero by $S_5$ while fused-text decodability remains nominally
$4$--$8$ points above chance throughout; noise and compression retain a
CI-supported positive association. Nor does it track visual degradation
-- compression has the largest terminal visual separation ($.253$) yet
one of the two CI-supported positive state–output associations. Thus, the association does not vary monotonically with this visual-separation summary, arguing against a simple one-to-one account while not ruling out visual mediation.

\subsection{Limitations}
\label{sec:limitations}

DISTRACT uses controlled comparative evidence across five systems, four
datasets, four perturbations, and fixed $S_0$--$S_5$ schedules; native
interfaces make cross-system differences whole-system rather than
architectural, and three datasets are stratified toward conflict-relevant
pairs, so raw error frequencies are not prevalence estimates. Controlled
perturbations and four priors isolate image quality and directed
target--distractor semantics, respectively, but not natural adverse conditions
or relational, temporal, multi-turn, and pragmatic language. They also do not
reproduce the coupled weather, scene, and sensor shifts of deployment.
Model-internal clean sweeps use 115 MeViS videos, and degraded internals are observational and
JointNLT-only; state replacement localizes sensitivity, not target-specific
semantic mediation. Broader models, datasets, and held-out causal tests are
therefore needed to further validate these claims.

\section{Discussion and Future Work}
\label{sec:discussion}

DISTRACT treats text as an experimental variable, testing whether counterfactual descriptions redirect tracking decisions and how these effects change under degraded vision. Three of five systems show limited sensitivity to text semantics and no directed behavioral effect surviving FDR correction, demonstrating that correct-text accuracy alone does not establish discriminative language use. Molmo2 and JointNLT show supported directed effects, but their severity trajectories vary across settings: some contract significantly, while others remain inconclusive after correction.

Internal analyses localize this distinction. Text condition is decodable in four of five systems, showing that decodability alone is insufficient evidence of behavioral use. In the responsive systems, replacement at selected earlier sites produces larger changes than at peak-decoding states, but correct donors do not outperform unrelated donors, identifying sensitive stages without establishing target-specific semantic mediation. Neither architecture nor clean-input responsiveness consistently anticipates degradation behavior across the tested panel; robustness and correct language use should therefore be evaluated separately.

\paragraph{Future work.} Our internal analyses are restricted to MeViS and, under degradation, to JointNLT (Sec.~\ref{sec:limitations}). Extending them across datasets, models, and newer VLT architectures could reveal whether architectural properties explain the responsive/non-responsive split. Additional datasets and relational, temporal, and multi-turn text conditions are natural extensions of DISTRACT.

\clearpage
{
    \small
    \bibliographystyle{ieeenat_fullname}
    \bibliography{main}
}

\end{document}


\makeatletter
\setlength{\@fptop}{0pt}
\setlength{\@fpsep}{8pt plus 0fil}
\setlength{\@fpbot}{0pt plus 1fil}
\setlength{\@dblfptop}{0pt}
\setlength{\@dblfpsep}{8pt plus 0fil}
\setlength{\@dblfpbot}{0pt plus 1fil}
\makeatother

\maketitlesupplementary
\appendix

\newcommand{\rawurl}[1]{{\tiny\ttfamily\detokenize{#1}}}
\newcommand{\splitrawurl}[2]{{\tiny\ttfamily\detokenize{#1}\newline\detokenize{#2}}}

\newcommand{\suppgridpanel}[3]{%
  \begin{minipage}[t]{0.485\textwidth}
    \centering
    \includegraphics[width=\linewidth]{#1}\par\vspace{-2pt}
    \captionsetup{type=figure}
    \captionof{figure}{#2}
    \label{#3}
  \end{minipage}%
}
\newcommand{\suppgridrow}[6]{%
  \suppgridpanel{#1}{#2}{#3}\hfill
  \suppgridpanel{#4}{#5}{#6}%
}
\newcommand{\suppiadset}[2]{%
  \begin{figure*}[t]
    \centering
    \suppgridrow
      {results/paper_s0_s5_v5_recalibrated/analysis/iad_matrices/#2/blur_iad_matrix_all_datasets_singlecol.png}
      {#1, blur: IAD matrix (Eq.~\ref{eq:iad}) across all four datasets.}
      {fig:heatmap_#2_blur}
      {results/paper_s0_s5_v5_recalibrated/analysis/iad_matrices/#2/noise_iad_matrix_all_datasets_singlecol.png}
      {#1, noise: IAD matrix (Eq.~\ref{eq:iad}) across all four datasets.}
      {fig:heatmap_#2_noise}
    \par\vspace{2pt}
    \suppgridrow
      {results/paper_s0_s5_v5_recalibrated/analysis/iad_matrices/#2/pixelation_iad_matrix_all_datasets_singlecol.png}
      {#1, pixelation: IAD matrix (Eq.~\ref{eq:iad}) across all four datasets.}
      {fig:heatmap_#2_pixelation}
      {results/paper_s0_s5_v5_recalibrated/analysis/iad_matrices/#2/compression_iad_matrix_all_datasets_singlecol.png}
      {#1, compression: IAD matrix (Eq.~\ref{eq:iad}) across all four datasets.}
      {fig:heatmap_#2_compression}
  \end{figure*}%
}
\newcommand{\supptargetset}[2]{%
  \begin{figure*}[t]
    \centering
    \suppgridrow
      {results/paper_s0_s5_v5_recalibrated/analysis/combined_curves/#2/dancetrack/target_accuracy_2x2.png}
      {#1 on DanceTrack: target accuracy across all four perturbations, for all four text conditions.}
      {fig:supp_#2_dancetrack_acc}
      {results/paper_s0_s5_v5_recalibrated/analysis/combined_curves/#2/sportsmot/target_accuracy_2x2.png}
      {#1 on SportsMOT: target accuracy across all four perturbations, for all four text conditions.}
      {fig:supp_#2_sportsmot_acc}
    \par\vspace{2pt}
    \suppgridrow
      {results/paper_s0_s5_v5_recalibrated/analysis/combined_curves/#2/personpath22/target_accuracy_2x2.png}
      {#1 on PersonPath22: target accuracy across all four perturbations, for all four text conditions.}
      {fig:supp_#2_personpath22_acc}
      {results/paper_s0_s5_v5_recalibrated/analysis/combined_curves/#2/mevis/target_accuracy_2x2.png}
      {#1 on MeViS: target accuracy across all four perturbations, for all four text conditions.}
      {fig:supp_#2_mevis_acc}
  \end{figure*}%
}

\section{Benchmark and Protocol}

\subsection{Models}
\label{sec:supp_models}

Table~\ref{tab:supp_models_access} lists the exact model revisions,
checkpoints, source URLs, and stated license status used in this study.

\begin{table*}[t]
  \centering
  \scriptsize
  \setlength{\tabcolsep}{4pt}
  \renewcommand{\arraystretch}{1.03}
  \begin{tabular}{@{}p{1.05cm}p{2.65cm}p{2.95cm}p{4.45cm}p{1.75cm}@{}}
    \toprule
    System & Native operation & Pinned artifact used & Source / download URL & License status \\
    \midrule
    Molmo2 & 8B VLM; clean point plus native tracking prompt $\rightarrow$ point trajectory &
    \texttt{allenai/Molmo2-8B}, revision \texttt{e28fa285} &
    \rawurl{https://huggingface.co/allenai/Molmo2-8B}\newline
    \rawurl{https://github.com/allenai/molmo2} & Apache-2.0 \\
    JointNLT & BERT text encoder and visual--language fusion; clean box $\rightarrow$ box trajectory &
    commit \texttt{94b338e1}; \texttt{JointNLT\_ep0300.}\newline\texttt{pth.tar}; \texttt{swin\_b\_ep300\_track} &
    \rawurl{https://github.com/lizhou-cs/JointNLT}\newline
    \rawurl{https://drive.google.com/drive/folders/}\newline
    \splitrawurl{1ZKBDhgXgV1GLSzFxw5fqpqd}{ZuX1hjfQe?usp=share_link} & MIT \\
    UVLTrack & Language-conditioned box tracker; clean box and raw expression $\rightarrow$ box trajectory &
    commit \texttt{6ca34055}; \texttt{UVLTrack\_ep0300.}\newline\texttt{pth.tar} (baseline-base) &
    \rawurl{https://github.com/OpenSpaceAI/UVLTrack}\newline
    \rawurl{https://drive.google.com/drive/folders/}\newline
    \splitrawurl{1UZTrGcL3YlxvNpHi0wKsO_s}{KsTYuYnFo?usp=sharing} & MIT \\
    MMTrack & Language-conditioned multimodal tracker; clean box and raw expression $\rightarrow$ box trajectory &
    commit \texttt{4586b013}; \texttt{MMTrack\_ep0150.}\newline\texttt{pth.tar} (baseline) &
    \rawurl{https://github.com/Azong-HQU/MMTrack}\newline
    \rawurl{https://drive.google.com/file/d/}\newline
    \splitrawurl{1kM8qxwJHJg4HuZNPPTMFZZTG}{EdtASVAf/view?usp=sharing} & MIT \\
    DUTrack & Description-conditioned box tracker; static external expression plus clean box $\rightarrow$ box trajectory &
    commit \texttt{490d9985}; \texttt{DUTrack\_ep0045.}\newline\texttt{pth.tar} (\texttt{dutrack\_256\_full}) &
    \rawurl{https://github.com/GXNU-ZhongLab/}\newline\rawurl{DUTrack}\newline
    \rawurl{https://drive.google.com/drive/folders/}\newline
    \splitrawurl{1edieIddvSzy9219F4WShajv}{Y4UmkKJV4?usp=sharing} & No license file in pinned repo \\
    \bottomrule
  \end{tabular}
  \caption{Models, pinned artifacts, upstream source URLs, and stated license status.}
  \label{tab:supp_models_access}
\end{table*}

\subsection{Datasets}
\label{sec:supp_datasets}

Table~\ref{tab:supp_datasets_access} records dataset and annotation
provenance. The benchmark uses MeViS and three MOT sources. MeViS supplies native
expressions and masks. For DanceTrack, SportsMOT, and PersonPath22, source
videos and identities come from the original releases, while expressions and
point/mask trajectories come from Molmo2-VideoTrackEval. The selected-pair
manifest retains source-video, clip, and annotation-row identifiers. Source-media
terms and the separate Molmo2-VideoTrackEval annotation license are reported
independently: the latter applies to its expressions, masks, and points, not to
the underlying videos.

\begin{table*}[t]
  \centering
  \scriptsize
  \setlength{\tabcolsep}{4pt}
  \renewcommand{\arraystretch}{1.03}
  \begin{tabular}{@{}p{1.45cm}p{4.9cm}p{4.65cm}p{2.7cm}@{}}
    \toprule
    Dataset & Source media and annotations used & Access URL(s) & License / use terms \\
    \midrule
    MeViS & 120 pairs from 120 source videos. Native human-written motion expressions and masks provide both text and geometry. Source media are obtained from the upstream release. &
    \rawurl{https://github.com/henghuiding/MeViS} &
    Upstream data: CC BY-NC-SA 4.0\newline\rawurl{https://creativecommons.}\newline\splitrawurl{org/licenses/}{by-nc-sa/4.0/}; not redistributed here \\
    DanceTrack & 60 pairs. DanceTrack provides source video and identities; Molmo2-VideoTrackEval provides human-written expressions and point/mask trajectories. &
    \rawurl{https://dancetrack.github.io/}\newline
    \splitrawurl{https://huggingface.co/datasets/allenai/}{Molmo2-VideoTrackEval} &
    Source video: non-commercial research; DanceTrack notes that Internet-sourced media are not project property.\newline Molmo2-VideoTrackEval expressions/masks/points: ODC-BY-1.0 \\
    SportsMOT & 30 pairs. SportsMOT provides source video and identities; Molmo2-VideoTrackEval provides human-written expressions and point/mask trajectories. &
    \rawurl{https://github.com/MCG-NJU/SportsMOT}\newline
    \splitrawurl{https://huggingface.co/datasets/allenai/}{Molmo2-VideoTrackEval} &
    Source video: CC BY-NC 4.0\newline\rawurl{https://creativecommons.}\newline\splitrawurl{org/licenses/}{by-nc/4.0/}\newline Molmo2-VideoTrackEval expressions/masks/points: ODC-BY-1.0 \\
    PersonPath22 & 30 pairs. PersonPath22 provides source video and identities; Molmo2-VideoTrackEval provides human-written expressions and point/mask trajectories. &
    \splitrawurl{https://github.com/amazon-science/}{tracking-dataset}\newline
    \splitrawurl{https://huggingface.co/datasets/allenai/}{Molmo2-VideoTrackEval} &
    Source release: CC BY-NC 4.0; some externally sourced videos are not redistributed by the provider.\newline Molmo2-VideoTrackEval expressions/masks/points: ODC-BY-1.0 \\
    \bottomrule
  \end{tabular}
  \caption{Dataset sources, annotation provenance, access URLs, and applicable upstream terms. Molmo2-VideoTrackEval annotation terms are distinct from source-media terms.}
  \label{tab:supp_datasets_access}
\end{table*}

\subsection{Inverse-probability-weighted sensitivity check}
\label{sec:supp_ipw}

For the three MOT datasets, Table~\ref{tab:supp_ipw} records an
inverse-probability-weighted sensitivity calculation for the 120-pair stress
subset and the 404-pair candidate pool. The guaranteed-quota and random-fill
draws are uniform random permutations within their defined strata, giving
the following marginal inclusion probabilities:
\begin{align*}
  P(\text{pos}) &= \tfrac{n_g}{n_{\mathrm{pos}}} + \big(1-\tfrac{n_g}{n_{\mathrm{pos}}}\big)\tfrac{f}{n_{\mathrm{pos}}-n_g+n_{\mathrm{neg}}}, \\
  P(\text{neg}) &= \tfrac{f}{n_{\mathrm{pos}}-n_g+n_{\mathrm{neg}}},
\end{align*}
where $n_g$ is the guaranteed-quota size and $f=$ target$-n_g$ is the
random-fill size.

\begin{table}[htbp]
  \centering
  \scriptsize
  \setlength{\tabcolsep}{3pt}
  \begin{tabular}{@{}llrrr@{}}
    \toprule
    Dataset & Effect & Naive (selected) & IPW & Full pool \\
    \midrule
    DanceTrack   & $B_T$ & $+12.8\%$ & $+11.3\%$ & $+11.4\%$ \\
                 & $N_T$ & $+7.6\%$  & $+7.3\%$  & $+9.4\%$ \\
                 & $C_T$ & $+25.3\%$ & $+21.5\%$ & $+18.0\%$ \\
    SportsMOT    & $B_T$ & $+0.2\%$  & $+0.2\%$  & $+1.2\%$ \\
                 & $N_T$ & $+3.8\%$  & $+2.2\%$  & $+2.6\%$ \\
                 & $C_T$ & $+5.5\%$  & $+1.6\%$  & $+4.9\%$ \\
    PersonPath22 & $B_T$ & $+0.7\%$  & $+0.6\%$  & $+6.7\%$ \\
                 & $N_T$ & $+3.6\%$  & $+4.0\%$  & $+6.7\%$ \\
                 & $C_T$ & $+14.3\%$ & $+9.3\%$  & $+12.5\%$ \\
    \midrule
    Pooled (3 datasets) & $B_T$ & $+6.6\%$  & $+5.8\%$  & $+7.7\%$ \\
                         & $N_T$ & $+5.7\%$  & $+5.2\%$  & $+7.0\%$ \\
                         & $C_T$ & $+17.6\%$ & $+13.5\%$ & $+13.4\%$ \\
    \bottomrule
  \end{tabular}
  \caption{Molmo2 clean-input $B_T$, $N_T$, and $C_T$: unweighted
  stress-subset estimate, inverse-probability-weighted estimate, and
  candidate-pool mean.}
  \label{tab:supp_ipw}
\end{table}

\subsection{Annotation conversion and common scoring}
\label{sec:supp_bbox}

For segmentation datasets, the initialization box is the tight, axis-aligned
bounding box of the mask foreground. Concretely, for mask $\mathcal{M}$ with
foreground pixel coordinates $(x,y)$:
\begin{align*}
  x_{\min} &= \min_{(x,y)\in\mathcal{M}} x, &
  y_{\min} &= \min_{(x,y)\in\mathcal{M}} y,\\
  x_{\max} &= \max_{(x,y)\in\mathcal{M}} x, &
  y_{\max} &= \max_{(x,y)\in\mathcal{M}} y,
\end{align*}
\[
  b = \big(x_{\min},\; y_{\min},\; x_{\max}-x_{\min},\; y_{\max}-y_{\min}\big).
\]
Tracker boxes are reduced to their centers
$\hat p=(x+w/2,y+h/2)$ for common scoring; Molmo2 supplies its point directly.
Future annotations are not used for initialization or inference.

\subsection{Text-condition construction}
\label{sec:supp_category}

Each pair has target and distractor expressions $e_T,e_D$. The neutral input
contains no target identity; the unrelated donor $e_U$ is drawn once from a
different annotated clip with the same category label and then fixed across
models and severities. The donor pool never relaxes this category constraint.
For DanceTrack, SportsMOT, and PersonPath22, the category label is the
dataset's single coarse class; MeViS uses the annotation category.
Table~\ref{tab:supp_text_realization} records the concrete input supplied to
each model under each condition.

\begin{table}[htbp]
  \centering
  \scriptsize
  \setlength{\tabcolsep}{3pt}
  \begin{tabular}{@{}lll@{}}
    \toprule
    Condition & Molmo2 input after its fixed scaffold & Specialized trackers \\
    \midrule
    Neutral & no appended identity description & \texttt{object} \\
    Unrelated & ``The target is: $e_U$'' & $e_U$ \\
    Correct & ``The target is: $e_T$'' & $e_T$ \\
    Conflicting & ``The target is: $e_D$'' & $e_D$ \\
    \bottomrule
  \end{tabular}
  \caption{Text input under each condition for Molmo2 and the specialized trackers.}
  \label{tab:supp_text_realization}
\end{table}

\subsection{Native inputs, outputs, and state reset}
\label{sec:supp_interface}

Table~\ref{tab:supp_native_interfaces} records each native interface and the
state cleared between runs.

\begin{table*}[htbp]
  \centering
  \scriptsize
  \setlength{\tabcolsep}{4pt}
  \begin{tabular}{@{}lllll@{}}
    \toprule
    System & Initialization & Text interface & Output & Reset safeguard \\
    \midrule
    Molmo2 & clean target point & native point-tracking scaffold & point trajectory & fresh generation state \\
    JointNLT & clean tight box & raw expression, native \textsc{track} mode & box trajectory & ROI queue, template features, tracker state \\
    UVLTrack & clean tight box & raw expression & box trajectory & state, frame counter, score and output cache \\
    MMTrack & clean tight box & raw expression & box trajectory & adapter/tracker and decode state \\
    DUTrack & clean tight box & static external expression & box trajectory & dynamic description and tracker state \\
    \bottomrule
  \end{tabular}
  \caption{Native inputs, outputs, and reset state.}
  \label{tab:supp_native_interfaces}
\end{table*}

DUTrack's released dynamic-description path would otherwise overwrite the
assigned condition. Table~\ref{tab:supp_dutrack_policy_audit} records the
condition-preservation audit used to select the primary static-external
policy; it is a protocol verification, not an efficacy result.

\begin{center}
  \centering
  \scriptsize
  \setlength{\tabcolsep}{3pt}
  \begin{tabular}{@{}p{0.22\columnwidth}p{0.38\columnwidth}p{0.29\columnwidth}@{}}
    \toprule
    Policy/check & Text handling & Audit outcome \\
    \midrule
    Static external (primary) & Adapter resets the external provider to the assigned phrase before every tracked frame. & Holds the assigned benchmark condition fixed. \\
    Seeded dynamic (audit) & Assigned phrase seeds the first tracked step; native updater remains enabled. & Description string and active token IDs changed by frame two in 192/192 runs. \\
    \bottomrule
  \end{tabular}
  \captionof{table}{DUTrack condition-preservation audit. The seeded-dynamic row is an
  audit mode, not a benchmark condition.}
  \label{tab:supp_dutrack_policy_audit}
\end{center}

\subsection{Outcome definitions and behavioral inference}
\label{sec:supp_behavioral_inference}

For every post-initialization frame, a usable prediction is target,
distractor, or other according to the common point-in-support rule above;
missing, nonfinite, or invalid predictions are absent. Target takes precedence
when supports overlap, and absent frames remain in every denominator. For each
pair, condition, and displayed severity, frame fractions are aggregated within
pair before pairs are aggregated. In addition to $B_T,N_T,C_T$ in the main
paper, the endpoint family contains
\begin{align}
 L_T(s) &= A_T(\mathrm{unrelated},s)-A_T(\mathrm{conflicting},s),\\
 S_D(s) &= A_D(\mathrm{conflicting},s)-A_D(\mathrm{unrelated},s).
\end{align}
Persistent takeover and recovery require three consecutive distractor and
subsequent target labels, respectively. For a fixed model, dataset,
perturbation, and displayed severity $s\in\{0,\ldots,5\}$, let $d_{p,s}$
denote a paired effect for pair $p$; for example,
$d_{p,s}=A_T^p(\mathrm{correct},s)-A_T^p(\mathrm{unrelated},s)$ for
$B_T$. The reported point estimate is the pair-weighted mean
\begin{equation}
 \hat\theta_s=\frac{1}{N}\sum_{p=1}^{N}d_{p,s}.
 \label{eq:supp_behavioral_mean}
\end{equation}
Pairs from the same source video are not treated as independent. Let
$g(p)\in\mathcal V$ assign pair $p$ to its source video, and let
$S_v=\sum_{p:g(p)=v}d_{p,s}$ and $n_v=\sum_p\mathbf{1}[g(p)=v]$.

For each of $R=10{,}000$ bootstrap replicates, we draw $|\mathcal V|$
videos with replacement, $Z_1^{(r)},\ldots,Z_{|\mathcal V|}^{(r)}$, and
retain every selected pair. The replicate estimate is
\begin{equation}
 \hat\theta_s^{(r)}=
 \frac{\sum_{j=1}^{|\mathcal V|}S_{Z_j^{(r)}}}
      {\sum_{j=1}^{|\mathcal V|}n_{Z_j^{(r)}}}.
 \label{eq:supp_behavioral_bootstrap}
\end{equation}
The 95\% CI is the empirical $[2.5,97.5]$ percentile interval.

For a two-sided paired sign-flip test, we independently draw
$\epsilon_v^{(r)}\in\{-1,+1\}$ for every source video and form
\begin{equation}
 \hat\theta_{s,0}^{(r)}=
 \frac{1}{N}\sum_{v\in\mathcal V}\epsilon_v^{(r)}S_v,
 \qquad
 p=\frac{1}{R}\sum_{r=1}^{R}
 \mathbf{1}\!\left[|\hat\theta_{s,0}^{(r)}|\geq|\hat\theta_s|\right].
 \label{eq:supp_behavioral_signflip}
\end{equation}
For a severity-endpoint test, we apply the same procedure to
$d_{p,S_5}-d_{p,S_0}$.

For every model--dataset experiment, the endpoint family contains the five
effects $B_T,N_T,C_T,L_T,S_D$, each tested for nonzero effect at $S_5$ and
for the $S_0\rightarrow S_5$ change within each of the four perturbations
($40$ tests), plus one nonzero-effect-at-$S_0$ test per effect ($5$ tests):
the $S_0$ record is the shared clean baseline and is identical across every
perturbation's ladder, so it is tested once per effect rather than once per
perturbation, for $45$ tests total. We apply Benjamini--Hochberg to all
$m$ p-values in this family. For ordered p-values $p_{(1)}\leq\cdots\leq
p_{(m)}$, the reported adjusted value is
\begin{equation}
 q_{(i)}=\min\!\left\{1,\min_{j\geq i}\frac{m}{j}p_{(j)}\right\}.
 \label{eq:supp_behavioral_bh}
\end{equation}
IAD is excluded from this testing family because it is descriptive.

\subsection{Perturbation realization and displayed severity}

Each corruption is applied independently to every post-initialization frame
(the initialization frame is always clean); corruptions are never composed.
For a frame $I\in[0,255]^{H\times W\times 3}$, the four perturbations are
implemented as follows.

\emph{Blur.} Per-channel Gaussian smoothing with $\sigma_s=k_s\cdot\min(H,W)$,
kernel truncated at $\pm3\sigma_s$: $I'_{\cdot,\cdot,c}=I_{\cdot,\cdot,c}\ast
\mathcal N(0,\sigma_s^2)$ for each channel $c$. $k_0=0$ is an exact identity.

\emph{Noise.} The frame is normalized to $[0,1]$; a single standard-normal
field $\eta\sim\mathcal N(0,1)^{H\times W\times 3}$, seeded deterministically
from a hash of (video, frame index) only -- never condition, pair, or
severity -- is drawn once per frame and reused unchanged across all four
text conditions and every severity for that frame:
$I'=\mathrm{round}\!\left(255\cdot\mathrm{clip}\!\left(\frac{I}{255}+\sigma_s\,
\eta,\,0,\,1\right)\right)$. $\sigma_0=0$ is an exact identity; because only
the scale $\sigma_s$ changes for a fixed draw of $\eta$, severities differ in
dose, not in noise realization, so contrasts across conditions at one
severity are never confounded by different noise instances.

\emph{Pixelation.} Block-mosaic resampling: downscale to
$(\mathrm{round}(r_sW),\mathrm{round}(r_sH))$ and upscale back to $(W,H)$,
both steps using PIL's BOX (area-average) filter -- the ImageNet-C
\texttt{pixelate} corruption. $r_0=1$ is an exact identity.

\emph{Compression.} The frame is re-encoded to JPEG at quality $q_s$
(libjpeg's 1--100 scale) and immediately decoded back to RGB -- the
ImageNet-C \texttt{jpeg\_compression} corruption. $q_0=100$ is treated as an
exact identity (the untouched frame is passed through, not a quality-100
re-encode).

\begin{figure}[htbp]
  \centering
  \includegraphics[width=\columnwidth]{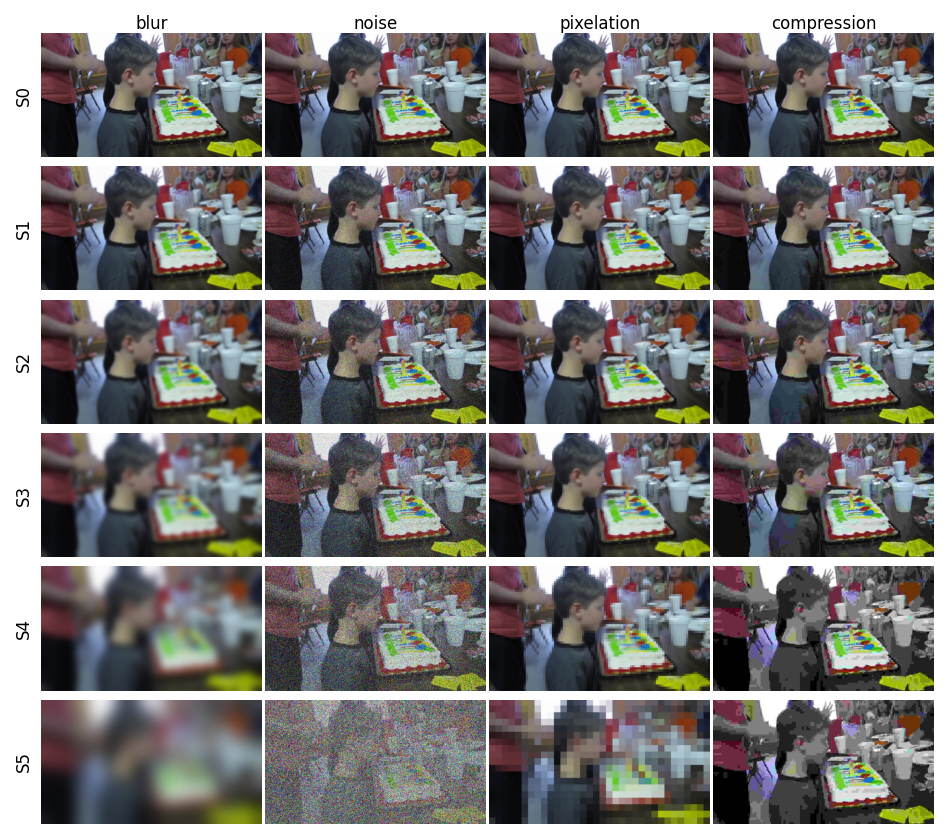}
  \caption{Displayed $S_0$--$S_5$ perturbation schedule on one MeViS frame.}
  \label{fig:supp_all_perturbation_calibration}
\end{figure}

Table~\ref{tab:supp_severity_remap} gives the raw-index subsequence used for
the displayed $S_0$--$S_5$ schedule. $S_0$ is raw index 0 for every
perturbation. Figure~\ref{fig:supp_all_perturbation_calibration} shows one
frame under the displayed schedule.

\begin{table}[htbp]
  \centering
  \scriptsize
  \setlength{\tabcolsep}{4pt}
  \begin{tabular}{@{}lrrrrrrr@{}}
    \toprule
    Perturbation & raw $n$ & $S_0$ & $S_1$ & $S_2$ & $S_3$ & $S_4$ & $S_5$ \\
    \midrule
    Blur        & 8 & 0 & 2 & 3 & 4 & 5 & 6 \\
    Noise       & 9 & 0 & 4 & 5 & 6 & 7 & 8 \\
    Pixelation  & 8 & 0 & 1 & 2 & 3 & 4 & 7 \\
    Compression & 7 & 0 & 2 & 3 & 4 & 5 & 6 \\
    \bottomrule
  \end{tabular}
  \caption{Raw implementation indices used for each displayed $S_0$--$S_5$ severity.}
  \label{tab:supp_severity_remap}
\end{table}

\subsection{Objective distortion audit}
\label{sec:supp_distortion_audit}

Table~\ref{tab:supp_distortion_audit} reports clean-to-$S_5$ PSNR and SSIM
over a deterministic 10-video, one-frame-per-video sample.

\begin{center}
  \centering
  \scriptsize
  \setlength{\tabcolsep}{4pt}
  \begin{tabular}{@{}lrrrr@{}}
    \toprule
    Perturbation & PSNR mean (dB) & PSNR range & SSIM mean & SSIM range \\
    \midrule
    Blur        & 19.5 & [17.2, 24.0] & 0.72 & [0.40, 0.85] \\
    Noise       & 6.1  & [5.6, 6.4]   & 0.02 & [0.01, 0.03] \\
    Pixelation  & 22.1 & [19.2, 26.1] & 0.70 & [0.40, 0.82] \\
    Compression & 23.7 & [20.8, 26.1] & 0.73 & [0.55, 0.84] \\
    \bottomrule
  \end{tabular}
  \captionof{table}{Clean-to-displayed-$S_5$ PSNR and SSIM on 10 fixed
  videos (one frame per video); ranges are min--max.}
  \label{tab:supp_distortion_audit}
\end{center}

\section{Results}

\subsection{Behavioral degradation study}
\label{sec:supp_extended_results}

This section provides the complete behavioral-result grid across five models,
four datasets, and four perturbations.

For standalone readability, with $\Delta_{c_1,c_2}(s)$ denoting the
target-accuracy difference between conditions at severity $s$, we use
\begin{equation}
\mathrm{IAD}(c_1,c_2)=\sum_{s=0}^{4}\tfrac12\left(
|\Delta_{c_1,c_2}(s)|+|\Delta_{c_1,c_2}(s{+}1)|\right).
\label{eq:iad}
\end{equation}

\subsection{IAD matrices by model}
\label{sec:supp_heatmaps}

Each figure gives the upper triangle of the neutral/unrelated/correct/
conflicting IAD matrix for one model and perturbation, with datasets as
panels. Each cell is the accumulated absolute target-accuracy difference
between its row and column conditions.
For each system, the blur, noise, pixelation, and compression matrices are
given in Figures~\ref{fig:heatmap_molmo2_blur}--\ref{fig:heatmap_molmo2_compression}
(Molmo2); Figures~\ref{fig:heatmap_uvltrack_blur}--\ref{fig:heatmap_uvltrack_compression}
(UVLTrack); Figures~\ref{fig:heatmap_mmtrack_blur}--\ref{fig:heatmap_mmtrack_compression}
(MMTrack); Figures~\ref{fig:heatmap_jointnlt_blur}--\ref{fig:heatmap_jointnlt_compression}
(JointNLT); and Figures~\ref{fig:heatmap_dutrack_blur}--\ref{fig:heatmap_dutrack_compression}
(DUTrack).
\clearpage
\suppiadset{Molmo2}{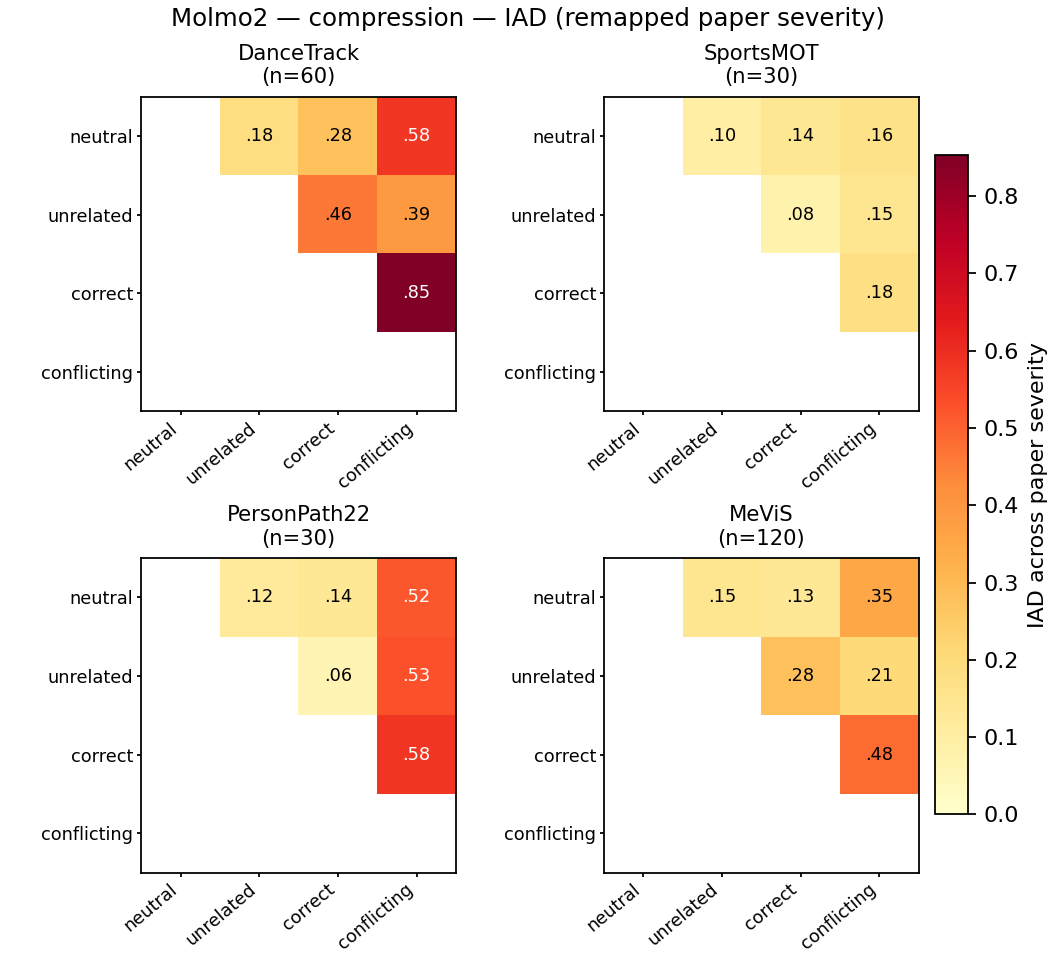}

\clearpage
\suppiadset{UVLTrack}{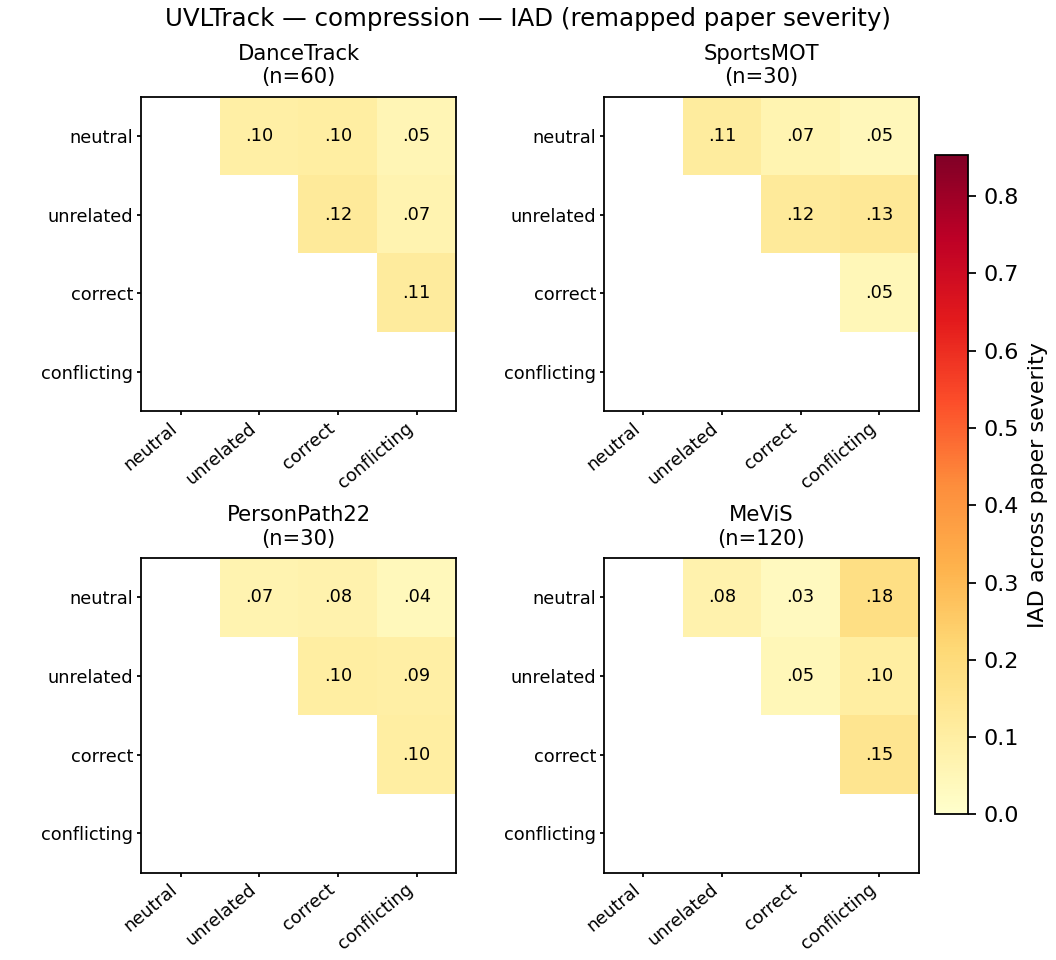}

\clearpage
\suppiadset{MMTrack}{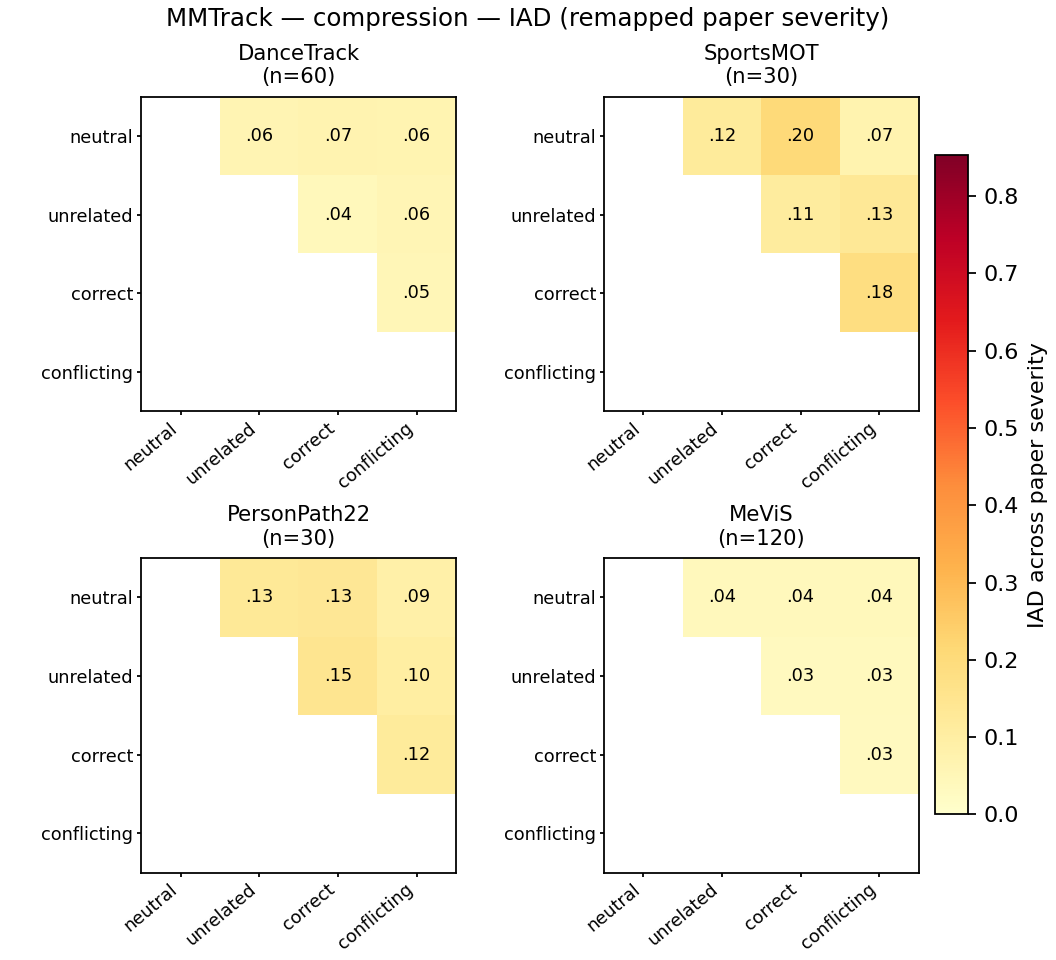}

\clearpage
\suppiadset{JointNLT}{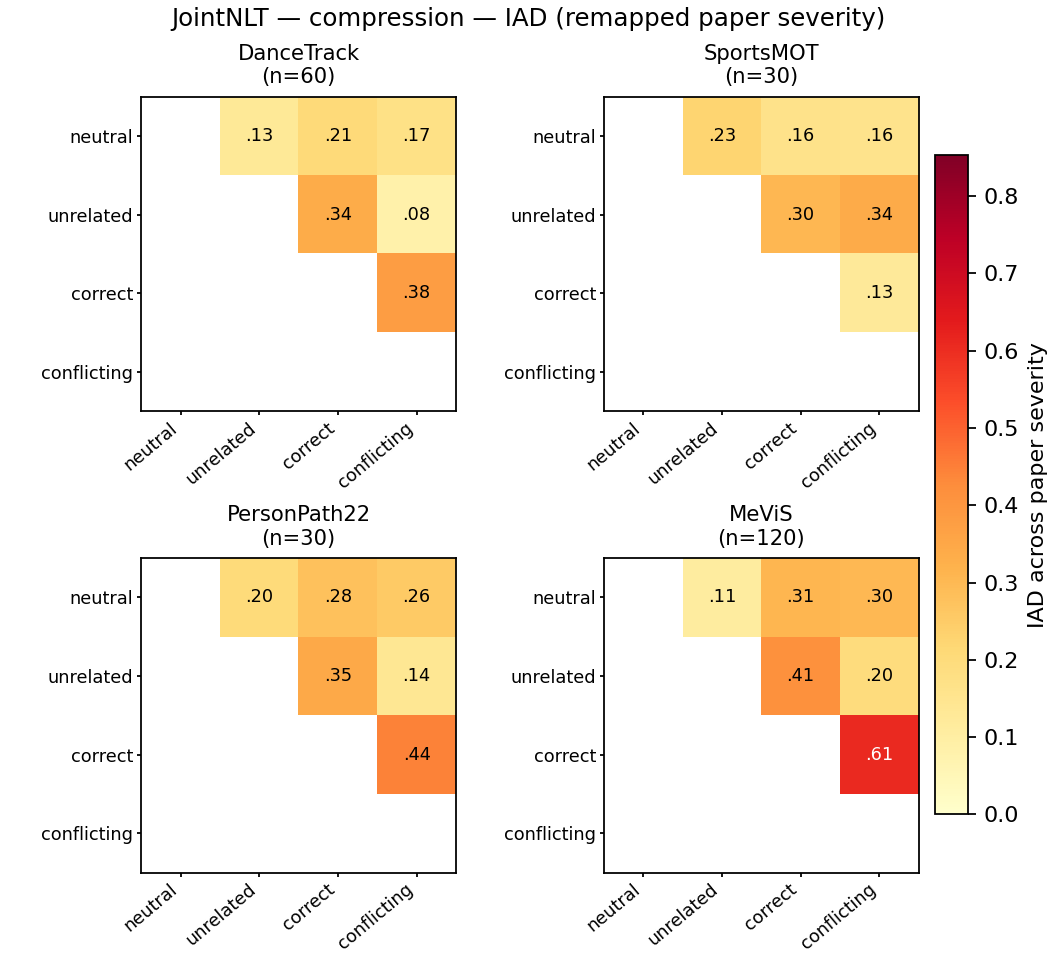}

\clearpage
\suppiadset{DUTrack}{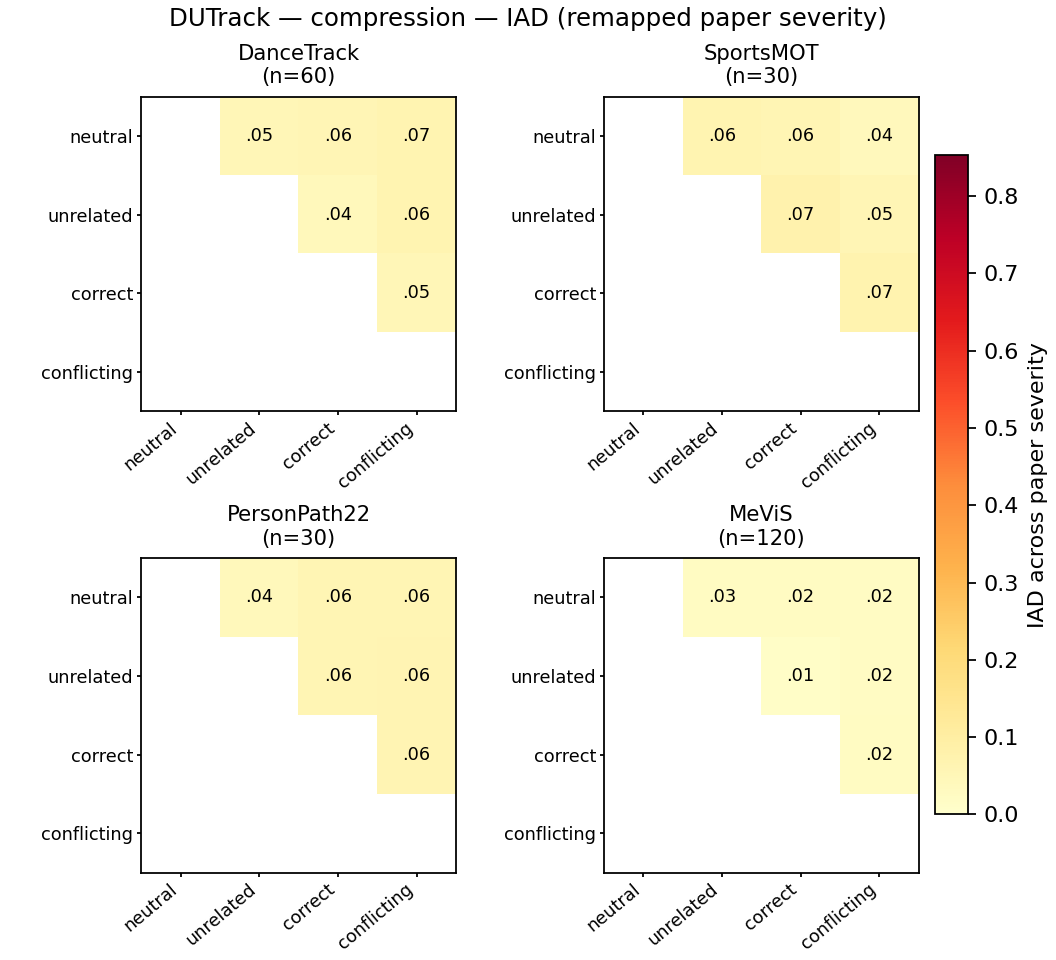}

\clearpage
\subsection{Signed severity trajectories}
\label{sec:supp_regime_grid}

\begin{center}
  \captionsetup{type=figure}
  \begin{subfigure}[t]{\columnwidth}
    \centering
    \includegraphics[width=\linewidth]{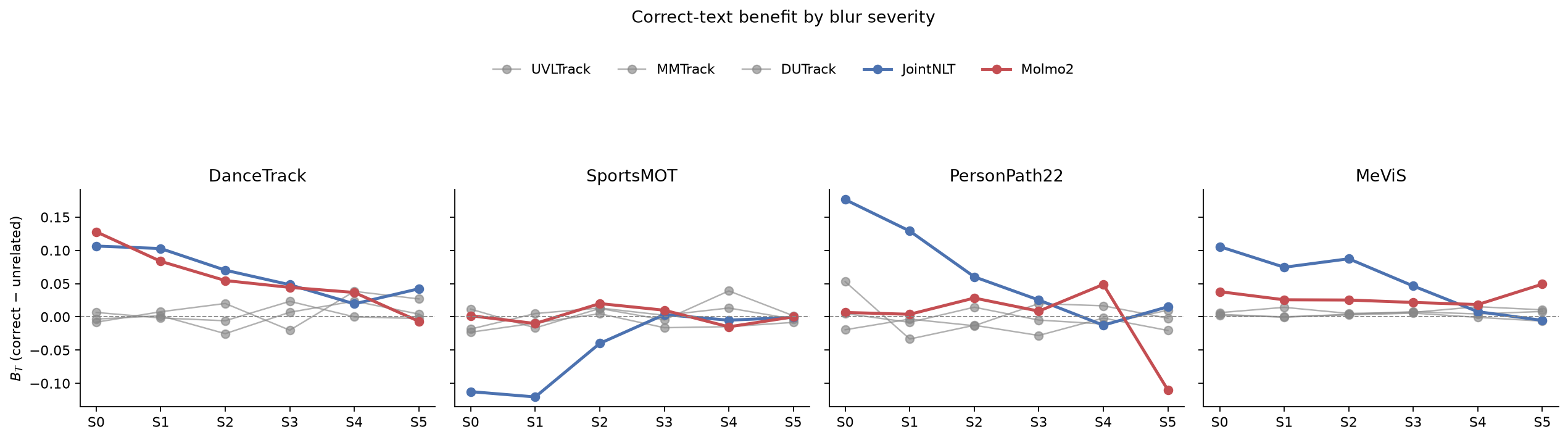}
    \caption{Blur}
  \end{subfigure}\\[2pt]
  \begin{subfigure}[t]{\columnwidth}
    \centering
    \includegraphics[width=\linewidth]{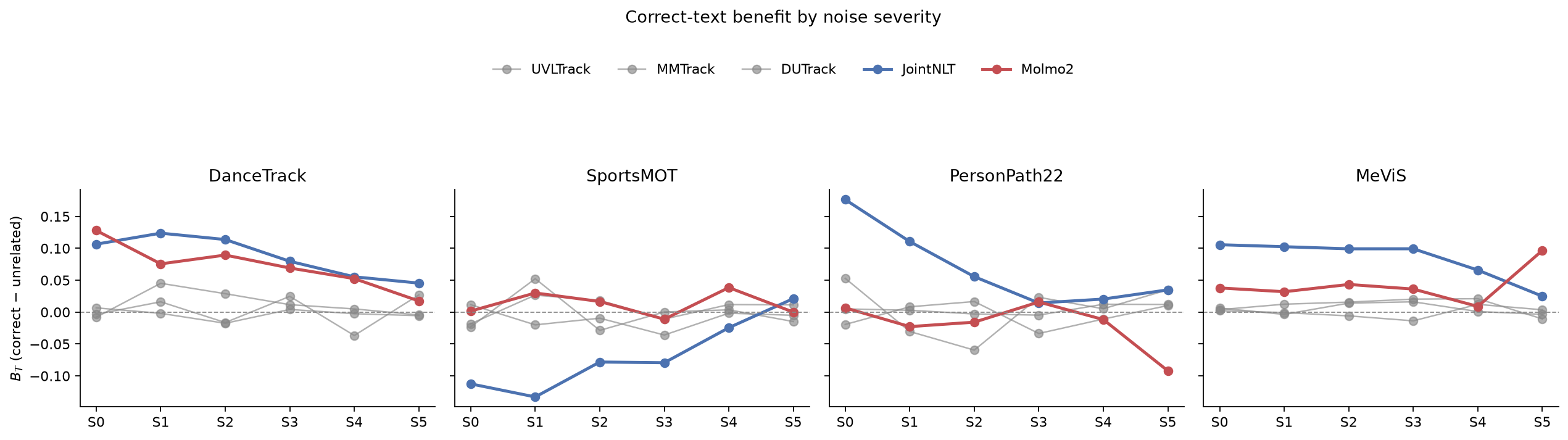}
    \caption{Noise}
  \end{subfigure}\\[2pt]
  \begin{subfigure}[t]{\columnwidth}
    \centering
    \includegraphics[width=\linewidth]{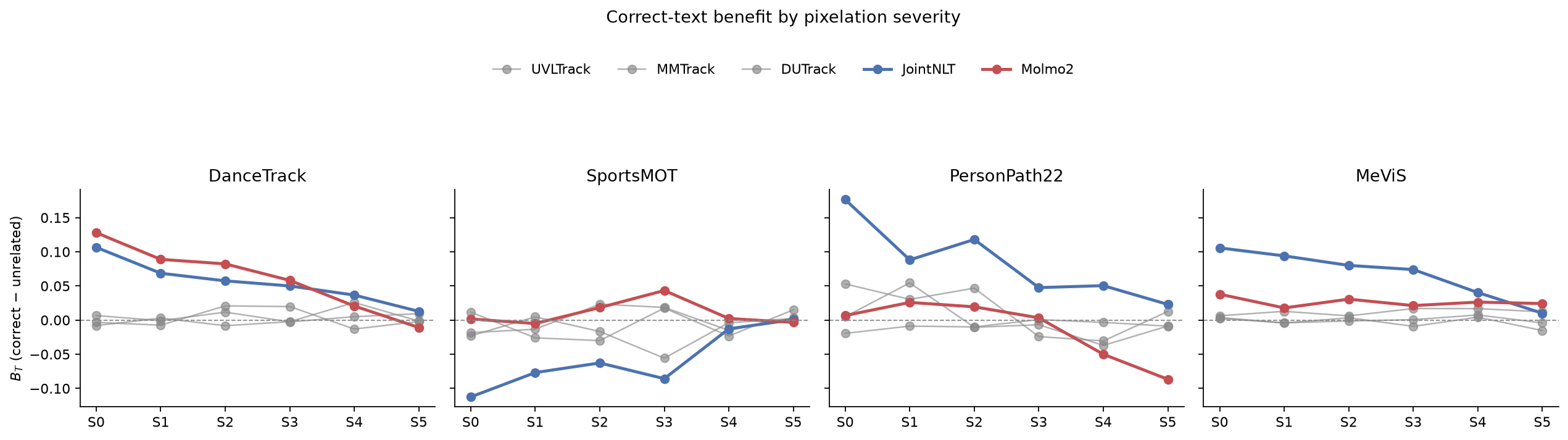}
    \caption{Pixelation}
  \end{subfigure}\\[2pt]
  \begin{subfigure}[t]{\columnwidth}
    \centering
    \includegraphics[width=\linewidth]{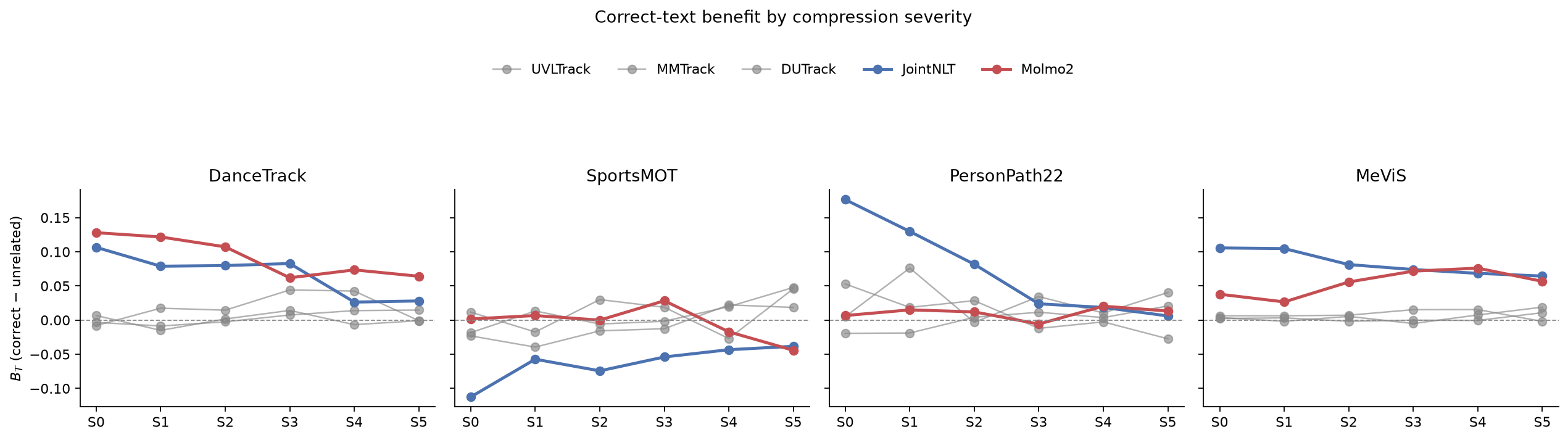}
    \caption{Compression}
  \end{subfigure}
  \captionof{figure}{Correct-versus-unrelated target-accuracy effect $B_T$
  across severity, systems, datasets, and perturbations.}
  \label{fig:supp_bt_all}
\end{center}

Figure~\ref{fig:supp_bt_all} provides the signed $B_T$ trajectories that
complement the unsigned IAD matrices.

\subsection{Four-condition trajectories, all systems}

The following figures provide condition-specific target-accuracy and
distractor-capture trajectories for every model--dataset pair.
Figure~\ref{fig:supp_capture_iad_pooled} gives the pooled
distractor-capture summary before the per-model--dataset trajectories.

\begin{center}
  \includegraphics[width=\columnwidth]{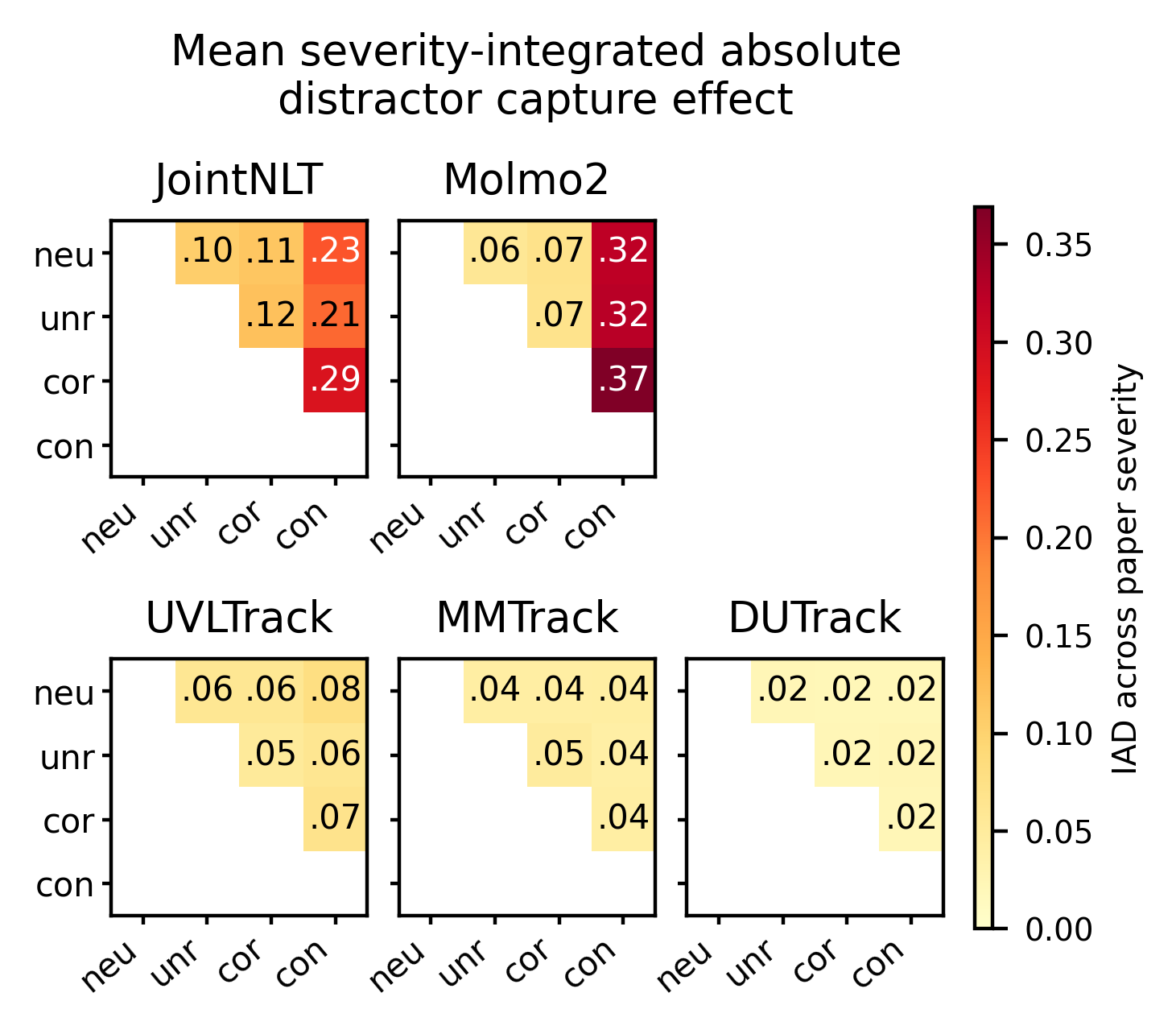}
  \captionof{figure}{Mean severity-integrated distractor-capture differences,
  pooled across datasets and perturbations.}
  \label{fig:supp_capture_iad_pooled}
\end{center}

Figures~\ref{fig:supp_molmo2_dancetrack_acc}--\ref{fig:supp_molmo2_mevis_acc}
give Molmo2 target-accuracy trajectories; Figures~\ref{fig:supp_molmo2_dancetrack_cap}--\ref{fig:supp_molmo2_mevis_cap}
give distractor-capture trajectories. \\
Figures~\ref{fig:supp_uvltrack_dancetrack_acc}--\ref{fig:supp_uvltrack_mevis_acc}
give UVLTrack target-accuracy trajectories; Figures~\ref{fig:supp_uvltrack_dancetrack_cap}--\ref{fig:supp_uvltrack_mevis_cap}
give distractor-capture trajectories.\\
Figures~\ref{fig:supp_mmtrack_dancetrack_acc}--\ref{fig:supp_mmtrack_mevis_acc}
give MMTrack target-accuracy trajectories; Figures~\ref{fig:supp_mmtrack_dancetrack_cap}--\ref{fig:supp_mmtrack_mevis_cap}
give distractor-capture trajectories.\\
Figures~\ref{fig:supp_jointnlt_dancetrack_acc}--\ref{fig:supp_jointnlt_mevis_acc}
give JointNLT target-accuracy trajectories; Figures~\ref{fig:supp_jointnlt_dancetrack_cap}--\ref{fig:supp_jointnlt_mevis_cap}
give distractor-capture trajectories.\\
Figures~\ref{fig:supp_dutrack_dancetrack_acc}--\ref{fig:supp_dutrack_mevis_acc}
give DUTrack target-accuracy trajectories; Figures~\ref{fig:supp_dutrack_dancetrack_cap}--\ref{fig:supp_dutrack_mevis_cap}
give distractor-capture trajectories.\\
\clearpage
\supptargetset{Molmo2}{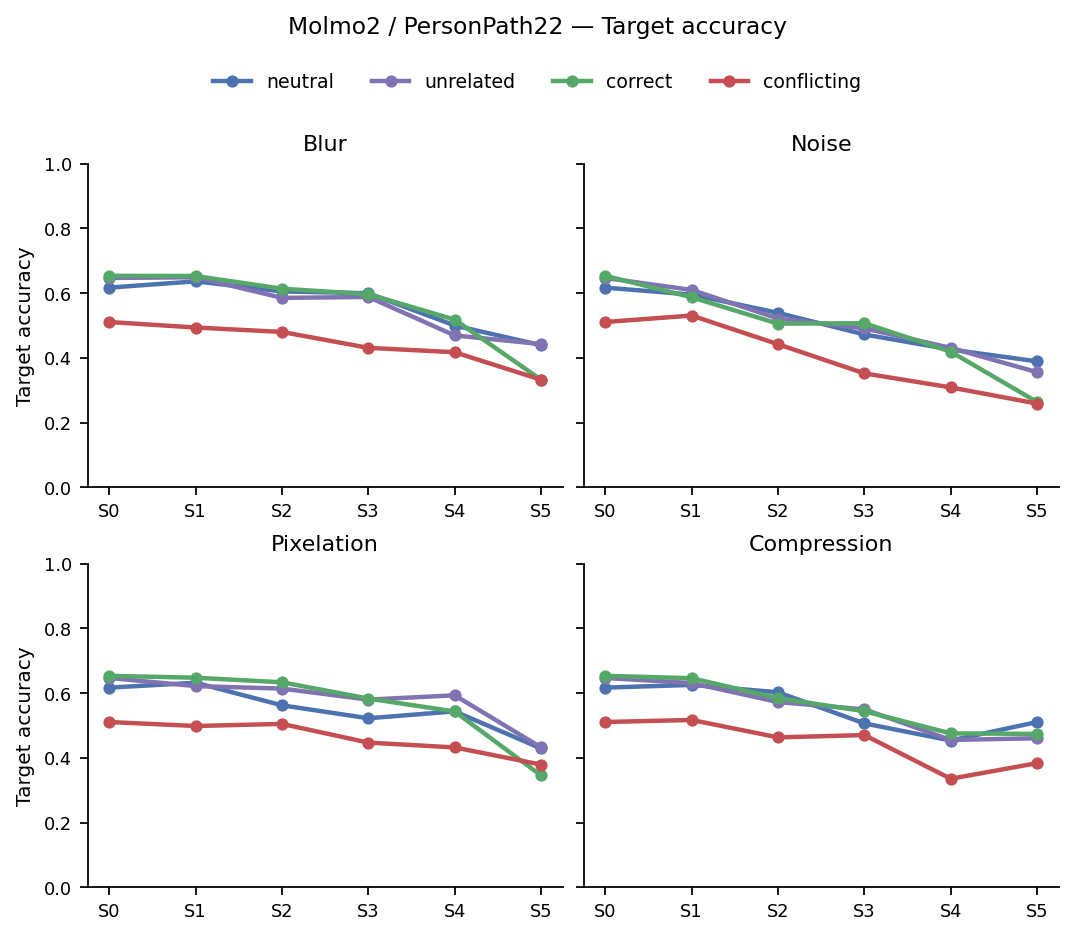}
\clearpage
\begin{center}
  \includegraphics[width=\columnwidth]{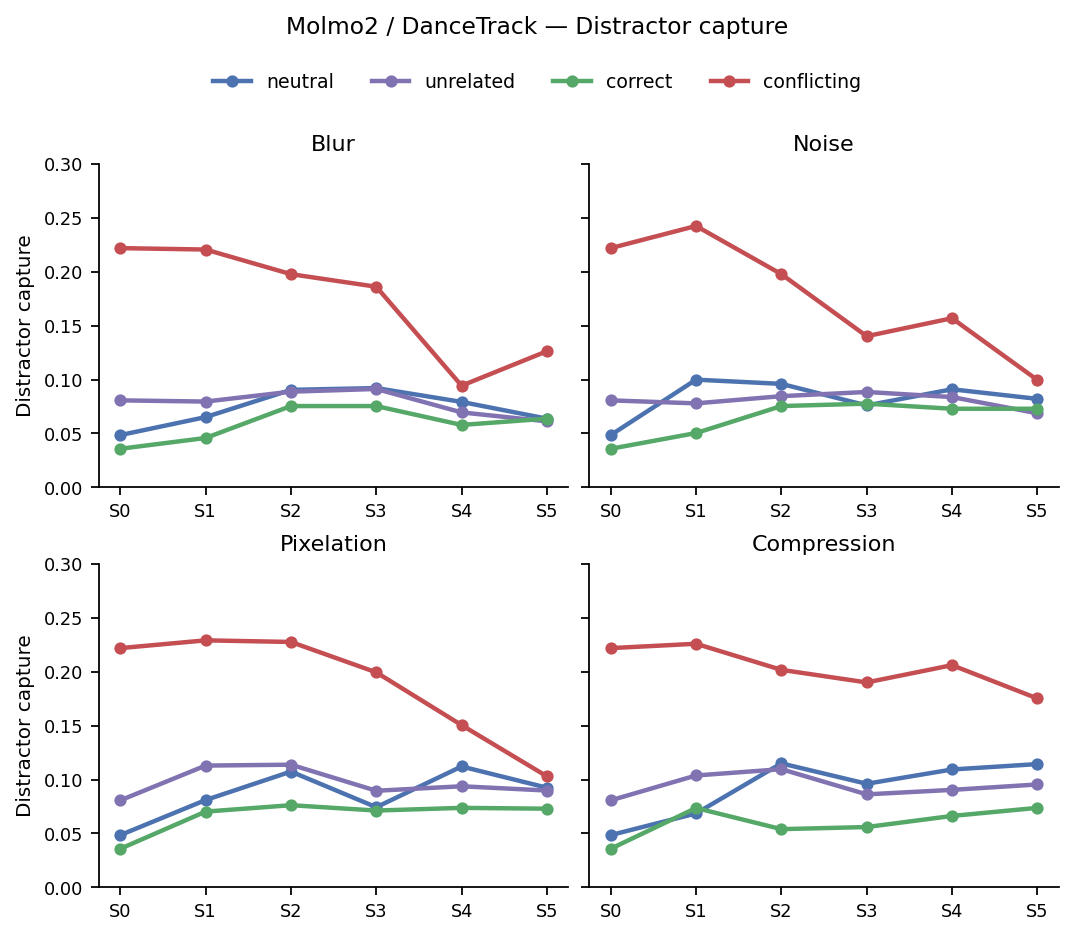}
  \captionof{figure}{Molmo2 on DanceTrack: distractor capture across all four perturbations, for all four text conditions.}
  \label{fig:supp_molmo2_dancetrack_cap}
\end{center}
\begin{center}
  \includegraphics[width=\columnwidth]{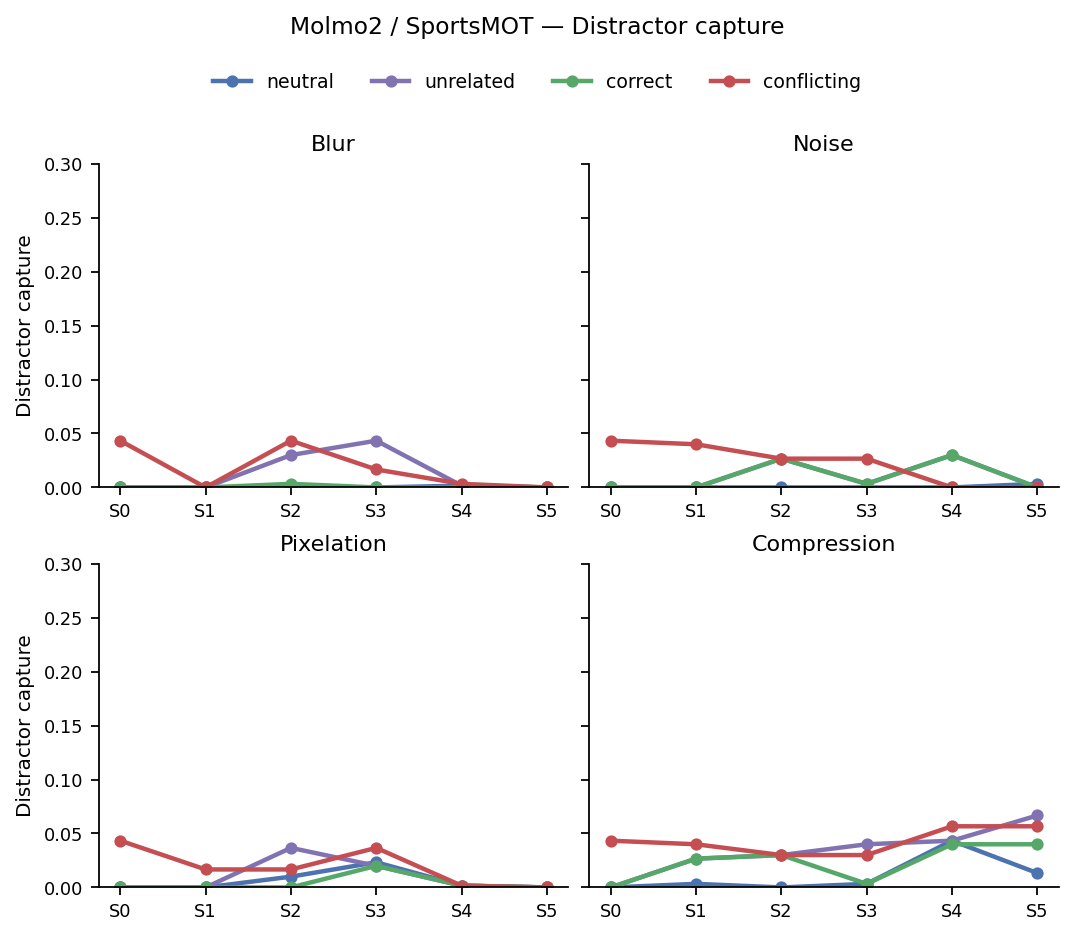}
  \captionof{figure}{Molmo2 on SportsMOT: distractor capture across all four perturbations, for all four text conditions.}
  \label{fig:supp_molmo2_sportsmot_cap}
\end{center}
\begin{center}
  \includegraphics[width=\columnwidth]{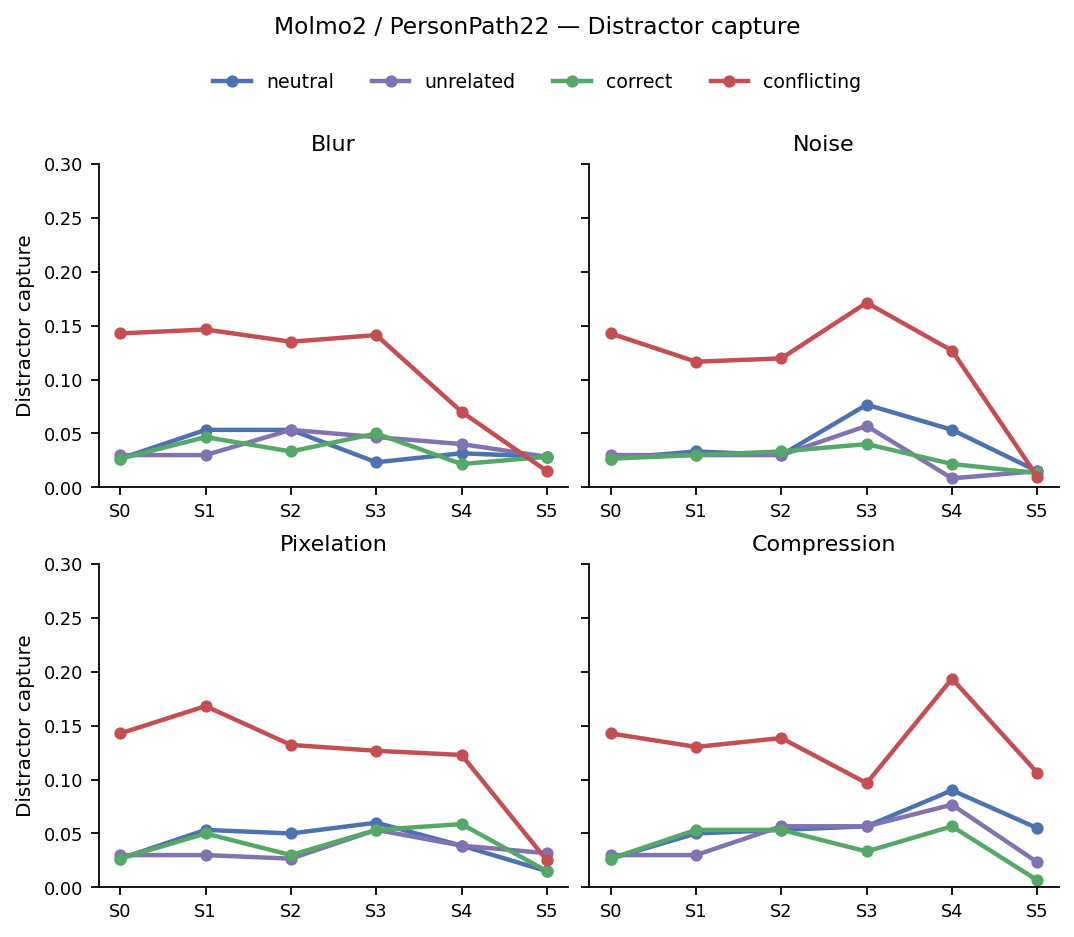}
  \captionof{figure}{Molmo2 on PersonPath22: distractor capture across all four perturbations, for all four text conditions.}
  \label{fig:supp_molmo2_personpath22_cap}
\end{center}
\begin{center}
  \includegraphics[width=\columnwidth]{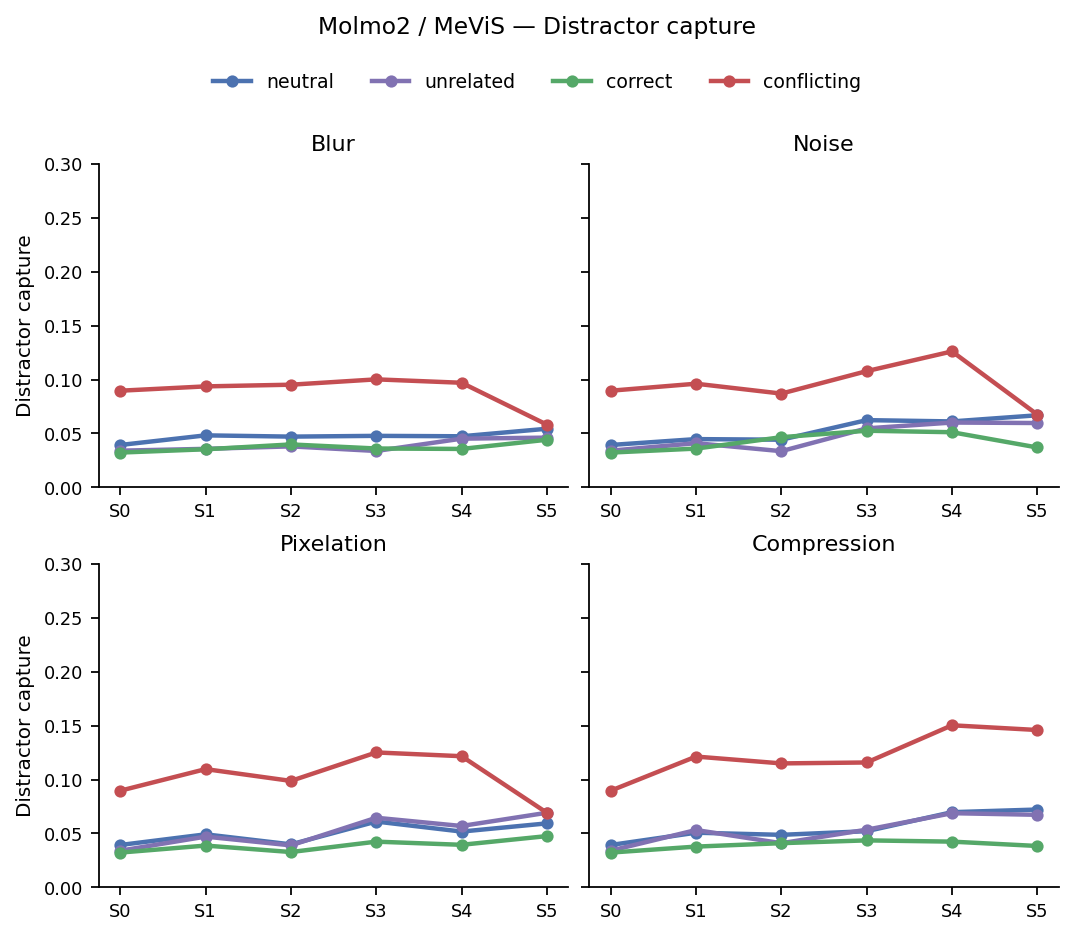}
  \captionof{figure}{Molmo2 on MeViS: distractor capture across all four perturbations, for all four text conditions.}
  \label{fig:supp_molmo2_mevis_cap}
\end{center}

\clearpage
\supptargetset{UVLTrack}{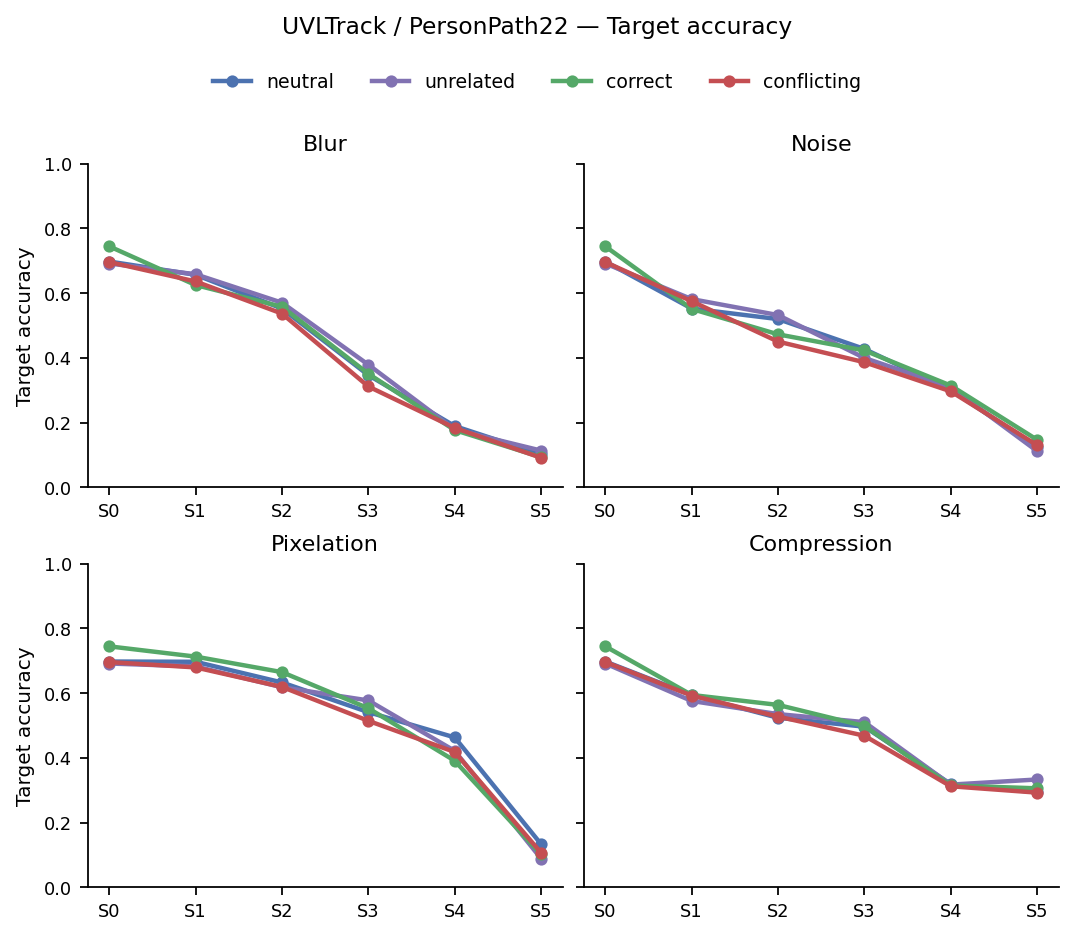}
\clearpage
\begin{center}
  \includegraphics[width=\columnwidth]{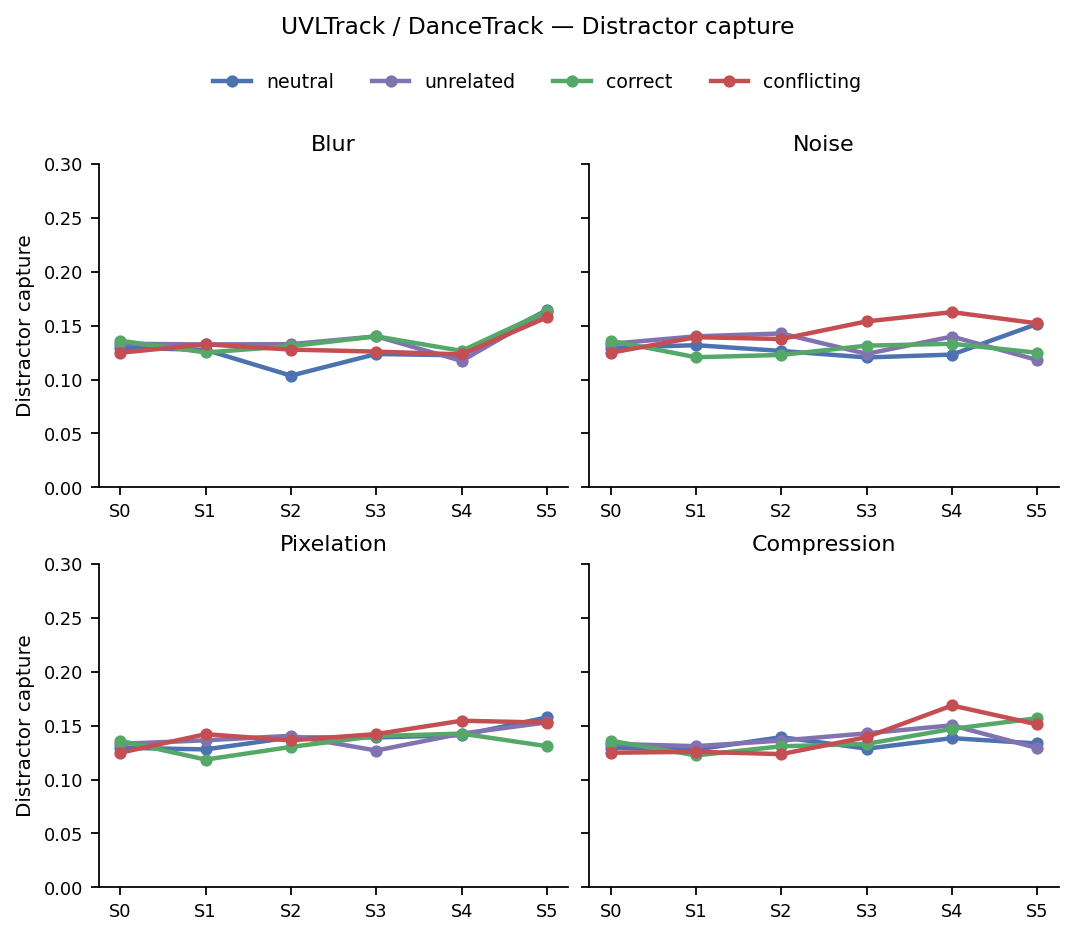}
  \captionof{figure}{UVLTrack on DanceTrack: distractor capture across all four perturbations, for all four text conditions.}
  \label{fig:supp_uvltrack_dancetrack_cap}
\end{center}
\begin{center}
  \includegraphics[width=\columnwidth]{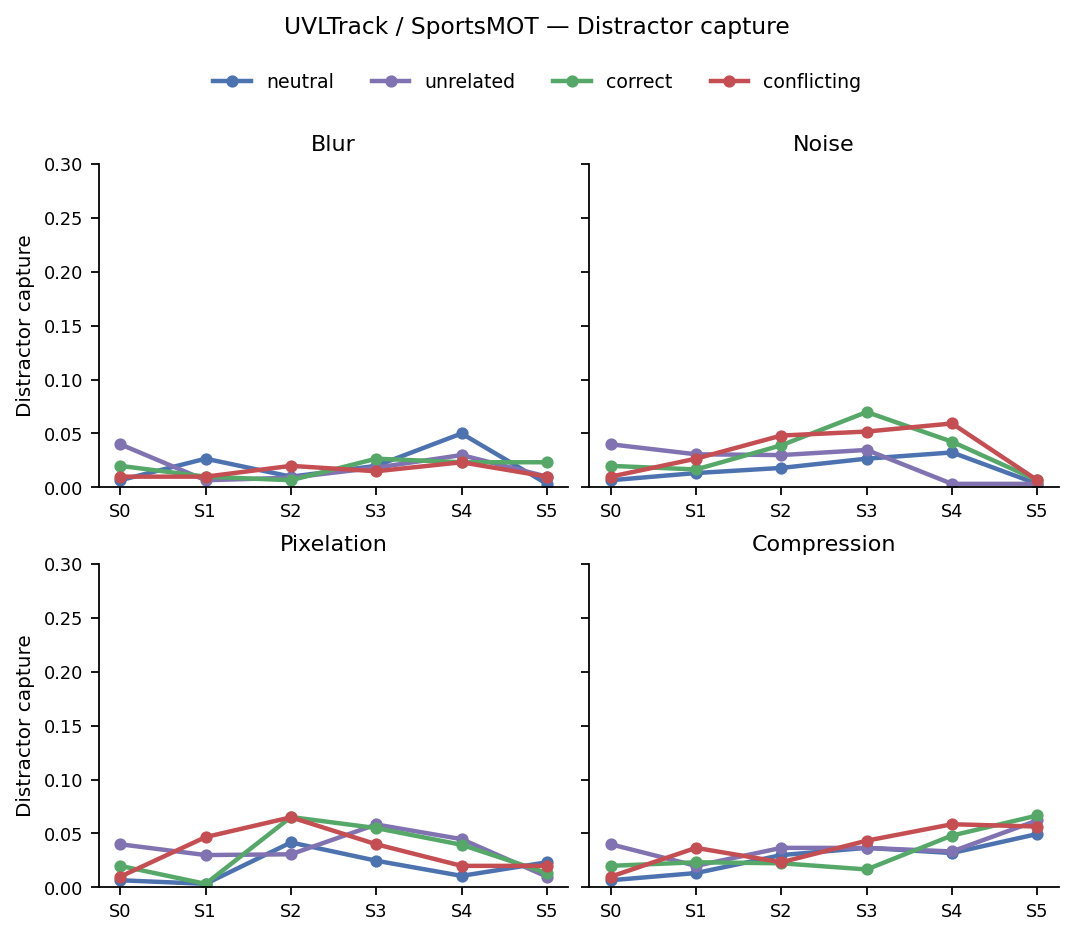}
  \captionof{figure}{UVLTrack on SportsMOT: distractor capture across all four perturbations, for all four text conditions.}
  \label{fig:supp_uvltrack_sportsmot_cap}
\end{center}
\begin{center}
  \includegraphics[width=\columnwidth]{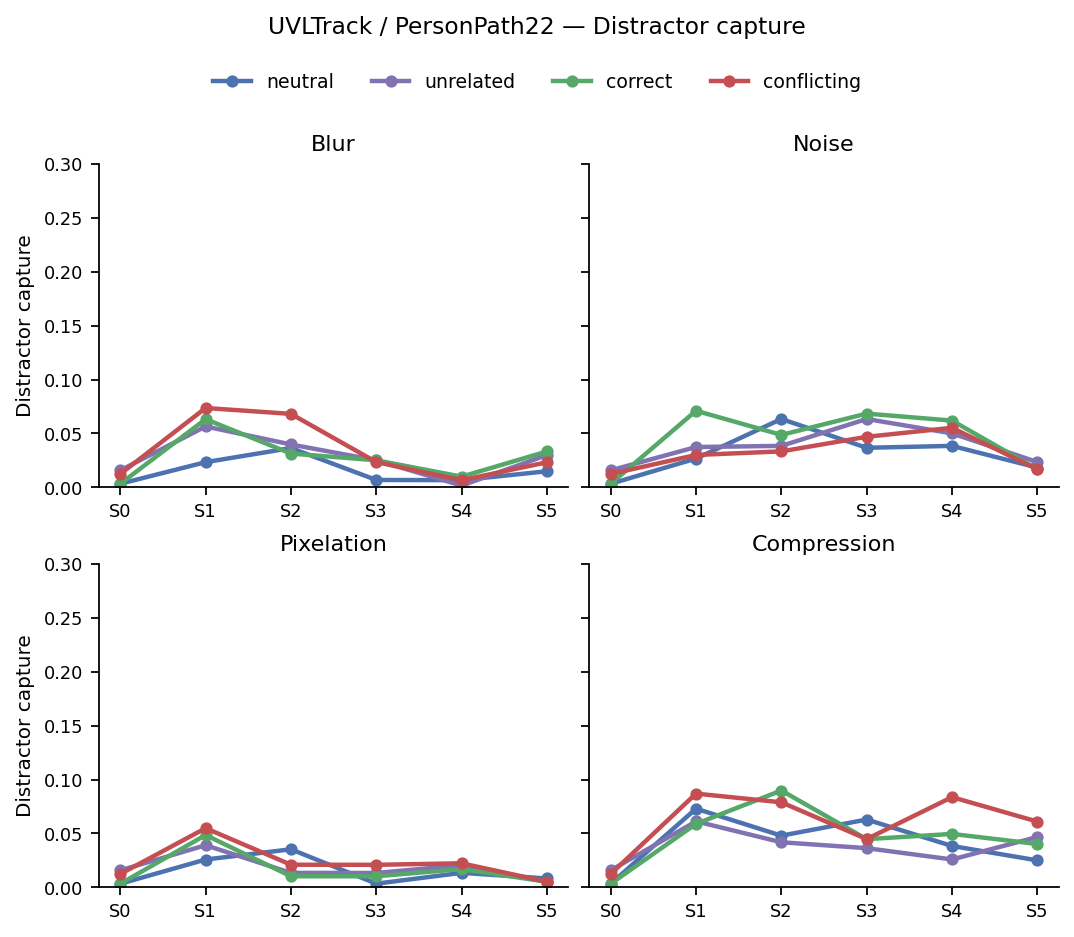}
  \captionof{figure}{UVLTrack on PersonPath22: distractor capture across all four perturbations, for all four text conditions.}
  \label{fig:supp_uvltrack_personpath22_cap}
\end{center}
\begin{center}
  \includegraphics[width=\columnwidth]{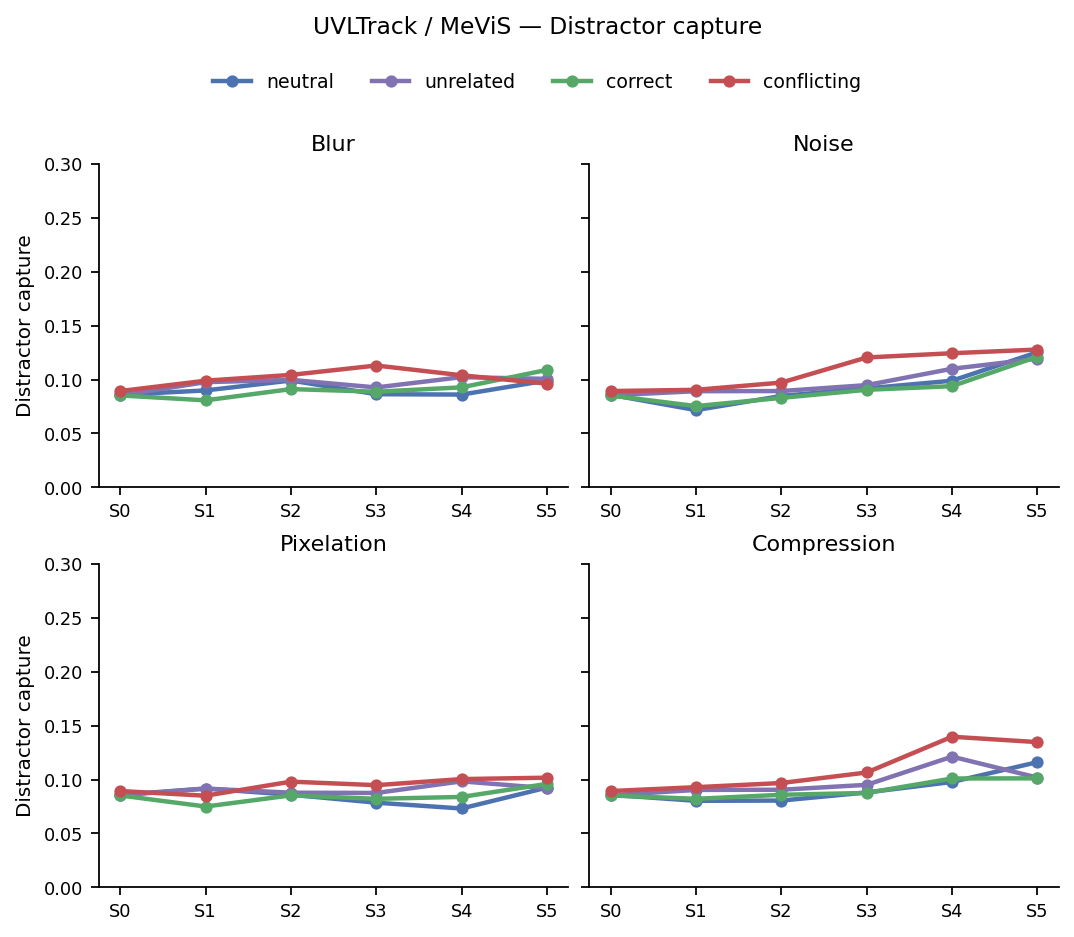}
  \captionof{figure}{UVLTrack on MeViS: distractor capture across all four perturbations, for all four text conditions.}
  \label{fig:supp_uvltrack_mevis_cap}
\end{center}

\clearpage
\supptargetset{MMTrack}{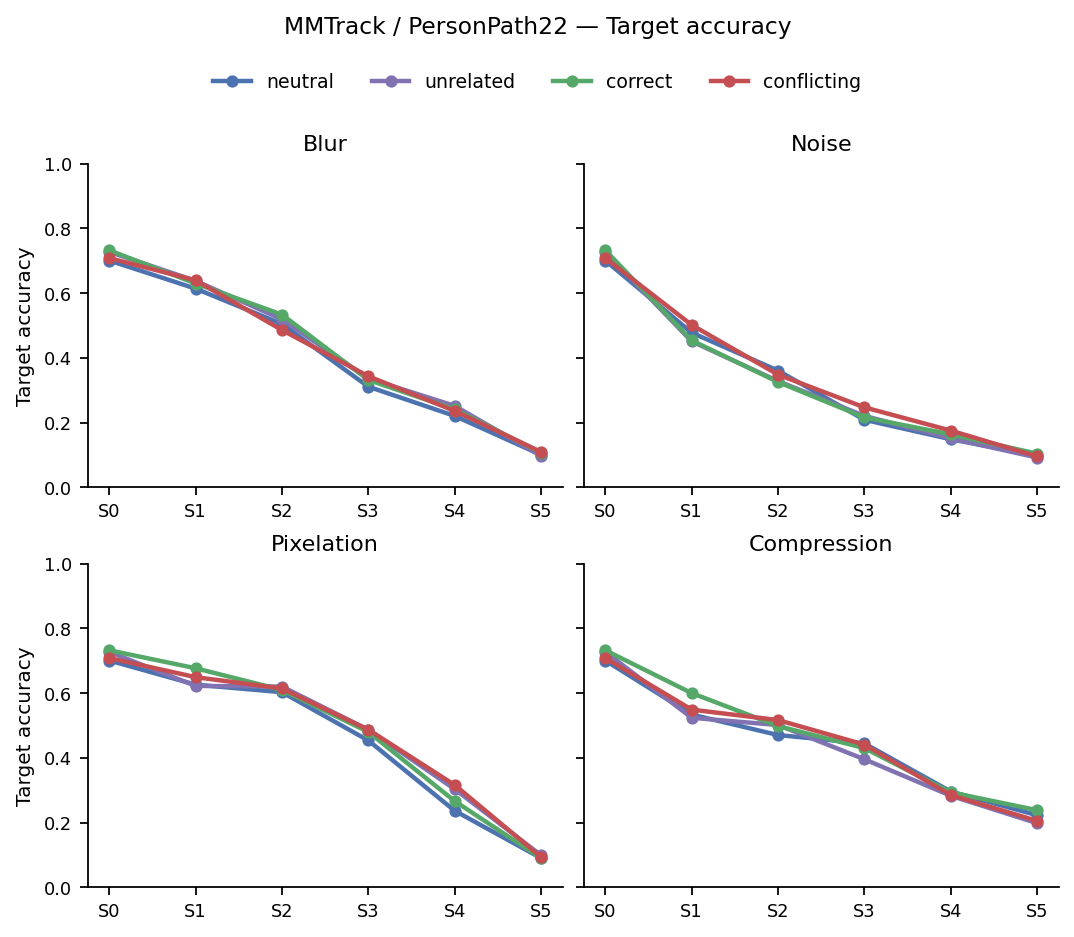}
\clearpage
\begin{center}
  \includegraphics[width=\columnwidth]{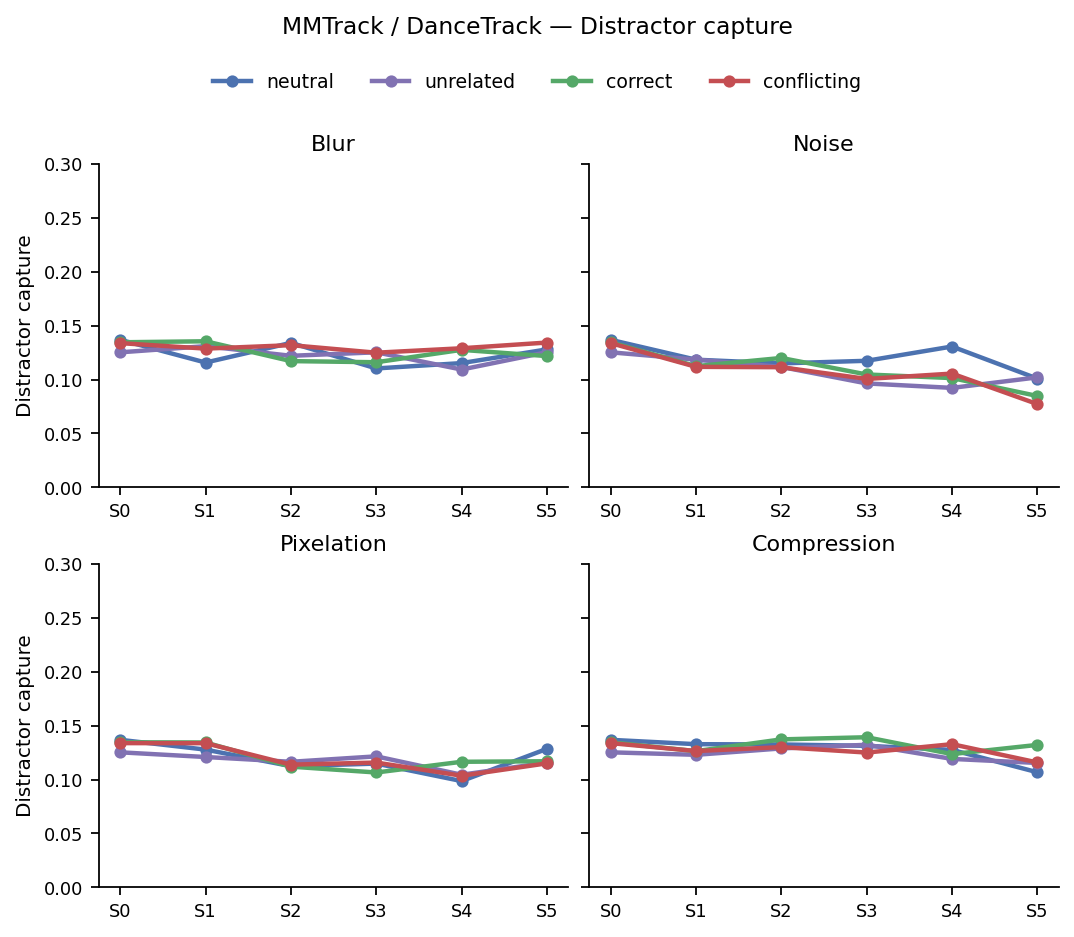}
  \captionof{figure}{MMTrack on DanceTrack: distractor capture across all four perturbations, for all four text conditions.}
  \label{fig:supp_mmtrack_dancetrack_cap}
\end{center}
\begin{center}
  \includegraphics[width=\columnwidth]{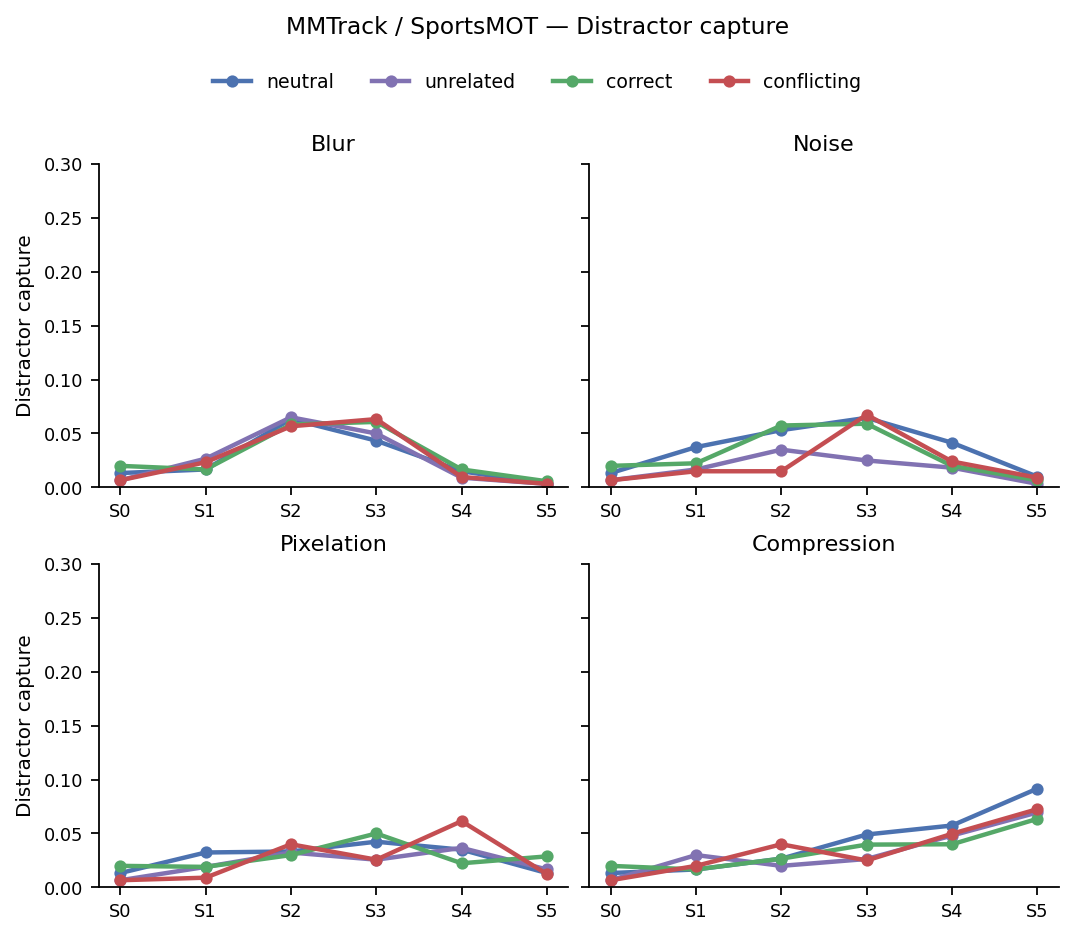}
  \captionof{figure}{MMTrack on SportsMOT: distractor capture across all four perturbations, for all four text conditions.}
  \label{fig:supp_mmtrack_sportsmot_cap}
\end{center}
\begin{center}
  \includegraphics[width=\columnwidth]{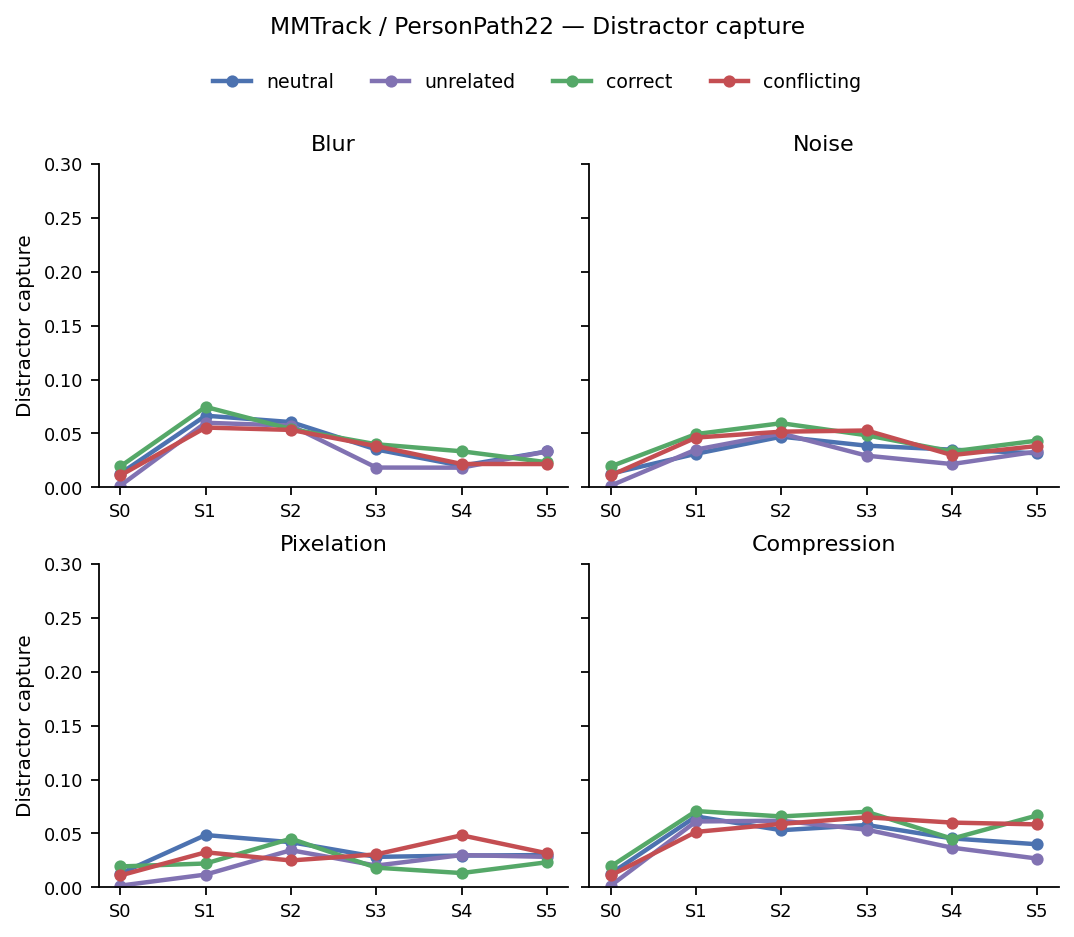}
  \captionof{figure}{MMTrack on PersonPath22: distractor capture across all four perturbations, for all four text conditions.}
  \label{fig:supp_mmtrack_personpath22_cap}
\end{center}
\begin{center}
  \includegraphics[width=\columnwidth]{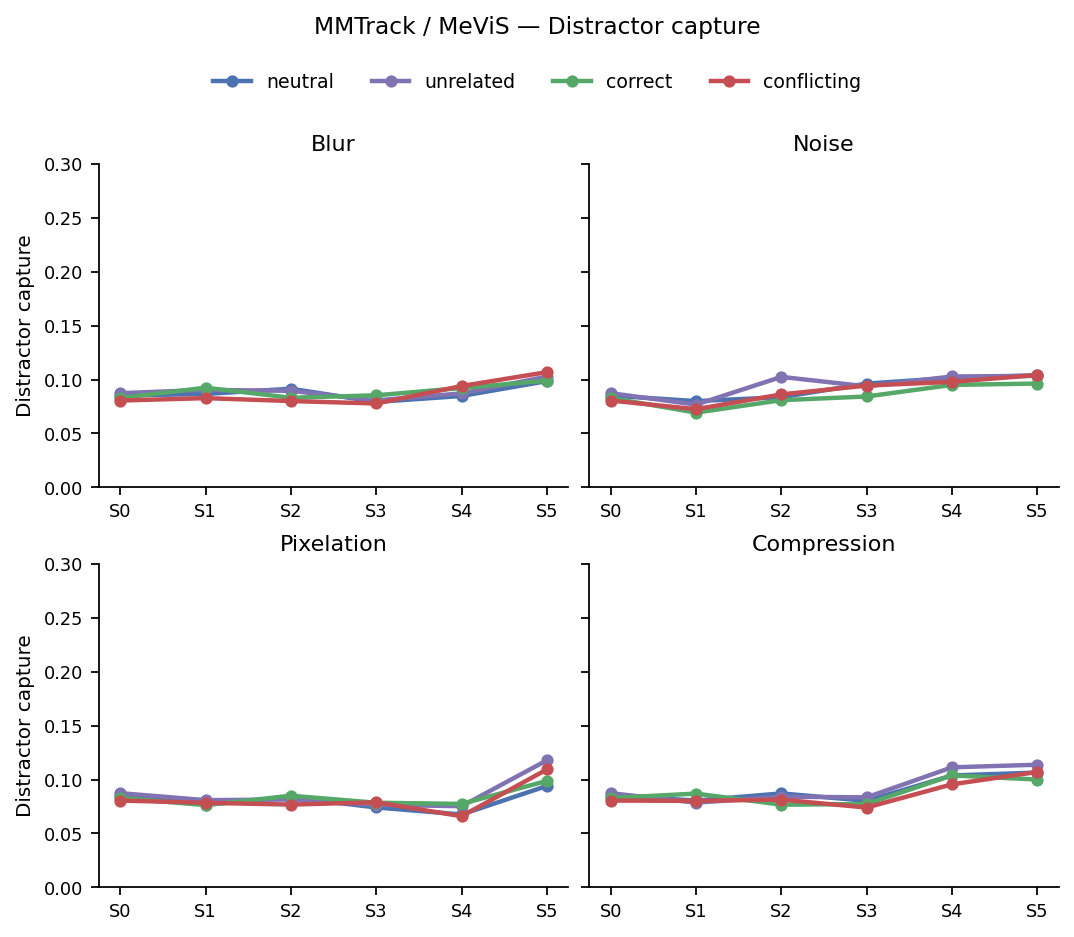}
  \captionof{figure}{MMTrack on MeViS: distractor capture across all four perturbations, for all four text conditions.}
  \label{fig:supp_mmtrack_mevis_cap}
\end{center}

\clearpage
\supptargetset{JointNLT}{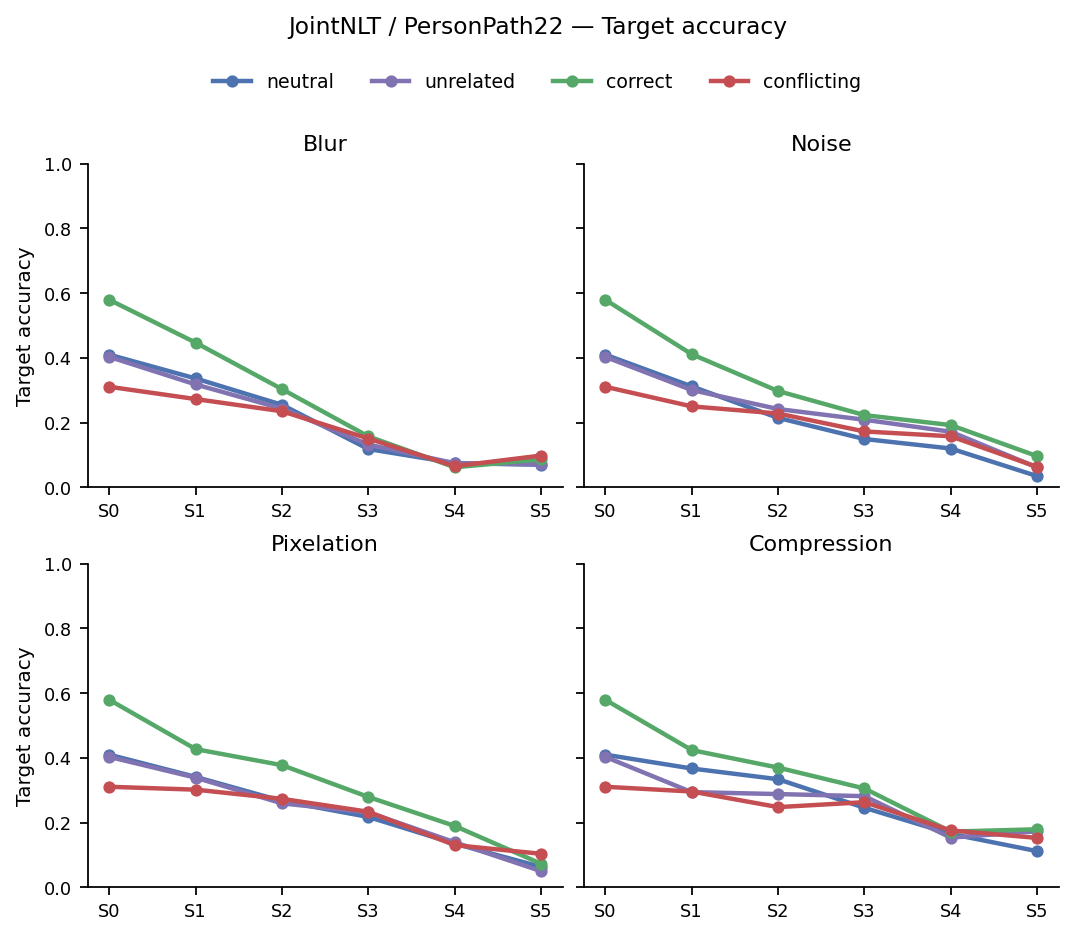}
\clearpage
\begin{center}
  \includegraphics[width=\columnwidth]{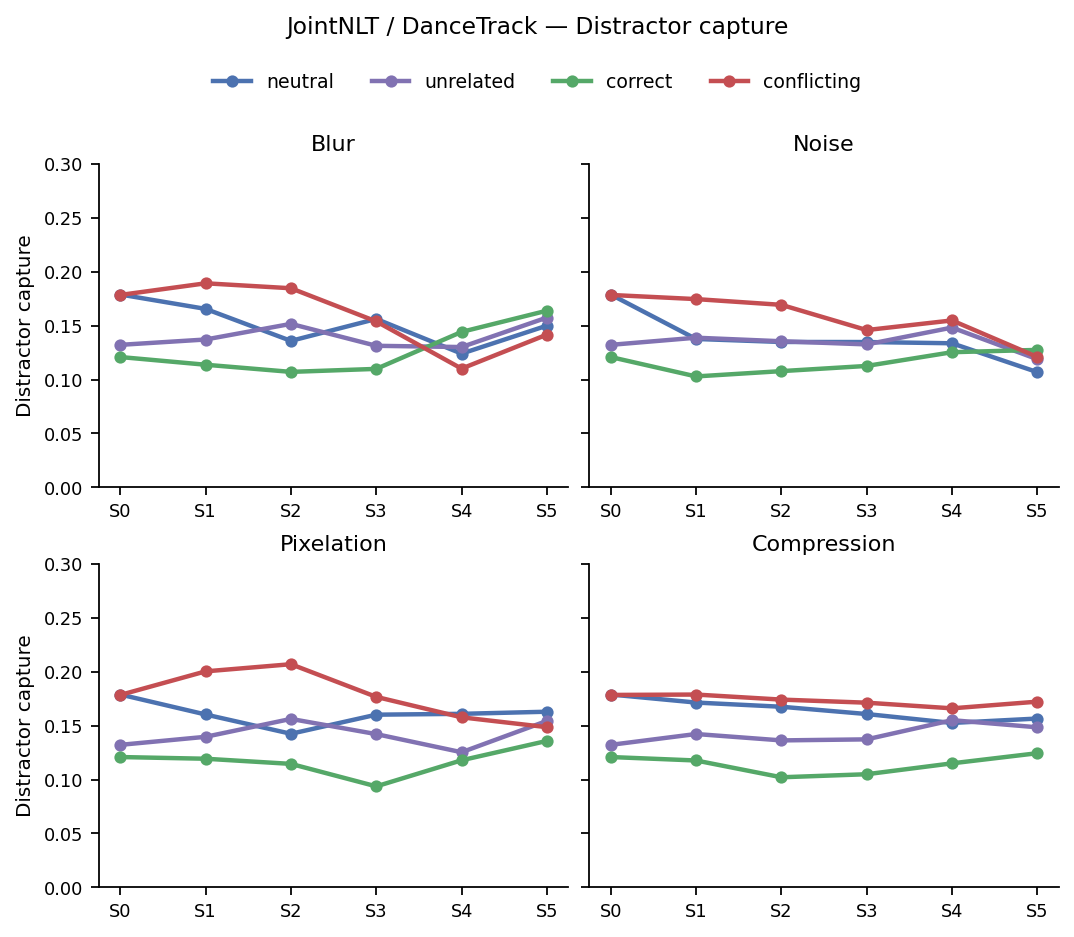}
  \captionof{figure}{JointNLT on DanceTrack: distractor capture across all four perturbations, for all four text conditions.}
  \label{fig:supp_jointnlt_dancetrack_cap}
\end{center}
\begin{center}
  \includegraphics[width=\columnwidth]{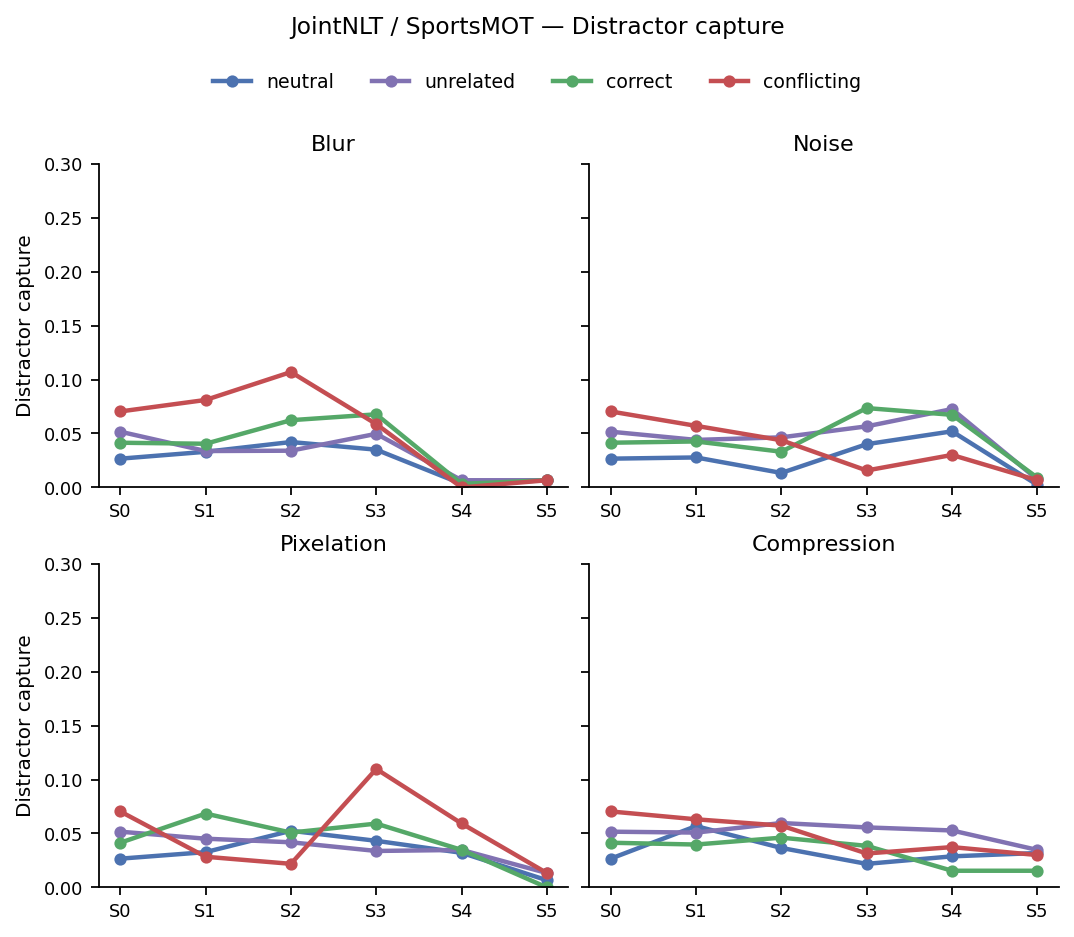}
  \captionof{figure}{JointNLT on SportsMOT: distractor capture across all four perturbations, for all four text conditions.}
  \label{fig:supp_jointnlt_sportsmot_cap}
\end{center}
\begin{center}
  \includegraphics[width=\columnwidth]{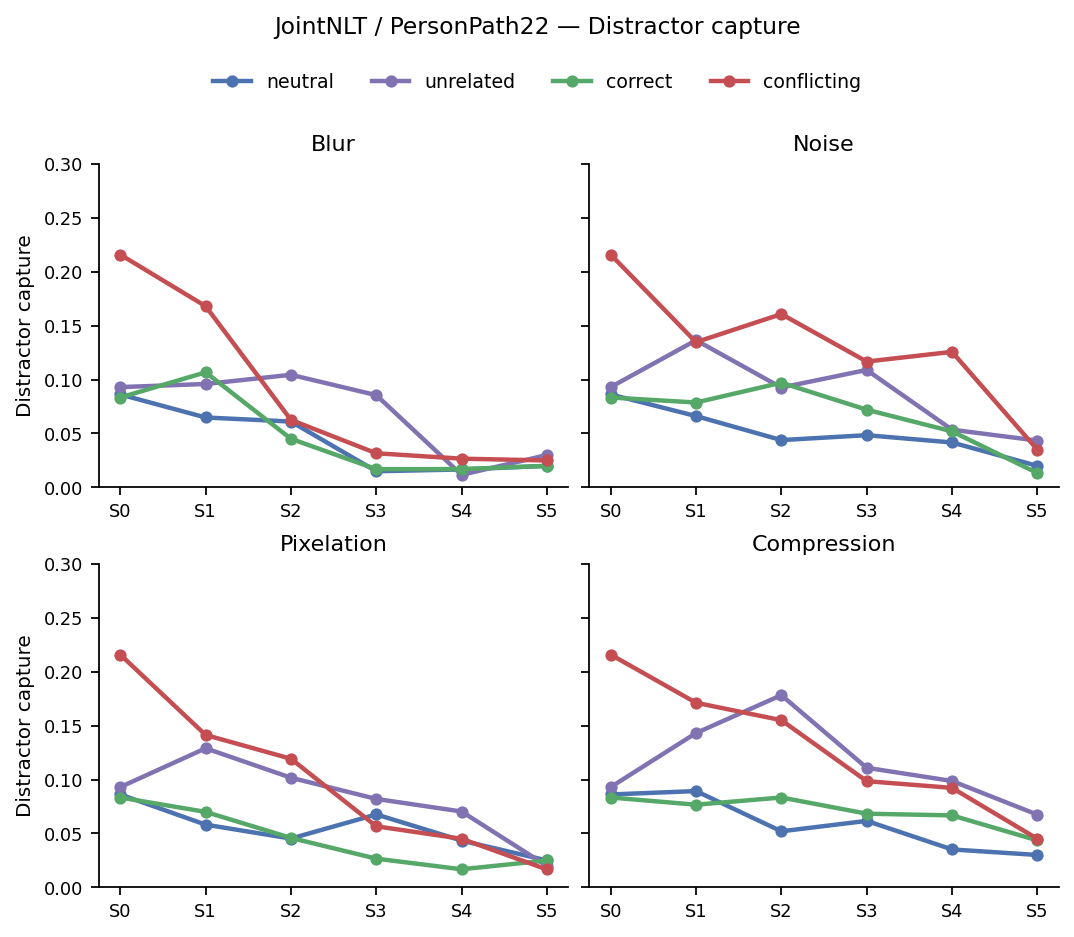}
  \captionof{figure}{JointNLT on PersonPath22: distractor capture across all four perturbations, for all four text conditions.}
  \label{fig:supp_jointnlt_personpath22_cap}
\end{center}
\begin{center}
  \includegraphics[width=\columnwidth]{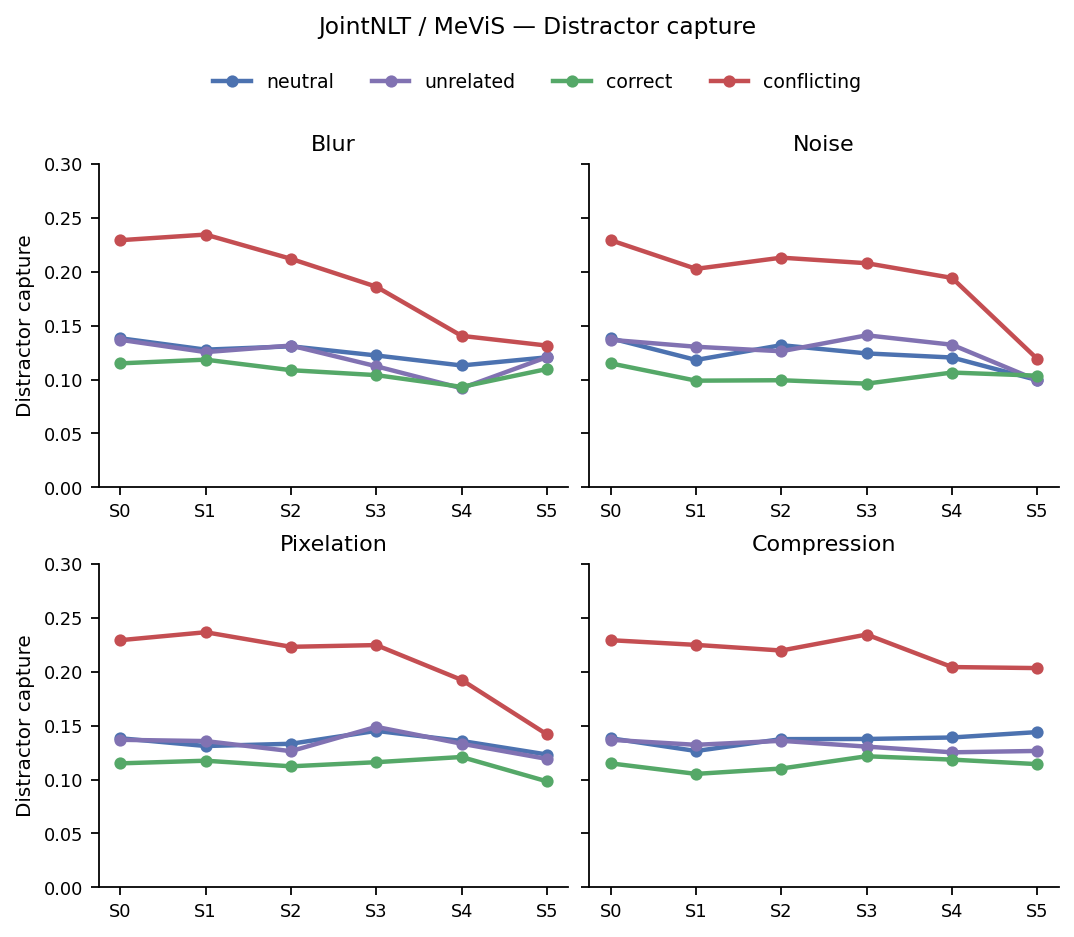}
  \captionof{figure}{JointNLT on MeViS: distractor capture across all four perturbations, for all four text conditions.}
  \label{fig:supp_jointnlt_mevis_cap}
\end{center}

\clearpage
\supptargetset{DUTrack}{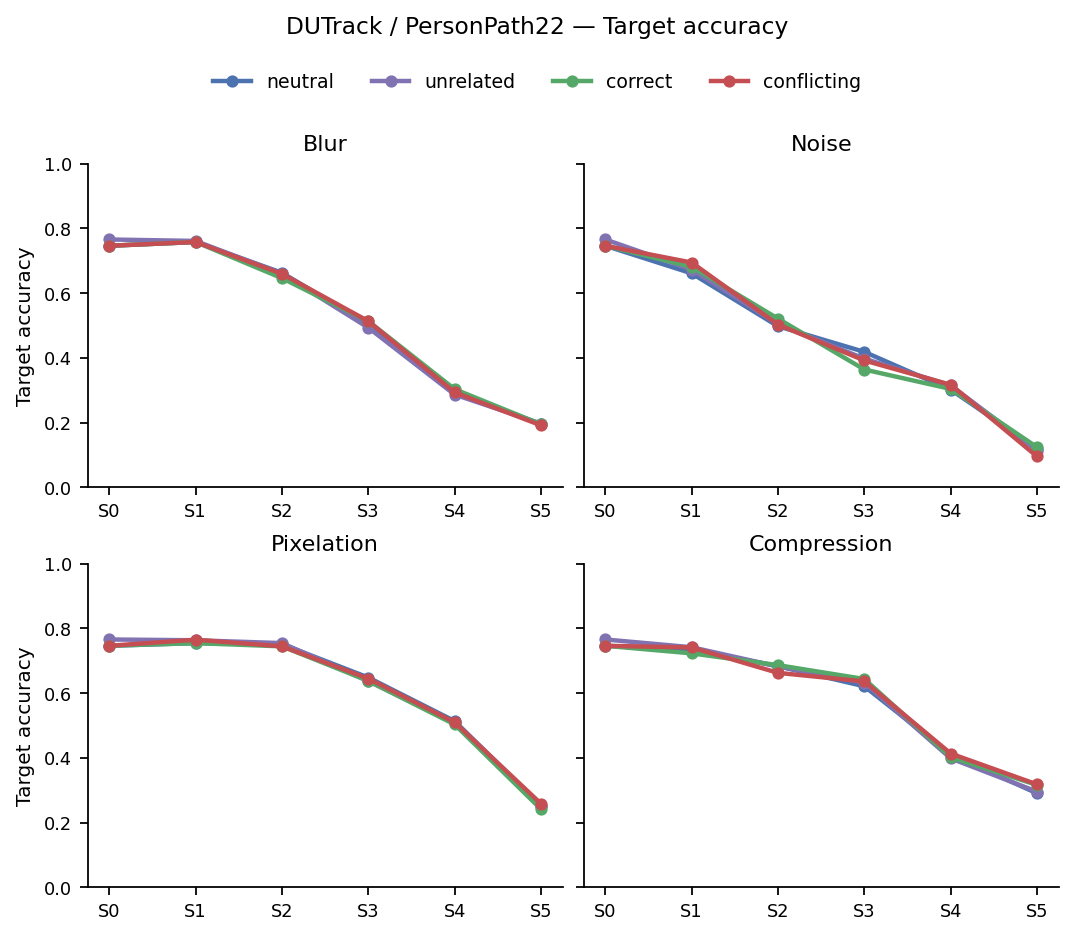}
\clearpage
\begin{center}
  \includegraphics[width=\columnwidth]{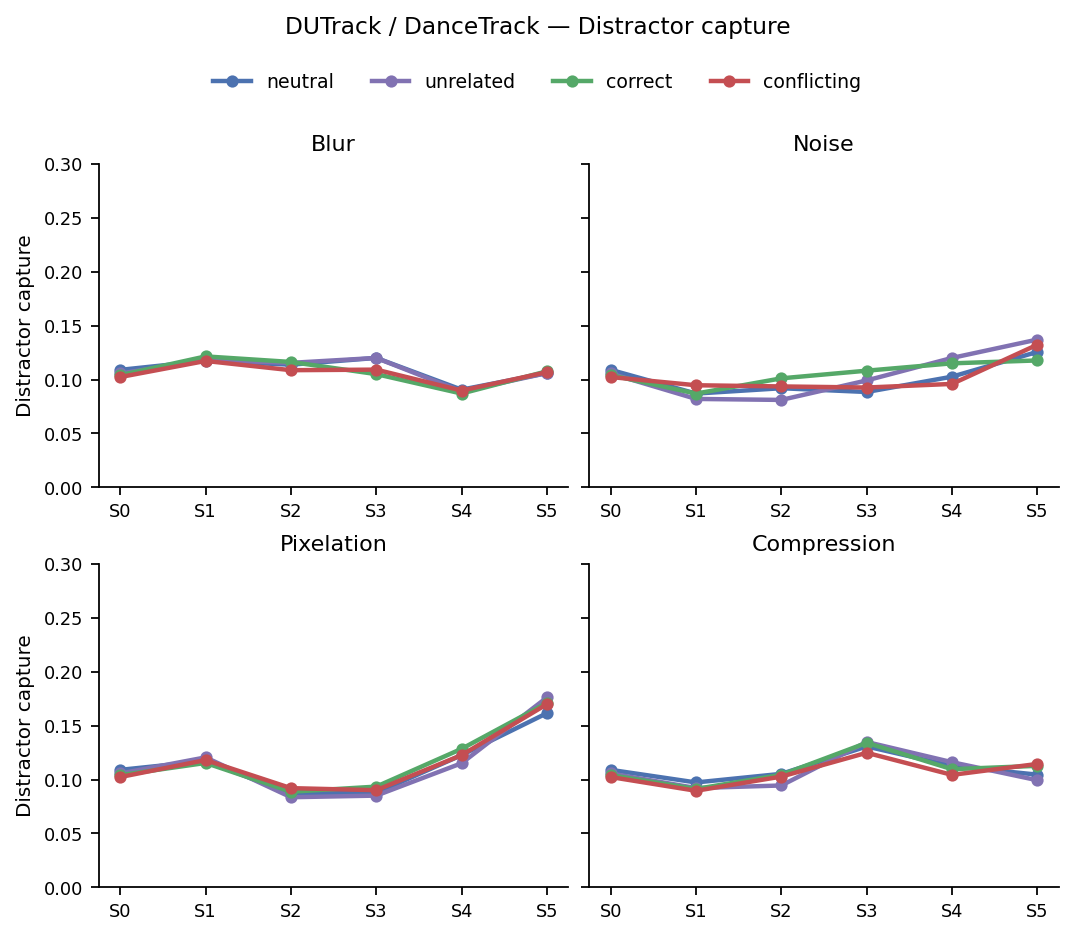}
  \captionof{figure}{DUTrack on DanceTrack: distractor capture across all four perturbations, for all four text conditions.}
  \label{fig:supp_dutrack_dancetrack_cap}
\end{center}
\begin{center}
  \includegraphics[width=\columnwidth]{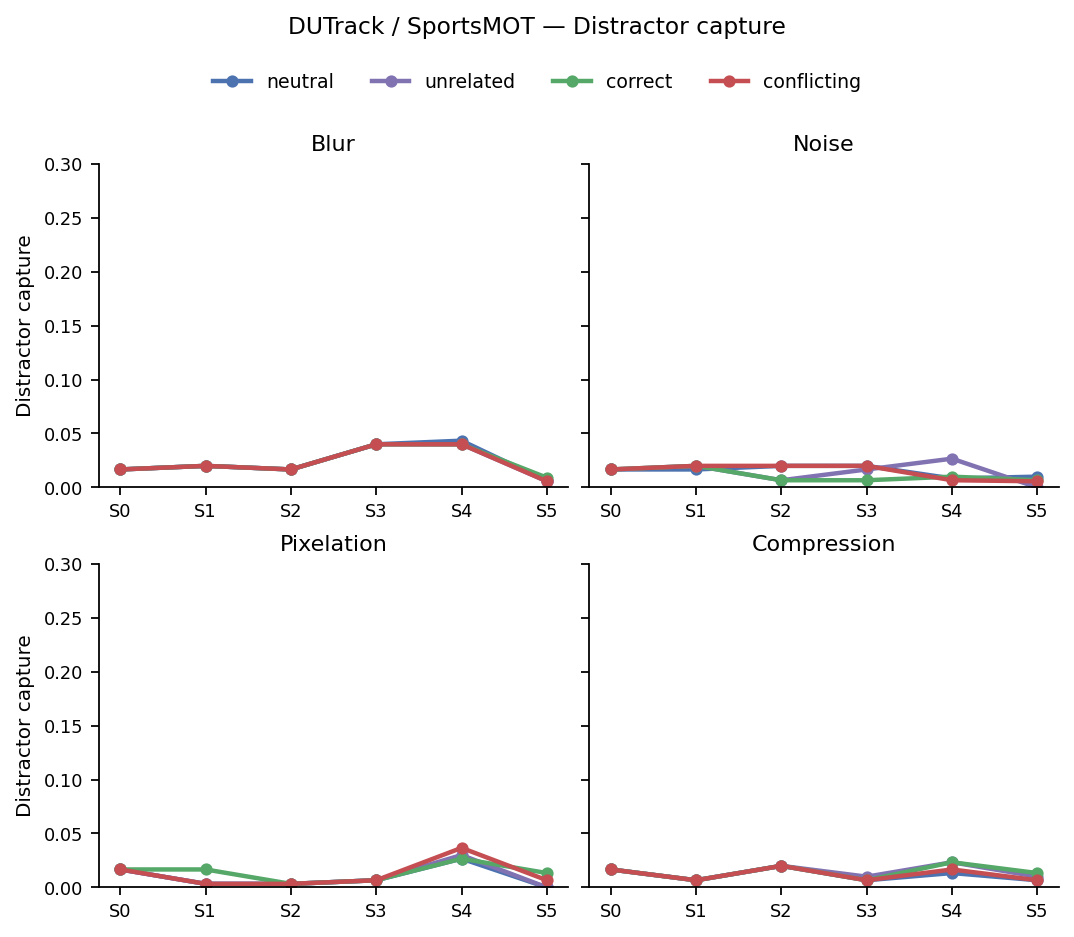}
  \captionof{figure}{DUTrack on SportsMOT: distractor capture across all four perturbations, for all four text conditions.}
  \label{fig:supp_dutrack_sportsmot_cap}
\end{center}
\begin{center}
  \includegraphics[width=\columnwidth]{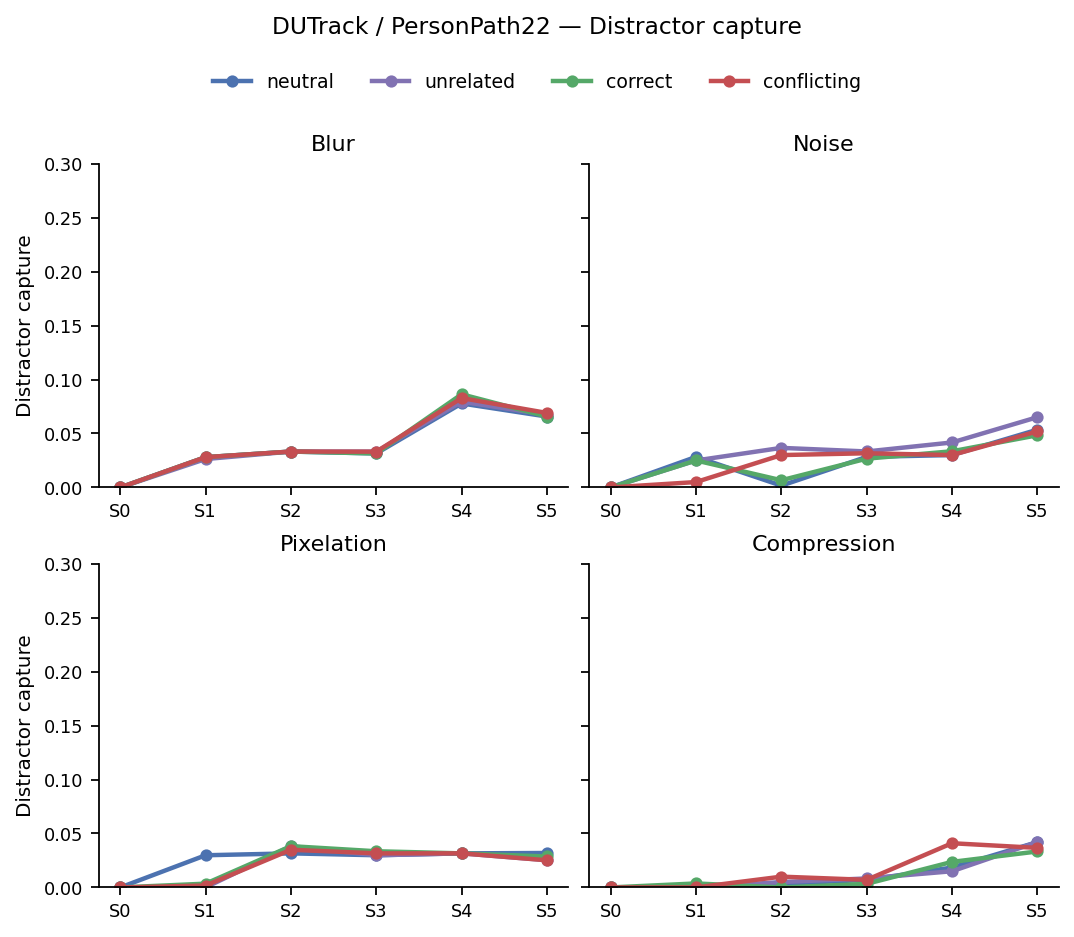}
  \captionof{figure}{DUTrack on PersonPath22: distractor capture across all four perturbations, for all four text conditions.}
  \label{fig:supp_dutrack_personpath22_cap}
\end{center}
\begin{center}
  \includegraphics[width=\columnwidth]{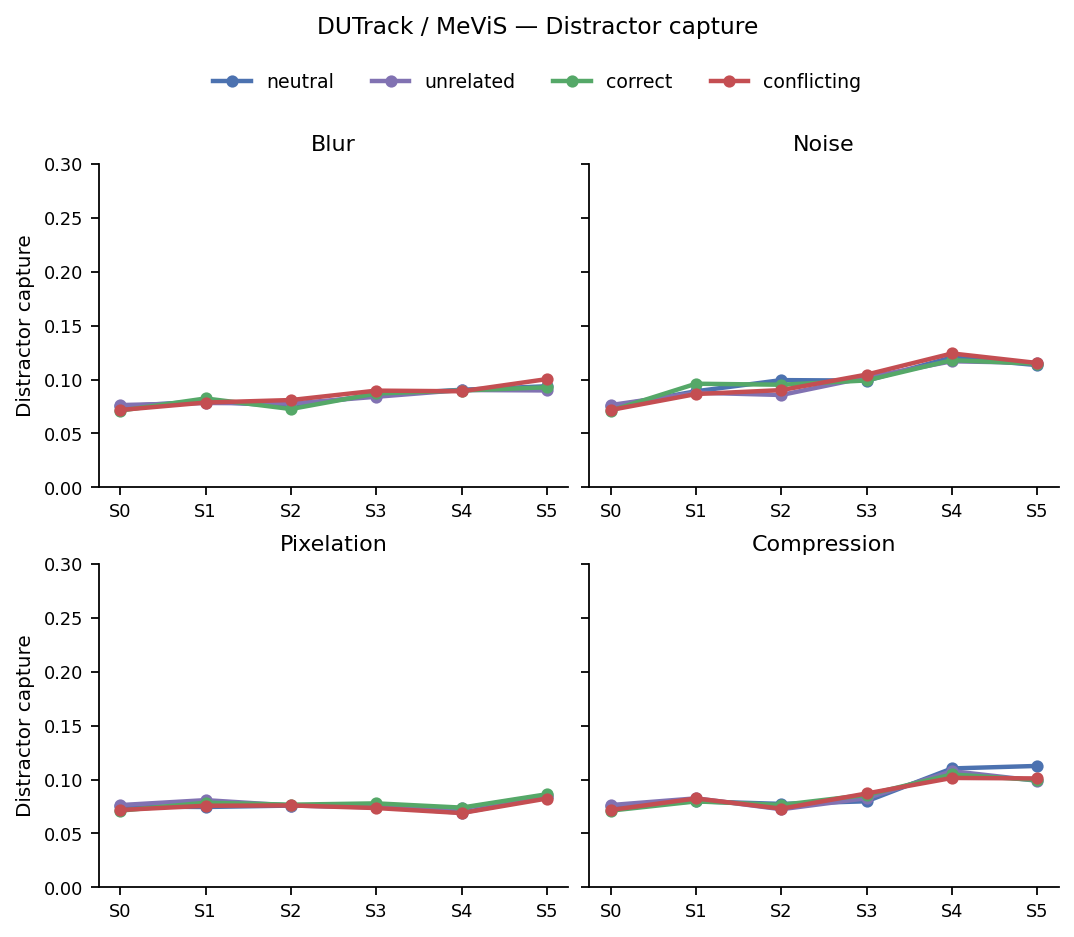}
  \captionof{figure}{DUTrack on MeViS: distractor capture across all four perturbations, for all four text conditions.}
  \label{fig:supp_dutrack_mevis_cap}
\end{center}
\clearpage

\subsection{Textual direction under degradation: full per-case statistics}
\label{sec:supp_contraction}

\begin{figure}[htbp]
  \centering
  \includegraphics[width=\columnwidth]{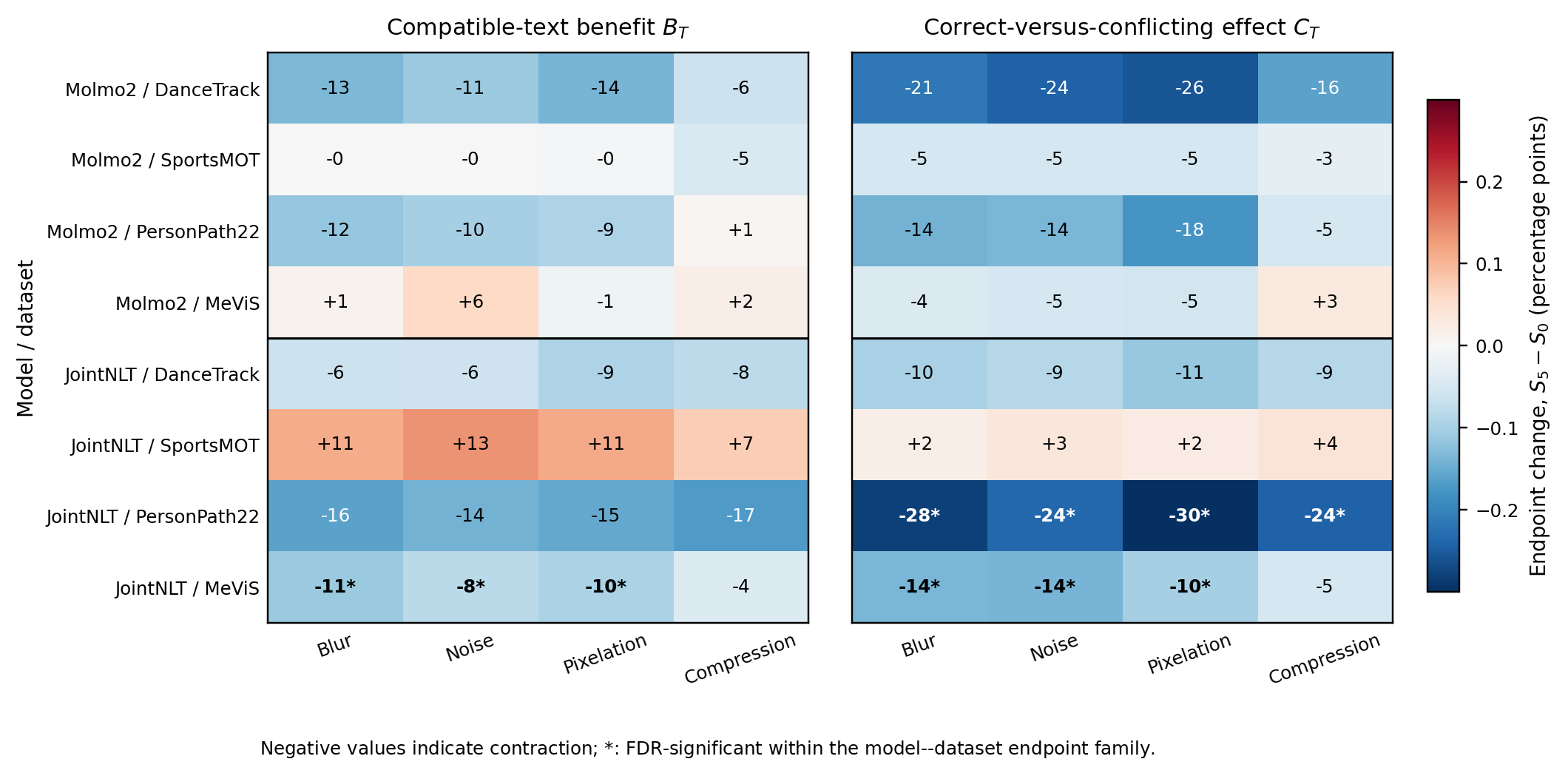}
  \caption{Endpoint change ($S_5-S_0$) in responsive-system target-accuracy
  effects; asterisks mark FDR-significant endpoint tests.}
  \label{fig:supp_contraction_heatmap}
\end{figure}

\begin{figure}[htbp]
  \centering
  \includegraphics[width=\columnwidth]{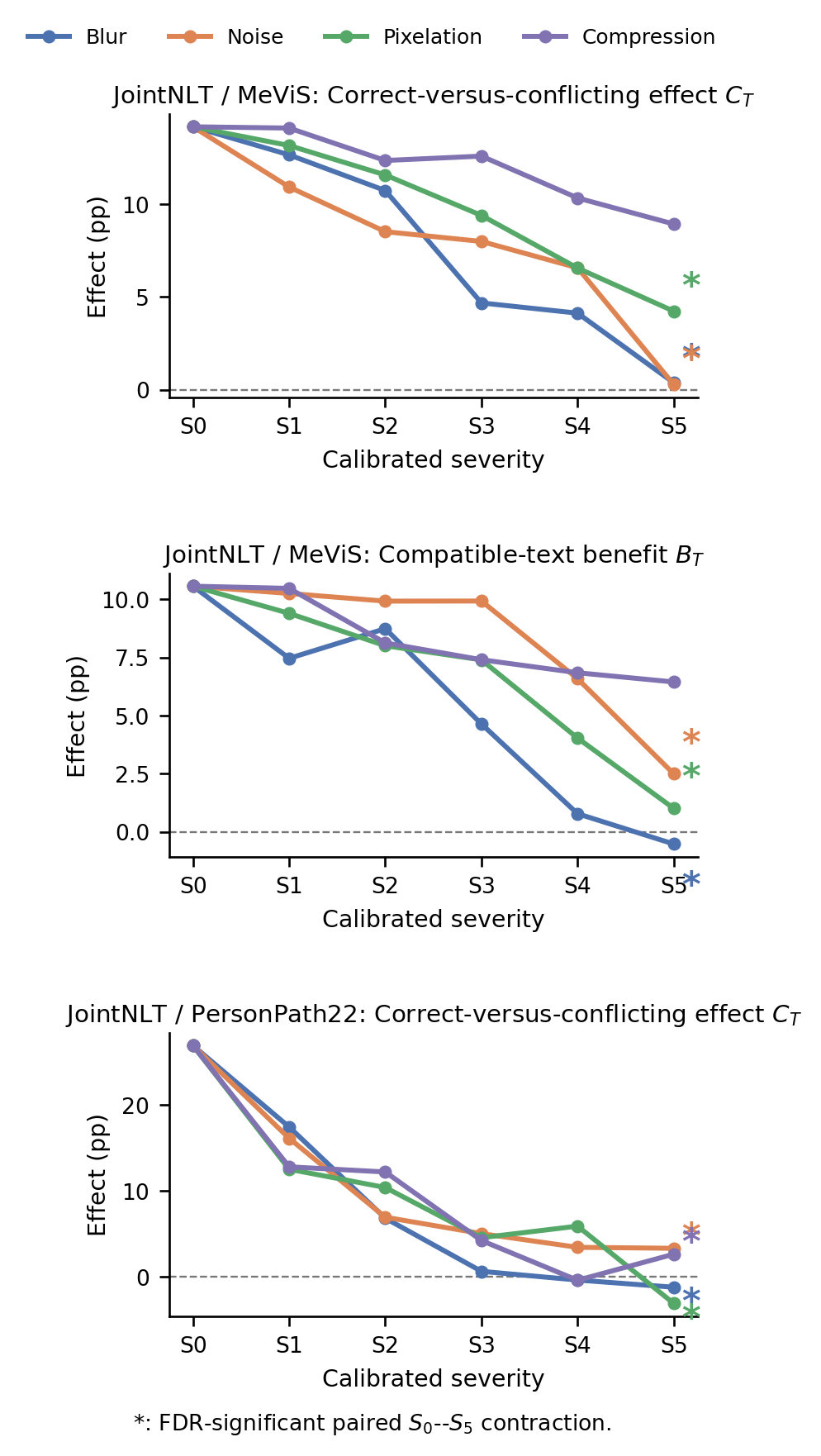}
  \caption{Full severity trajectories for the three model--dataset--effect
  cases with FDR-supported endpoint contraction.}
  \label{fig:supp_contraction_trajectories}
\end{figure}

These tables list the responsive-system effects with a CI-supported clean
input value. $S_0$ is the clean signed effect; later entries are signed
$S_5-S_0$ changes in percentage points. Asterisks mark FDR-significant
endpoint tests. Figure~\ref{fig:supp_contraction_heatmap} visualizes the
endpoint inventory, and Fig.~\ref{fig:supp_contraction_trajectories} shows
the corresponding full trajectories. Tables~\ref{tab:supp_contraction_target_accuracy}
and~\ref{tab:supp_contraction_distractor_capture} provide the numerical
target-accuracy and distractor-capture values.

\textit{Target-accuracy effects ($B_T$, $C_T$, or $N_T$).}

\begin{table}[htbp]
  \centering
  \scriptsize
  \setlength{\tabcolsep}{3pt}
  \resizebox{\columnwidth}{!}{%
  \begin{tabular}{@{}lllrrrrr@{}}
    \hline
    Model & Dataset & Eff. & $S_0$ & Blur & Noise & Pixel. & Compr. \\
    \hline
    JointNLT & MeViS        & $C_T$ & +14.2 & $-13.8^*$ & $-13.9^*$ & $-9.9^*$ & $-5.2$ \\
    JointNLT & PersonPath22 & $C_T$ & +26.9 & $-28.0^*$ & $-23.5^*$ & $-29.9^*$ & $-24.2^*$ \\
    Molmo2   & DanceTrack   & $B_T$ & +12.8 & $-13.5$ & $-11.0$ & $-13.9$ & -6.4 \\
    JointNLT & MeViS        & $B_T$ & +10.6 & $-11.1^*$ & $-8.1^*$ & $-9.6^*$ & -4.1 \\
    Molmo2   & DanceTrack   & $C_T$ & +25.3 & -21.4 & $-24.3$ & $-25.8$ & -16.0 \\
    JointNLT & MeViS        & $N_T$ & +7.8 & $-8.9^*$ & $-7.8^*$ & -5.1 & -3.4 \\
    JointNLT & DanceTrack   & $B_T$ & +10.7 & -6.4 & -6.1 & -9.4 & -7.8 \\
    JointNLT & DanceTrack   & $C_T$ & +10.9 & -9.6 & -8.6 & -11.5 & -8.5 \\
    JointNLT & PersonPath22 & $B_T$ & +17.6 & -16.1 & -14.1 & -15.3 & -17.0 \\
    JointNLT & PersonPath22 & $N_T$ & +17.0 & -15.3 & -10.8 & -16.0 & -10.2 \\
    JointNLT & SportsMOT    & $B_T$ & $-11.2^\dagger$ & +11.2 & +13.4 & +11.3 & +7.4 \\
    Molmo2   & MeViS        & $B_T$ & +3.8 & +1.1 & +5.9 & -1.4 & +1.9 \\
    Molmo2   & MeViS        & $C_T$ & +8.1 & -4.4 & -5.0 & -5.4 & +2.9 \\
    Molmo2   & DanceTrack   & $N_T$ & +7.6 & -9.8 & -6.4 & -8.6 & -6.3 \\
    \hline
  \end{tabular}%
  }
  \caption{Target-accuracy effects with a CI-supported clean-input value.
  Values after $S_0$ are signed endpoint changes; asterisks mark
  FDR-significant endpoint tests.}
  \label{tab:supp_contraction_target_accuracy}
\end{table}

\textit{Distractor-capture effects ($S_D$).}

\begin{table}[htbp]
  \centering
  \scriptsize
  \setlength{\tabcolsep}{2.8pt}
  \resizebox{\columnwidth}{!}{%
  \begin{tabular}{@{}llrrrrr@{}}
    \hline
    Model & Dataset & $S_0$ & Blur & Noise & Pixel. & Compr. \\
    \hline
    JointNLT & MeViS & +9.2 & $-8.2^*$ & $-7.3^*$ & $-7.0^*$ & -1.6 \\
    Molmo2 & DanceTrack & +14.1 & -7.6 & $-11.0$ & $-12.8$ & -6.1 \\
    Molmo2 & MeViS & +5.6 & -4.4 & -4.8 & $-5.6^*$ & +2.3 \\
    Molmo2 & PersonPath22 & +11.3 & -12.6 & -11.8 & -11.9 & -3.0 \\
    \hline
  \end{tabular}%
  }
  \caption{Distractor-capture effects with a CI-supported clean-input value.
  Values after $S_0$ are signed endpoint changes; asterisks mark
  FDR-significant endpoint tests.}
  \label{tab:supp_contraction_distractor_capture}
\end{table}
\FloatBarrier

\subsection{Absolute performance and floor/ceiling diagnostics}
\label{sec:supp_absolute_performance}

Table~\ref{tab:supp_absolute_performance} reports absolute target accuracy,
distractor capture, and other/absent rate under correct text at $S_0$ and
$S_5$.

\begin{table}[htbp]
  \centering
  \scriptsize
  \setlength{\tabcolsep}{3pt}
  \begin{tabular}{@{}lrrrrrr@{}}
    \toprule
     & \multicolumn{3}{c}{$S_0$} & \multicolumn{3}{c}{$S_5$ (pooled, 4 perturbations)} \\
    \cmidrule(lr){2-4} \cmidrule(lr){5-7}
    Model & $A_T$ & $A_D$ & other/absent & $A_T$ & $A_D$ & other/absent \\
    \midrule
    Molmo2   & 67.9\% & 2.4\% & 29.7\% & 37.4\% & 3.7\% & 58.8\% \\
    JointNLT & 43.6\% & 9.0\% & 47.4\% & 19.3\% & 7.3\% & 73.4\% \\
    UVLTrack & 64.4\% & 6.1\% & 29.5\% & 27.4\% & 7.6\% & 65.0\% \\
    MMTrack  & 62.9\% & 6.4\% & 30.7\% & 24.2\% & 7.0\% & 68.8\% \\
    DUTrack  & 68.1\% & 4.8\% & 27.1\% & 33.2\% & 6.6\% & 60.2\% \\
    \bottomrule
  \end{tabular}
  \caption{Correct-text outcome rates, pooled across datasets at $S_0$ and
  across datasets and perturbations at $S_5$. ``other/absent'' combines the
  two non-target, non-distractor labels.}
  \label{tab:supp_absolute_performance}
\end{table}
\subsection{Model internals}
\label{sec:supp_internals}

\subsubsection{Saved states and hook semantics}

Table~\ref{tab:supp_hook_map} records the clean-MeViS hook map and pooled
state semantics. Language states use [CLS] tokens or sentence vectors;
visual states pool search tokens in target or distractor regions.

\begin{center}
  \centering
  \scriptsize
  \setlength{\tabcolsep}{3pt}
  \begin{tabular}{@{}lp{6.3cm}@{}}
    \toprule
    Model & Hooked modules/stages and stored pooled semantics \\
    \midrule
    Molmo2 & Decoder inputs $0$--$36$ (post-connector input through final normalized decoder state): description-token state before coordinate generation and target/distractor visual states pooled over mask-covered image tokens. \\
    JointNLT & BERT layers $0$--$11$ under \texttt{language\_backbone}; \texttt{text\_proj}; six \texttt{vl\_joint\_trans} encoder blocks; target decoder; box-head input. We retain BERT/projected/fused [CLS], target/distractor search-region means, target query, and the spatially averaged pre-box-head feature. \\
    UVLTrack & \texttt{backbone.bert} embeddings/layers $0$--$5$; ViT blocks $0$--$11$; backbone output. We retain text [CLS], target/distractor search-region means, and final text/search states before the box head. \\
    MMTrack & Text encoder and \texttt{text\_adj}; visual blocks; \texttt{bottleneck}; \texttt{vl\_fusion}; coordinate decoder and predictor. We retain [CLS]/sentence states, target/distractor search-region means, and coordinate-query states through localization. \\
    DUTrack & \texttt{backbone.descript\_embedding}; first joint-block input; every text-reachable joint block; backbone output. We retain description/fusion-input [CLS], target/distractor search-region means, and the final visual state before temporal weighting/CENTER localization. \\
    \bottomrule
  \end{tabular}
  \captionof{table}{Model-specific hook map and stored pooled states.}
  \label{tab:supp_hook_map}
\end{center}

\subsubsection{Full-depth Text BA sweeps}

Tables~\ref{tab:supp_jointnlt_layerwise}--\ref{tab:supp_molmo_layerwise}
give the complete clean-input Text BA sweeps for JointNLT and Molmo2.

\begin{center}
  \centering
  \scriptsize
  \setlength{\tabcolsep}{4pt}
  \begin{tabular}{@{}lrr@{}}
    \toprule
    Stage & Text BA & $q$ \\
    \midrule
    text encoder L0  & $0.602$ & $.002$ \\
    text encoder L1  & $0.525$ & $.098$ \\
    text encoder L2  & $0.509$ & $.316$ \\
    text encoder L3  & $0.516$ & $.202$ \\
    text encoder L4  & $0.531$ & $.040$ \\
    text encoder L5  & $0.516$ & $.196$ \\
    text encoder L6  & $0.566$ & $.002$ \\
    text encoder L7  & $0.490$ & $.760$ \\
    text encoder L8  & $0.569$ & $.002$ \\
    text encoder L9  & $0.526$ & $.067$ \\
    text encoder L10 & $0.544$ & $.006$ \\
    text encoder L11 & $0.575$ & $.002$ \\
    text projection  & $0.563$ & $.002$ \\
    fusion L0 text   & $0.566$ & $.002$ \\
    fusion L0 $T{-}D$      & $0.518$ & $.196$ \\
    fusion L1 text         & $0.585$ & $.002$ \\
    fusion L1 $T{-}D$      & $0.535$ & $.049$ \\
    fusion L2 text         & $0.577$ & $.002$ \\
    fusion L2 $T{-}D$      & $0.690$ & $.002$ \\
    fusion L3 text         & $0.576$ & $.002$ \\
    fusion L3 $T{-}D$      & $0.699$ & $.002$ \\
    fusion L4 text         & $0.601$ & $.002$ \\
    fusion L4 $T{-}D$      & $0.673$ & $.002$ \\
    fusion L5 text         & $0.553$ & $.006$ \\
    fusion L5 $T{-}D$      & $0.685$ & $.002$ \\
    confirmatory late fusion $T{-}D$ & $\mathbf{0.695}$ & $.002$ \\
    pre-loc.\ target query & $0.519$ & $.202$ \\
    pre-loc.\ box feature  & $0.573$ & $.002$ \\
    \bottomrule
  \end{tabular}
  \captionof{table}{JointNLT correct-versus-conflicting Text BA at every
  hooked stage ($0.5$ is chance); $q$ is FDR-corrected over the sweep.
  Bold denotes the decision-near stage used in the main paper.}
  \label{tab:supp_jointnlt_layerwise}
\end{center}

\begin{center}
  \centering
  \scriptsize
  \setlength{\tabcolsep}{4pt}
  \begin{tabular}{@{}lrr@{}}
    \toprule
    Layer & BA & $q$ \\
    \midrule
    L0  & $0.499$ & $1.000$ \\
    L1  & $0.501$ & $.567$  \\
    L2  & $0.497$ & $.688$  \\
    L3  & $0.492$ & $.702$  \\
    L4  & $0.486$ & $.797$  \\
    L5  & $0.510$ & $.441$  \\
    L6  & $0.508$ & $.459$  \\
    L7  & $0.499$ & $.688$  \\
    L8  & $0.490$ & $.709$  \\
    L9  & $0.492$ & $.701$  \\
    L10 & $0.510$ & $.441$  \\
    L11 & $0.506$ & $.441$  \\
    L12 & $0.505$ & $.554$  \\
    L13 & $0.521$ & $.275$  \\
    L14 & $0.519$ & $.370$  \\
    L15 & $0.523$ & $.252$  \\
    L16 & $0.514$ & $.441$  \\
    L17 & $0.545$ & $.053$  \\
    L18 & $0.543$ & $.057$  \\
    L19 & $0.554$ & $.033$ \\
    L20 & $0.561$ & $.033$ \\
    L21 & $\mathbf{0.634}$ & $.015$ \\
    L22 & $0.629$ & $.015$ \\
    L23 & $0.614$ & $.015$ \\
    L24 & $0.594$ & $.015$ \\
    L25 & $0.569$ & $.015$ \\
    L26 & $0.565$ & $.015$ \\
    L27 & $0.567$ & $.015$ \\
    L28 & $0.569$ & $.015$ \\
    L29 & $0.563$ & $.015$ \\
    L30 & $0.556$ & $.025$ \\
    L31 & $0.558$ & $.015$ \\
    L32 & $0.563$ & $.015$ \\
    L33 & $0.556$ & $.025$ \\
    L34 & $0.550$ & $.057$ \\
    L35 & $0.565$ & $.015$ \\
    L36 & $0.574$ & $.015$ \\
    \bottomrule
  \end{tabular}
  \captionof{table}{Molmo2 correct-versus-conflicting Text BA at all saved
  coordinate-query decoder depths ($0.5$ is chance); $q$ is FDR-corrected
  over the 37-layer sweep. Bold denotes the peak.}
  \label{tab:supp_molmo_layerwise}
\end{center}

\subsubsection{JointNLT across visual degradation}
\label{sec:supp_jointnlt_degraded_internals}

\begin{figure*}[p]
  \centering
  \includegraphics[width=.94\textwidth]{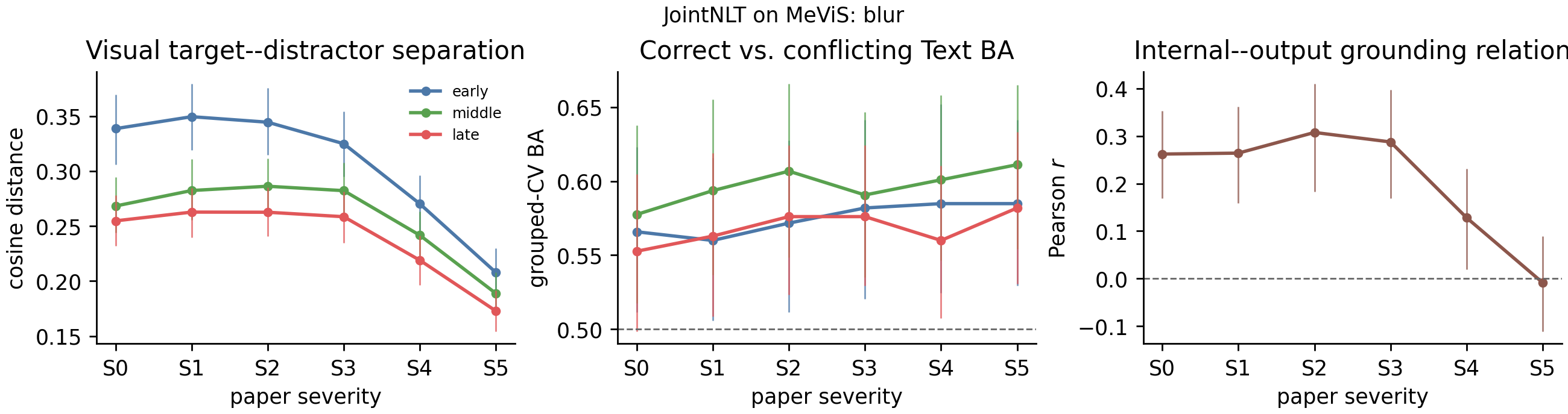}
  \caption{JointNLT MeViS internal trajectories under blur. Panels show
  visual separation, Text BA, and grounding-to-output association; error
  bars are 95\% source-video bootstrap CIs.}
  \label{fig:supp_jointnlt_degradation_blur}
\end{figure*}

\begin{figure*}[p]
  \centering
  \includegraphics[width=.94\textwidth]{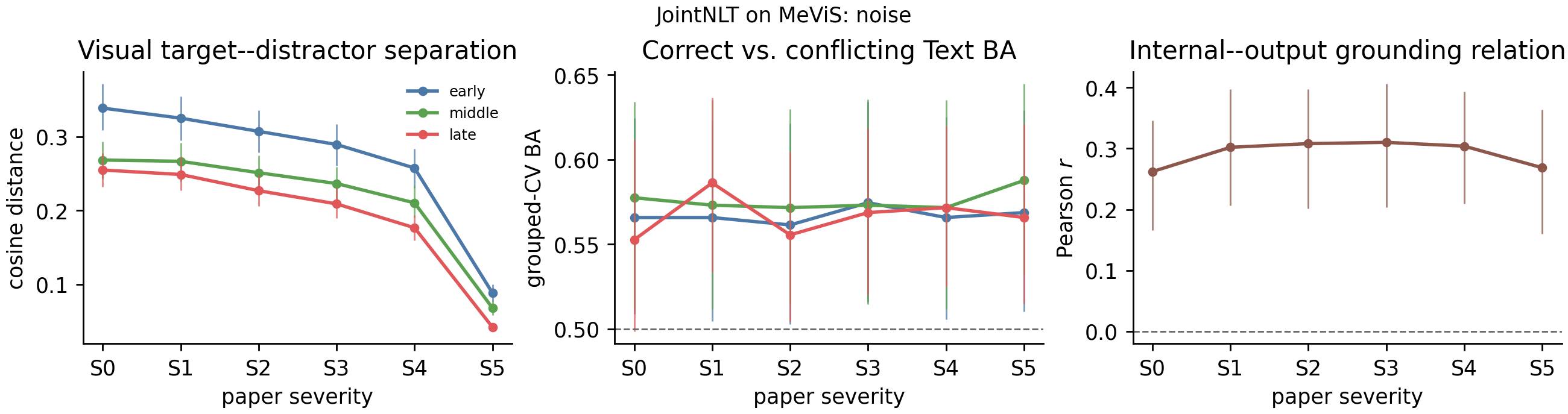}
  \caption{JointNLT MeViS internal trajectories under noise.}
  \label{fig:supp_jointnlt_degradation_noise}
\end{figure*}

\begin{figure*}[p]
  \centering
  \includegraphics[width=.94\textwidth]{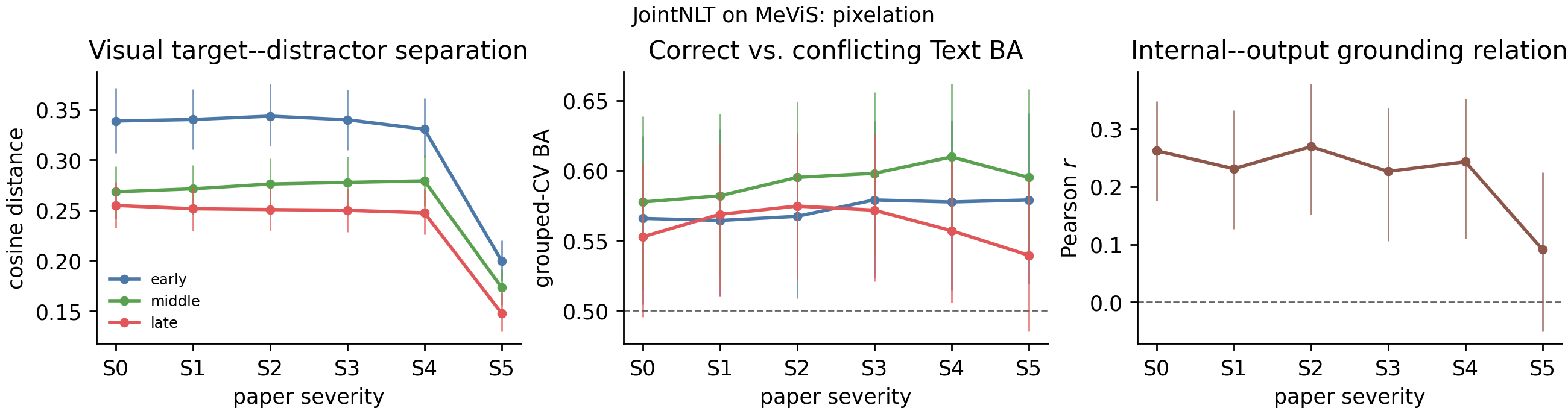}
  \caption{JointNLT MeViS internal trajectories under pixelation.}
  \label{fig:supp_jointnlt_degradation_pixelation}
\end{figure*}

\begin{figure*}[p]
  \centering
  \includegraphics[width=.94\textwidth]{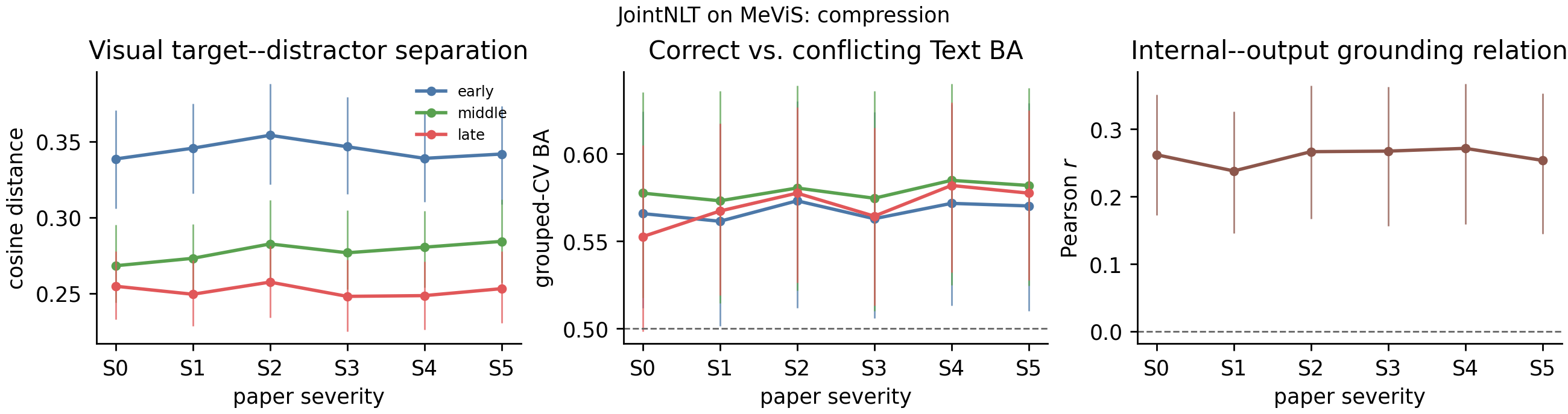}
  \caption{JointNLT MeViS internal trajectories under compression.}
  \label{fig:supp_jointnlt_degradation_compression}
\end{figure*}

JointNLT is evaluated on the clean MeViS manifest across all displayed
severities. The query frame is clean; tracker state is reset for every run;
and features are sampled at three fixed co-visible frames per source video.
Figures~\ref{fig:supp_jointnlt_degradation_blur},
\ref{fig:supp_jointnlt_degradation_noise},
\ref{fig:supp_jointnlt_degradation_pixelation}, and
\ref{fig:supp_jointnlt_degradation_compression} provide the four
perturbation-specific trajectories, and
Table~\ref{tab:supp_jointnlt_degradation_terminal} records their terminal
late-fusion values.

\FloatBarrier
\begin{center}
  \centering
  \scriptsize
  \setlength{\tabcolsep}{3pt}
  \begin{tabular}{@{}lrrr@{}}
    \toprule
    Perturbation & late visual sep. & late fused-text BA & output association $r$ \\
    \midrule
    blur        & $.173$ [$ .154,.192$] & $.582$ [$ .531,.633$] & $-.009$ [$-.111,.089$] \\
    noise       & $.042$ [$ .036,.048$] & $.566$ [$ .515,.621$] & $+.268$ [$ .160,.363$] \\
    pixelation  & $.148$ [$ .130,.168$] & $.539$ [$ .485,.593$] & $+.091$ [$-.051,.225$] \\
    compression & $.253$ [$ .230,.278$] & $.577$ [$ .528,.625$] & $+.254$ [$ .145,.353$] \\
    \bottomrule
  \end{tabular}
  \captionof{table}{JointNLT late-fusion $S_5$ values by perturbation.
  Brackets are 95\% source-video bootstrap CIs.}
  \label{tab:supp_jointnlt_degradation_terminal}
\end{center}

\section{Ablations}
\label{sec:supp_ablations}

This section reports protocol and intervention sensitivity checks.

\subsection{Native-box-metrics sensitivity check}
\label{sec:supp_bbox_sensitivity}

The four box trackers are also scored with native box IoU against tight
mask-extent ground-truth boxes ($\mathrm{IoU}\!\ge\!0.5$), using the same
predictions and clustered inference as the point-in-mask metric.
Table~\ref{tab:supp_bbox_s0} reports the clean-input point and box-IoU
effects side by side.

\begin{table}[t]
  \centering
  \scriptsize
  \setlength{\tabcolsep}{3pt}
  \begin{tabular}{@{}llrrr@{}}
    \toprule
    Model & Dataset & point & box-IoU & diff \\
    \midrule
    \multicolumn{5}{@{}l}{\textit{$B_T$}} \\
    UVLTrack & DanceTrack   & $-0.8\%$  & $+0.4\%$  & $+1.2$ \\
    UVLTrack & SportsMOT    & $+1.2\%$  & $+0.5\%$  & $-0.7$ \\
    UVLTrack & PersonPath22 & $\underline{+5.3\%}$  & $+4.1\%$  & $-1.2$ \\
    UVLTrack & MeViS        & $+0.6\%$  & $+1.3\%$  & $+0.7$ \\
    MMTrack  & DanceTrack   & $-0.3\%$  & $+1.3\%$  & $+1.7$ \\
    MMTrack  & SportsMOT    & $-2.3\%$  & $-0.5\%$  & $+1.8$ \\
    MMTrack  & PersonPath22 & $+0.5\%$  & $-2.5\%$  & $-3.0$ \\
    MMTrack  & MeViS        & $+0.4\%$  & $+0.5\%$  & $+0.1$ \\
    JointNLT & DanceTrack   & $\underline{+10.7\%}$ & $\underline{+11.1\%}$ & $+0.4$ \\
    JointNLT & SportsMOT    & $\underline{-11.2\%}$ & $\underline{-11.6\%}$ & $-0.3$ \\
    JointNLT & PersonPath22 & $\underline{+17.6\%}$ & $\underline{+17.1\%}$ & $-0.6$ \\
    JointNLT & MeViS        & $\underline{+10.6\%}$ & $\underline{+12.0\%}$ & $+1.5$ \\
    DUTrack  & DanceTrack   & $+0.7\%$  & $+0.2\%$  & $-0.5$ \\
    DUTrack  & SportsMOT    & $-1.8\%$  & $-1.3\%$  & $+0.5$ \\
    DUTrack  & PersonPath22 & $-1.9\%$  & $-0.2\%$  & $+1.7$ \\
    DUTrack  & MeViS        & $+0.2\%$  & $-0.0\%$  & $-0.2$ \\
    \midrule
    \multicolumn{5}{@{}l}{\textit{$S_D$}} \\
    UVLTrack & DanceTrack   & $-0.8\%$  & $+0.9\%$  & $+1.7$ \\
    UVLTrack & SportsMOT    & $-3.0\%$  & $-3.3\%$  & $-0.3$ \\
    UVLTrack & PersonPath22 & $-0.3\%$  & $-0.5\%$  & $-0.2$ \\
    UVLTrack & MeViS        & $+0.4\%$  & $+0.3\%$  & $-0.1$ \\
    MMTrack  & DanceTrack   & $+0.9\%$  & $\underline{+1.3\%}$  & $+0.5$ \\
    MMTrack  & SportsMOT    & $+0.0\%$  & $+0.0\%$  & $+0.0$ \\
    MMTrack  & PersonPath22 & $+0.9\%$  & $+0.2\%$  & $-0.8$ \\
    MMTrack  & MeViS        & $-0.7\%$  & $-0.6\%$  & $+0.1$ \\
    JointNLT & DanceTrack   & $+4.6\%$  & $+6.0\%$  & $+1.3$ \\
    JointNLT & SportsMOT    & $+1.9\%$  & $+2.6\%$  & $+0.8$ \\
    JointNLT & PersonPath22 & $+12.3\%$ & $+11.4\%$ & $-0.9$ \\
    JointNLT & MeViS        & $\underline{+9.2\%}$  & $\underline{+7.4\%}$  & $-1.9$ \\
    DUTrack  & DanceTrack   & $-0.3\%$  & $+0.0\%$  & $+0.3$ \\
    DUTrack  & SportsMOT    & $+0.0\%$  & $+0.0\%$  & $+0.0$ \\
    DUTrack  & PersonPath22 & $+0.0\%$  & $+0.0\%$  & $+0.0$ \\
    DUTrack  & MeViS        & $-0.5\%$  & $-0.1\%$  & $+0.3$ \\
    \bottomrule
  \end{tabular}
  \caption{Clean-input $B_T$ and $S_D$ under point-in-mask and box-IoU
  scoring. ``diff'' is box-IoU minus point, in percentage points; underlining
  denotes a 95\% source-video bootstrap CI excluding zero.}
  \label{tab:supp_bbox_s0}
\end{table}

\subsection{Causal donor-content controls}
\label{sec:supp_donor_content_controls}

Each control replaces the state at the main-paper patch site while holding
visual input, initialization, and sequential replay fixed.

\paragraph{JointNLT content controls.}

Table~\ref{tab:supp_jointnlt_contrasts} lists the donor-state contrasts for
the retention-control and persistent-capture strata.

\begin{center}
  \centering
  \scriptsize
  \setlength{\tabcolsep}{3pt}
  \begin{tabular}{@{}lrrr@{}}
    \hline
    Contrast (correct $-$) & Mean $\Delta$ & 95\% CI & $q$ \\
    \hline
    \multicolumn{4}{@{}l}{\textit{Retention controls (n=67)}} \\
    neutral       & $+0.176$ & [$0.090$,$0.274$] & $.0001$ \\
    unrelated     & $+0.184$ & [$0.101$,$0.280$] & $.0001$ \\
    cross-video   & $+0.226$ & [$0.140$,$0.326$] & $.0000$ \\
    self (effect) & $+0.049$ & [$0.022$,$0.080$] & $.0006$ \\
    \hline
    \multicolumn{4}{@{}l}{\textit{Persistent captures (n=23)}} \\
    neutral       & $+0.359$ & [$0.077$,$0.622$] & $.0251$ \\
    unrelated     & $+0.193$ & [$-0.045$,$0.413$] & $.1241$ \\
    cross-video   & $+0.368$ & [$0.110$,$0.634$] & $.0206$ \\
    self (effect) & $+1.193$ & [$0.971$,$1.410$] & $.0000$ \\
    \hline
  \end{tabular}
  \captionof{table}{JointNLT donor-state contrasts by outcome stratum.
  Values are FDR-corrected within each layer--stratum family.}
  \label{tab:supp_jointnlt_contrasts}
\end{center}

\paragraph{Molmo2 confirmation strata.}

Table~\ref{tab:supp_molmo_patch_confirmation} lists the L6/L12 donor-state
contrasts by confirmation stratum.

\begin{center}
  \centering
  \scriptsize
  \setlength{\tabcolsep}{2.5pt}
  \begin{tabular}{@{}llrlrr@{}}
    \toprule
    Source state & Stratum & $n$ & Layer & Mean $\Delta m$ [95\% CI] & $p$ \\
    \midrule
    correct & conflict capture & 15 & L6  & $+.868$ [$+.477$,$+1.266$] & $.0008$ \\
            &                  & 15 & L12 & $+.861$ [$+.470$,$+1.261$] & $.0012$ \\
    correct & other/error      & 12 & L6  & $+.242$ [$+.068$,$+.429$] & $.0332$ \\
            &                  & 12 & L12 & $+.232$ [$+.073$,$+.410$] & $.0604$ \\
    correct & target retention & 88 & L6  & $+.013$ [$-.019$,$+.045$] & $.4253$ \\
            &                  & 88 & L12 & $+.021$ [$+.000$,$+.044$] & $.0618$ \\
    \midrule
    unrelated & conflict capture & 15 & L6  & $+.770$ [$+.411$,$+1.136$] & $.0037$ \\
              &                  & 15 & L12 & $+.789$ [$+.401$,$+1.181$] & $.0027$ \\
    unrelated & other/error      & 12 & L6  & $+.029$ [$-.157$,$+.218$] & $.8219$ \\
              &                  & 12 & L12 & $+.133$ [$+.010$,$+.281$] & $.1282$ \\
    unrelated & target retention & 88 & L6  & $+.007$ [$-.023$,$+.036$] & $.6450$ \\
              &                  & 88 & L12 & $+.005$ [$-.028$,$+.038$] & $.7749$ \\
    \midrule
    correct$-$unrelated & conflict capture & 15 & L6  & $+.098$ [$+.020$,$+.218$] & $.0241$ \\
                         &                  & 15 & L12 & $+.072$ [$-.024$,$+.204$] & $.3134$ \\
    correct$-$unrelated & other/error      & 12 & L6  & $+.213$ [$+.051$,$+.446$] & $.0308$ \\
                         &                  & 12 & L12 & $+.099$ [$+.028$,$+.185$] & $.0635$ \\
    correct$-$unrelated & target retention & 88 & L6  & $+.006$ [$-.037$,$+.051$] & $.7952$ \\
                         &                  & 88 & L12 & $+.016$ [$-.016$,$+.052$] & $.3774$ \\
    \bottomrule
  \end{tabular}
  \captionof{table}{Molmo2 L6/L12 donor-state contrasts. $\Delta m$ is the
  trajectory-mean target-minus-distractor point-margin change from the
  conflicting baseline; CIs use source-video bootstrap and $p$ is paired
  sign-flip. The bottom block compares correct and unrelated donors.}
  \label{tab:supp_molmo_patch_confirmation}
\end{center}

\paragraph{Causal audit of the behaviorally null systems.}
\label{sec:supp_null_causal_audit}

The three small-effect systems are tested on the 115-video manifest with
matched-correct replacement and zero-text ablation at their declared text
boundaries. Contrasts use the conflicting baseline; $q$ is FDR-corrected
within each system's two-contrast family. Table~\ref{tab:supp_null_causal_audit}
reports the resulting semantic and presence contrasts.

\begin{center}
  \centering
  \scriptsize
  \setlength{\tabcolsep}{3pt}
  \begin{tabular}{@{}llrrl@{}}
    \hline
    Model & Contrast (vs.\ conflicting) & Mean $\Delta$ & $q$ & Equiv.\ \\
    \hline
    UVLTrack & matched-correct (semantic) & $+0.024$ & $.365$ & No \\
             & zero-ablation (presence)   & $+0.036$ & $.007$ & No \\
    \hline
    MMTrack  & matched-correct (semantic) & $-0.012$ & $.099$ & \textbf{Yes} \\
             & zero-ablation (presence)   & $-0.022$ & $.021$ & No \\
    \hline
    DUTrack  & matched-correct (semantic) & $+0.002$ & $.573$ & \textbf{Yes} \\
             & zero-ablation (presence)   & $-0.013$ & $.379$ & \textbf{Yes} \\
    \hline
  \end{tabular}
  \captionof{table}{Output-margin change after matched-correct replacement
  (semantic) or zero-text ablation (presence), by system.}
  \label{tab:supp_null_causal_audit}
\end{center}

\clearpage
\subsection{Outcome-independent sample confirmation}
\label{sec:supp_outcome_independent_confirmation}

The main-paper MOT subsets intentionally retain rare Molmo2
conflicting-text takeover cases before seeded random filling. This confirmation
instead retains the complete annotation-eligible DanceTrack, SportsMOT, and
PersonPath22 pool: 404 pairs from 81 source videos (214/17, 110/31, and
80/33 pairs/videos, respectively). Pair inclusion requires at least five
co-visible annotated frames, mean co-visible mask IoU below 0.3, and the
required target, distractor, and deterministic category-matched unrelated
expressions; it uses no model output, accuracy, or takeover label.

JointNLT, UVLTrack, MMTrack, and DUTrack (Molmo2 is out of scope) were run under all four text conditions
at every displayed $S_0$--$S_5$ severity level of blur, noise, pixelation, and
compression.

\begin{figure}[t]
  \centering
  \includegraphics[width=\linewidth]{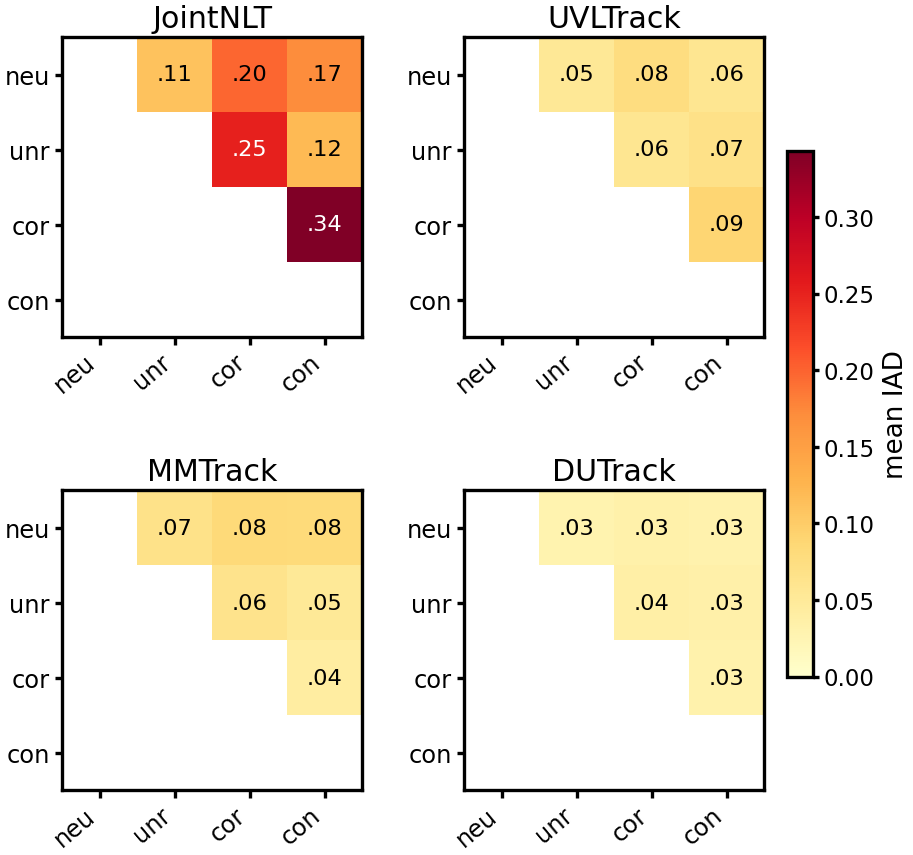}
  \caption{Outcome-independent pool: target-accuracy IAD, averaged equally
  across the three MOT sources and four perturbations over paper $S_0$--$S_5$.
  Labels abbreviate neutral/unrelated/correct/conflicting.}
  \label{fig:supp_oi_iad_confirmation_compact}
\end{figure}

Figure~\ref{fig:supp_oi_iad_confirmation_compact} compactly summarizes each
system's target-accuracy IAD matrices.  Within a model, we equally average the
three source-dataset matrices and, within a dataset, the four
perturbation-specific IAD matrices over the paper's six-position severity
ladder. Larger cells therefore indicate a larger average absolute condition
gap, not a directional or inferential test.

\paragraph{Effect pattern.}
Figure~\ref{fig:supp_oi_iad_confirmation_compact} retains the main behavioral
split under the outcome-independent distribution.
JointNLT's correct--conflicting IAD is $.50$, $.18$, and $.35$ on
DanceTrack, SportsMOT, and PersonPath22, respectively. The corresponding
cells remain smaller for UVLTrack ($.13$, $.06$, $.09$), MMTrack ($.03$,
$.06$, $.04$), and DUTrack ($.02$, $.04$, $.03$).
\section{Reproducibility}
\label{sec:supp_reproducibility}

We release the complete experimental pipeline as a self-contained,
anonymized repository in the supplementary material, structured to
reproduce every table, figure, and statistical test in this paper
through a single configuration file and entry point rather than the
roughly twenty scripts and five separate Python environments the
pipeline spans internally. The repository provides the inference,
validation, consolidation, statistics, and model-internals code exactly
as used to generate the reported results; it introduces no
reimplementation of the underlying methodology.

The archive does not redistribute raw third-party video or image frames.
It provides code, frozen manifests and provenance metadata, annotations only
where their upstream terms permit redistribution, and scripts that obtain and
prepare source media from the original providers (Table~\ref{tab:supp_models_access},
Table~\ref{tab:supp_datasets_access}). The packaged operational smoke fixture
is an author-created synthetic two-object sequence. It exercises the shared
specialized-tracker input, four-condition text, perturbation, output-schema,
and validation paths, but is not a scientific replication or an estimate of
tracking or text-prior performance. Reproducing the full five-model study
requires obtaining the source media, annotations, and checkpoints under their
respective upstream terms. A CPU-only test suite covering adapter contracts,
outcome-label invariants, and the statistical correction routines requires
neither a GPU nor a model checkpoint.

Scaling from the synthetic operational fixture to the full paper-scale grid
(every selected pair, all six severities) requires obtaining the complete
datasets and model checkpoints per their documented licenses
(Table~\ref{tab:supp_models_access}, Table~\ref{tab:supp_datasets_access})
and otherwise leaves the pipeline unchanged. The repository is
anonymized for review; the complete, de-anonymized version
will be made publicly available upon acceptance.

\section{Ethics and Societal Impact}
\label{sec:supp_ethics}

DISTRACT evaluates existing vision-language trackers; it does not train a
tracker, collect new personal data, or release raw third-party media. Its
purpose is to reveal failures under ordinary image degradation, not to certify
a model as safe for deployment. Tracking can be used in privacy-sensitive or
surveillance contexts, although the code adds no surveillance-specific
capability. Users must obtain data lawfully, respect upstream terms and privacy
obligations, and apply suitable oversight and safeguards before operational use.

\clearpage